\documentclass[runningheads]{llncs}
\usepackage{pifont}
\usepackage{multirow}
 
\usepackage{eccv}

\usepackage{eccvabbrv}
\usepackage{tabularx}
\usepackage{graphicx}
\usepackage{xcolor}
\usepackage{listings}
\usepackage{wrapfig}
\usepackage{booktabs}
\usepackage[skins]{tcolorbox}
\usepackage{colortbl}
\newcommand{\Qbox}[1]{%
    
    \begin{tcolorbox}[colframe=black!70, colback=cyan!2, boxrule=1pt, arc=2mm,   top=2pt, bottom=3pt, left=3pt, right=3pt,  fontupper=\scriptsize,
  boxsep=1pt]
        #1
    \end{tcolorbox}
}
\definecolor{backcolour}{rgb}{0.95,0.95,0.92}
\lstdefinestyle{mystyle}{
    backgroundcolor=\color{backcolour},
    basicstyle=\ttfamily\footnotesize,
    breaklines=true,
    numbers=left,
    numbersep=5pt,
    frame=single
}
\usepackage[accsupp]{axessibility}  

\usepackage{hyperref}

\usepackage{orcidlink}
\newcommand{\name}{OmniCoT}

\begin{document}

\title{OmniCoT: A Benchmark for Global and Multi-Step Panoramic Reasoning} 


\author{Haocong He\textsuperscript{1,*}, Chenfei Liao\textsuperscript{2,*,\dag}, Zichen Wen\textsuperscript{1,7}, Zihao Dongfang\textsuperscript{2},\\ Xu Zheng\textsuperscript{2},  
Bin Ren\textsuperscript{3}, Chang Su\textsuperscript{4},  Zixin Zhang\textsuperscript{2},  Harold H. Chen\textsuperscript{2}, \\ Hongfei Zhang\textsuperscript{2}, Weijia Li\textsuperscript{5,7}, Kailun Yang\textsuperscript{6}, Conghui He\textsuperscript{7}, Xuming Hu\textsuperscript{2,9}, \\ Nicu Sebe\textsuperscript{8}, Linfeng Zhang\textsuperscript{1,\ddag}}

\authorrunning{H. He et al.}

\institute{\textsuperscript{1}SJTU,
\textsuperscript{2}HKUST(GZ),
\textsuperscript{3}MBZUAI,
\textsuperscript{4}JLU, 
\textsuperscript{5}THU, 
\textsuperscript{6}HNU, \\
\textsuperscript{7}Shanghai AI Lab, 
\textsuperscript{8}UniTrento, 
\textsuperscript{9}HKUST  \\
\textsuperscript{*}Equal Contribution \quad 
\textsuperscript{\dag}Project Lead \quad
\textsuperscript{\ddag}Corresponding Author \\
\email{cliao127@connect.hkust-gz.edu.cn}\\
\email{\{haocong-he,zhanglinfeng\}@sjtu.edu.cn}\\
Dataset: \url{https://huggingface.co/datasets/Eustia1/OmniCoT/}\\
Model: \url{https://huggingface.co/Eustia1/OmniCoT-R1}\\
Code: \url{https://github.com/Chenfei-Liao/OmniCoT} \\
Page: \url{https://chenfei-liao.github.io/OmniCoT.github.io/}
}

\maketitle

\begin{figure}
    \centering

    \includegraphics[width=\linewidth]{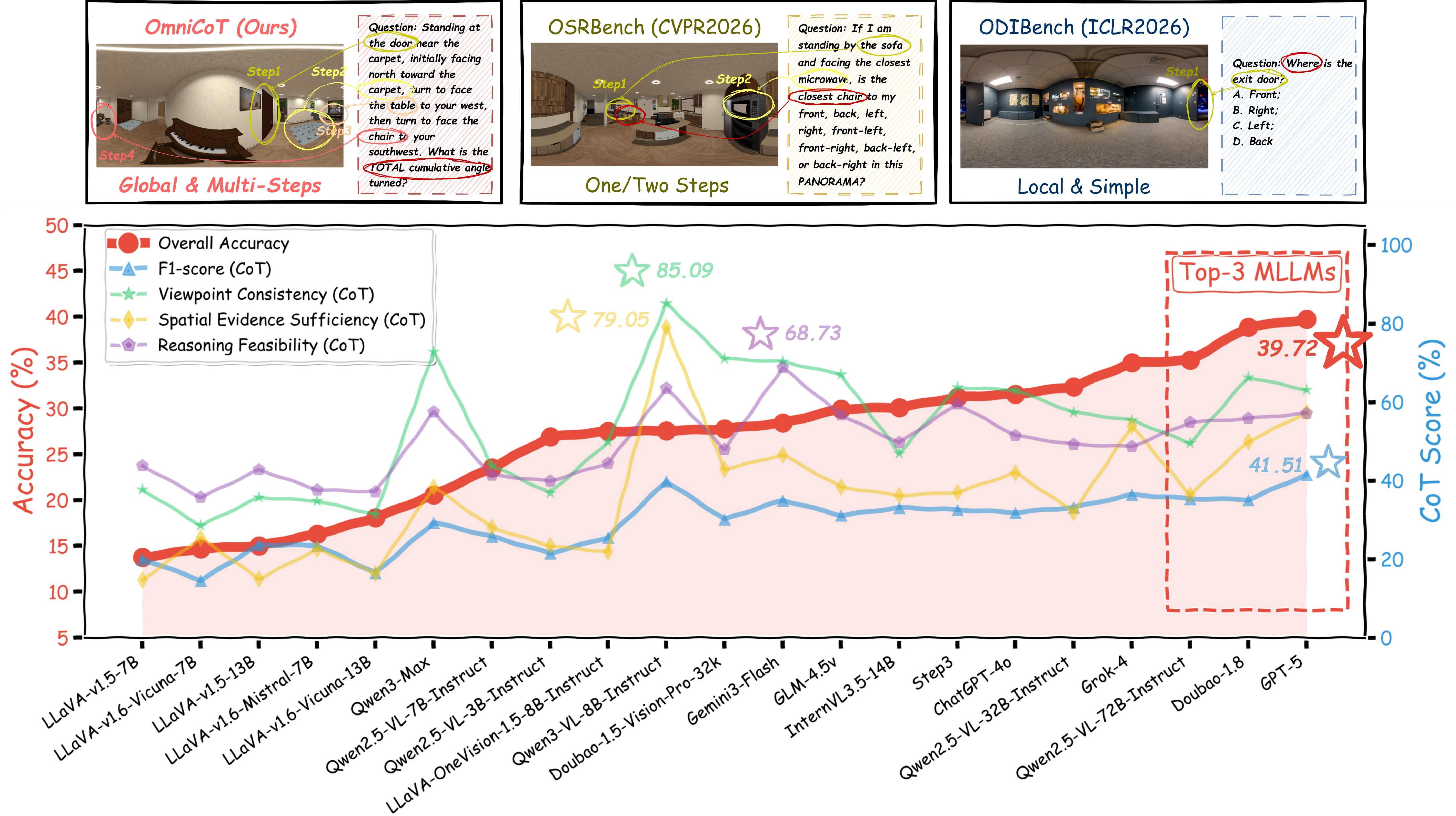}
    \vspace{-2.2em}
\caption{\textbf{(\textit{Top}) Comparison with existing panoramic benchmarks (\textit{e.g.}, OSRBench}~\cite{dongfang2025multimodal}\textbf{, ODIBench}~\cite{yang2025odi}\textbf{).} 
While prior works solely rely on synthetic data and primarily focus on few-step reasoning or local visual cues, OmniCoT demands comprehensive multi-hop inference across the full $360^\circ$ view and uniquely incorporates a manually annotated real-world subset, driving MLLMs to fundamentally \textit{\textbf{see more and reason more}}. 
\textbf{(\textit{Bottom}) Leaderboard of MLLMs on OmniCoT-B}.
It showcases both the overall accuracy and detailed Chain-of-Thought (CoT) quality metrics.}
    \label{fig:teaser}

\end{figure}

\begin{abstract}
Multimodal Large Language Models~(MLLMs) have demonstrated promising spatial reasoning capabilities, while these abilities remain underexplored in the emerging visual modality of panoramic imagery. The full $360^\circ \times 180^\circ$ field of view of panoramas essentially supports complex global multi-step reasoning, which is also the fundamental advantage of panoramas in applications such as embodied intelligence. However, existing panoramic benchmarks largely focus on simplistic queries that rely on local cues or single-/few-step reasoning, thereby ignoring the fundamental advantage of panoramas and failing to fully exploit their potential. To address this gap, we introduce \textbf{\textit{OmniCoT}}, a panoramic spatial reasoning suite designed to enable MLLMs to use global evidence and perform multi-step inference across viewpoints. It includes \textbf{\textit{OmniCoT-B}}~(6.7K data) for evaluation, which measures both answer accuracy and reasoning quality, \textbf{\textit{OmniCoT-Real}}~(1K data) as a manually annotated real-world subset to quantify the Sim-to-Real gap.  For training, \textbf{\textit{OmniCoT-T}}~(14.3K data) is purpose-built with structured stepwise Chain-of-Thought annotations that explicitly link intermediate reasoning steps to panoramic evidence. Based on OmniCoT-T, we introduce \textbf{\textit{OmniCoT-R1}} and adopt a two-stage training strategy tailored to the geometrically complex panoramic space,
where Supervised Fine-Tuning (SFT) anchors reasoning to panoramic evidence (e.g., bearings, proximity) and GRPO penalizes geometrically incoherent paths to consolidate global $360^\circ$ spatial consistency.
Through OmniCoT, we aim to \textbf{\textit{recalibrate the difficulty of panoramic spatial reasoning to better align with the intrinsic capabilities of panoramic imagery}}, thereby fostering meaningful progress in this research area.
\end{abstract}

\section{Introduction}

Panoramas, as an emerging visual modality, provide a unique $360^{\circ}{\times}180^{\circ}$ field of view (FoV) for robots, vehicles, and related applications~\cite{zhang2026panoramic,lin2025one,wang2026panoworld,zhong2025omnisam}. 
By capturing more of the surrounding environment, this wide field of view is especially useful for tasks such as autonomous driving and embodied intelligence~\cite{yang2025efficientvla,yang2026uaor,zhang2026panoramic}. 
However, severe geometric distortions and dense visual content introduce substantial domain gaps, creating new challenges when extending MLLMs to understand panoramas.
To evaluate the capability of MLLM in panoramic environments, VQA stands out as the most common way to test whether MLLMs can truly understand and reason about the entire $360^{\circ}$ environment.

With more panoramic benchmarks proposed, a closer look shows a significant flaw, as shown in Fig.~\ref{fig:teaser} and Table~\ref{tab::comparison}: most existing questions in current benchmarks remain overly simplistic, failing to compel state-of-the-art MLLMs~\cite{team2026kimi,wen2026innovator,wen2025efficient} to leverage the full $360^{\circ}$ view.
The currently proposed panoramic spatial reasoning benchmarks focus on only local regions (\textit{e.g.}, ``Where is the exit door''), and the reasoning tasks tend to require no more than two-hop inference.
We believe that \textbf{\textit{panoramas are built for complex spatial tasks}}, because they show how every object in a full $360^{\circ}$ view relates to the others, regardless of their distance or direction. 
When questions are too simple, they ignore the wealth of information in panoramas. 
To bridge this gap, we introduce \textbf{\textit{OmniCoT}}\footnote{Here ``Omni'' refers to panoramic (omni-directional) input rather than omni-modal input~\cite{yagi1999omnidirectional}.}, which includes a challenging panoramic reasoning benchmark\textbf{~\textit{OmniCoT-B}}~(6.7K data) that pushes models to \textit{\textbf{``see more and reason more''}}, a manually annotated real-world subset\textbf{~\textit{OmniCoT-Real}}~(1K data) to quantify the Sim-to-Real gap, and a training set\textbf{~\textit{OmniCoT-T}}~(14.3K data).

\begin{table*}[t]
    \caption{\textbf{Comparisons of our OmniCoT} with other omnidirectional spatial reasoning benchmarks.}
    \label{tab::comparison}
    \centering
    \scriptsize
    \renewcommand\arraystretch{1}
    \resizebox{\linewidth}{!}{
    \begin{tabular}{l r@{}c@{}l c|ccc|ccc}
        \toprule
        Benchmark & \multicolumn{3}{c}{Spatial Scope} & Annotation & Syn.  & Real. & CoT & \# of Q Types& \# of Images & \# of QA \\
        \midrule
        VQA 360\textdegree~\cite{chou2020visual} &Global & / & Single-step& Auto & \ding{56} &\ding{52}  &\ding{56}  & 5 & 1.4k
        & 16.9k\\
        \midrule
        PanoEnv-QA~\cite{lin2026panoenv}  & Global & / & Single-step & Auto & \ding{52} & \ding{56}  & \ding{56} & 5 & 595 & 14.8k\\ 
        \midrule
        OmniVQA ~\cite{zhang2025towards} & Local & / & Single-step & Auto & \ding{52} & \ding{56}  & \ding{56} & 3 & 1.2k & 4.8k\\ 
        \midrule
        OSR-Bench~\cite{dongfang2025multimodal} & Local & / & Few-step & Auto & \ding{52} & \ding{56}  & \ding{56} & 3 & 4.1k &153k\\ 
        \midrule
        ODI-Bench~\cite{yang2025odi} & Local & / & Few-step & Auto\&Manual & \ding{52} & \ding{52} & \ding{56} & 10 & 2k & 4.3k \\
        \midrule
        \textbf{\name~(Ours)}  & Global & / & Multi-hop & Auto\&Manual & \ding{52}& \ding{52} &  \ding{52}  & 6 & 4.2k  & 21.6k\\
        \bottomrule
    \end{tabular}
    }
\end{table*}

First, building upon real-world embodied perception requirements, we introduce a structured question taxonomy that decomposes panoramic spatial reasoning into three progressive dimensions: \textit{\textbf{See-Locate-Move}}. 
This taxonomy enables the systematic generation of multi-hop questions spanning viewpoint transformation, inter-object relational reasoning, and embodied action simulation in panoramic scenes.
Then, we design a hybrid pipeline that integrates automated generation with expert supervision to construct high-quality question–answer pairs requiring multi-hop reasoning and global information.
Based on OmniCoT-B, we perform extensive tests on a wide range of MLLMs, evaluating both their general and spatial reasoning abilities. 
Building upon the benchmark, we further develop OmniCoT-R1 based on OmniCoT-T through a two-stage training strategy, which serves as the baseline of panoramic reasoning MLLMs. 
The strategy first establishes structured reasoning via SFT, and then enhances the model's ability to solve complex multi-step tasks with GRPO. 
Moreover, we collect and annotate 1K real-world data to construct OmniCoT-Real, enabling the evaluation of the Sim-to-Real gap on OmniCoT.

At a glance, our contributions can be summarized as follows: 
\ding{182} We introduce \textbf{\textit{OmniCoT-B}}, a new benchmark that challenges MLLMs to fully use the $360^{\circ}$ space. It moves beyond simple, local questions, requiring MLLMs to see more and reason more.
\ding{183} We provide a thorough evaluation of current MLLMs, measuring not just if their final answers are correct, but whether their step-by-step reasoning actually makes sense.
\ding{184} We develop \textbf{\textit{OmniCoT-R1}} based on our proposed \textbf{\textit{OmniCoT-T}} through a two-stage training strategy, which serves as the new baseline of the complex panoramic spatial reasoning task.  
\ding{185} We present \textbf{\textit{OmniCoT-Real}}, a set of real-world panoramas used to further measure the Sim-to-Real gap of OmniCoT.
\ding{186} We provide diagnostic analyses in the Supplementary, including paraphrase robustness, input-projection robustness, tolerance sensitivity, human verification, and data scaling of OmniCoT-T.

\section{Related Work}
\label{sec:formatting}
\noindent \textbf{MLLMs for Omnidirectional Vision.}\quad
While pinhole images only show a narrow slice of a scene, panoramas capture everything in a full $360^{\circ}$ view~\cite{lin2025one}. 
This complete FoV is essential for embodied AI and virtual reality~\cite{zheng2025panorama}. 
This makes panoramic imagery an ideal testing ground for MLLMs aiming to master higher-level spatial reasoning~\cite{dongfang2025multimodal}.
Current research in this area focuses on two main goals: \ding{172} building better datasets and \ding{173} adapting models for panoramic vision. 
For instance, datasets like OSR-Bench~\cite{dongfang2025multimodal} and ODI-Bench~\cite{yang2025odi} use a mix of MLLMs and human effort to create panoramic question-answering tasks. 
Meanwhile, researchers are fine-tuning models with techniques like GRPO to help them navigate the distorted, wide-angle world of panoramas: examples include 360-R1~\cite{zhang2025towards} for spatial reasoning and ERP-RoPE~\cite{zhou2025dense360} for better image processing.
However, there is a catch: most of these models are still tested on very basic questions that only look at one small spot or require just one or two simple steps. 
This fails to use the full potential of panoramas and doesn't reflect how they would be used in the real world. 
Thus, we introduce OmniCoT, which is designed to calibrate the difficulty of panoramic spatial reasoning benchmarks by requiring MLLMs to see more and reason more.

\noindent \textbf{MLLM Benchmarks for Spatial Reasoning.}\quad 
Spatial reasoning research has progressed from object recognition to goal-oriented tasks such as navigation and relational reasoning~\cite{yu2025far, liu2023visual, zheng2025multimodal, daxberger2025mm}. 
These benchmarks evaluate how MLLMs understand the physical world by mapping surroundings and planning actions~\cite{guo2024surds,xu2025spatialbench,stogiannidis2025mind, cai2025spatialbot}. 
However, a key gap remains in evaluating reasoning within $360^{\circ}$ environments. 
Existing panoramic benchmarks often treat spatial reasoning as a ``black box''~\cite{yang2025odi, dongfang2025multimodal,zhang2025towards,lin2026panoenv, chou2020visual}, focusing only on final-answer correctness. 
This is inadequate for panoramas, where dense visual information enables models to exploit local shortcuts~\cite{geirhos2020shortcut, agrawal2018don} instead of true global reasoning. 
We argue that panoramic intelligence should be evaluated not only by outcomes but also by reasoning processes~\cite{lightman2023let, zheng2025processbench}. 
While prior benchmarks emphasize accuracy, OmniCoT-T focuses on reasoning chains~\cite{wei2022chain,jiang2025mme}, requiring step-by-step decomposition of thought processes to provide a more transparent and rigorous evaluation of spatial reasoning in $360^{\circ}$ environments.

\section{OmniCoT Design \& Generation}

\subsection{Question Dimension: ``See-Locate-Move''}

Inspired by real-world embodied requirements, we design our complex panoramic spatial reasoning questions from three progressive dimensions: \textit{\textbf{``see'', ``locate'' and ``move''}}. 
The first dimension, ``see'', focuses on viewpoint transformation. 
It tests the MLLMs' ability to decouple their view from the camera's initial pose and multi-hop viewpoint reasoning, requiring them to mentally synthesize views from arbitrary angles and understand the spherical geometry of the panorama.
The second dimension, ``locate'', advances to spatial object relationships. Questions of this dimension evaluate MLLMs' ability to reason about the inter-object spatial relationship, verifying whether it can perceive the scene not just as a bag of pixels, but as a structured layout of interconnected entities.
The final dimension, ``move'', introduces embodied action simulation. This dimension challenges MLLMs to execute virtual movements and predict the visual consequences of these actions, thus evaluating their potential for real-world scenarios.

\noindent    \ding{227} \textbf{[See] Multi-hop Viewpoint Transformation.}\quad
     This category assesses the MLLM's fundamental ability to observe the world through multi-step panoramas. 
This involves not only understanding spherical projections but also the ability to perform precise geometric integration.
Specifically, \ding{182} Multi-Step Orientation Tracking (MOT) requires the model to start from an initial pose and execute a series of relative rotations, finally identifying the target object. 
\ding{183} Relative Angular Calculation (RAC) challenges the model to compute the cumulative angular displacement or the specific bearing between multiple landmarks, treating objects as dynamic reference points. 
These tasks verify if the model can maintain spatial constancy even when the target undergoes significant perspective warping near the panoramic poles or across the image boundaries.

\Qbox{
\textit{\textbf{Type A1 Example:} Standing at the \textbf{\textcolor{blue!30}{[object\_desc]}}, facing \textbf{\textcolor{orange!30}{[cardinal direction]}}, turn \textbf{\textcolor{brown!50}{[angle1]\textdegree [dir1]}}, then turn \textbf{\textcolor{brown!50}{[angle2]\textdegree [dir2]}}, \textbf{\textcolor{red!50}{what is the NEAREST object?}}}

\textit{\textbf{Type A2 Example:} Standing at \textbf{\textcolor{blue!30}{[object\_desc]}}, initially facing \textbf{\textcolor{blue!30}{[object\_desc]}}, turn to face \textbf{\textcolor{blue!30}{[object\_desc]}}, then turn to face \textbf{\textcolor{blue!30}{[object\_desc]}}. \textbf{\textcolor{red!50}{What is the TOTAL cumulative angle turned?}}} 
}

\noindent \ding{227}  \textbf{[Locate] Inter-Object Spatial Relationship.}\quad
     This category assesses the MLLM’s ability to reason about the spatial relationships between multiple objects within a scene through multi-hop reasoning. 
Rather than answering questions based on direct observation, the model must navigate chains of spatial relations and use qualifiers such as ``nearest'' or ``second closest'' to either identify a specific target object or determine its directional relationship to an anchor object.
Specifically, \ding{182} Multi-Hop Object Identification (MOI) requires the model to sequentially apply spatial qualifiers across multiple reference objects to find a unique target.  \ding{183} Multi-Hop Direction Identification (MDI) challenges the model to determine the final direction (\textit{e.g.}, North) of a target relative to an anchor.

\Qbox{
\textit{\textbf{Type B1 Example:} \textbf{\textcolor{red!50}{What}} \textbf{\textcolor{blue!30}{[object\_desc]}} is directly to the \textbf{\textcolor{orange!30}{[cardinal direction]}} of the NEAREST object that is  \textbf{\textcolor{orange!30}{[cardinal direction]}} of the  \textbf{\textcolor{blue!30}{[object\_desc]}}?}

\textit{\textbf{Type B2 Example:} \textbf{\textcolor{red!50}{In which direction}} is the  \textbf{\textcolor{blue!30}{[object\_desc]}} to the \textbf{\textcolor{orange!30}{[cardinal direction]}} of the \textbf{\textcolor{blue!30}{[object\_desc]}}, relative to the \textbf{\textcolor{blue!30}{[object\_desc]}} itself?}
}

\noindent \ding{227}  \textbf{[Move] Embodied Action Simulation.}\quad Furthermore, to evaluate the MLLM's ability to perform dynamic spatial reasoning through the simulation of embodied actions, such as movement and turning, the questions of Type C are introduced.
Within this category, the model must track continuous state changes including updates to position, orientation, and field of view, for either itself or objects within the scene.
Specifically, \ding{182} Pure Translational Movement (PTM) simulates navigation along a single direction without rotation, testing the model's capacity to predict objects encountered along a linear path.
\ding{183} Rotation-Translational Movement (RTM) combines straight-line movement with a subsequent rotation, requiring the model to update its egocentric perspective and reason about the altered spatial relationships and visibility.

\Qbox{
\textit{\textbf{Type C1 Example:} From the  \textbf{\textcolor{blue!30}{[object\_desc]}}, near the  \textbf{\textcolor{blue!30}{[object\_desc]}},, walk straight \textbf{\textcolor{orange!30}{[cardinal direction]}} for \textbf{\textcolor{brown!50}{[number]}} meters. \textbf{\textcolor{red!50}{What is the FIRST object you will encounter?}}}

\textit{\textbf{Type C2 Example:} From the  \textbf{\textcolor{blue!30}{[object\_desc]}},, walk \textbf{\textcolor{orange!30}{[cardinal direction]}}  \textbf{\textcolor{brown!50}{[number]}} meters toward the  \textbf{\textcolor{blue!30}{[object\_desc]}}, area, then turn \textbf{\textcolor{brown!50}{[angle]}} to face \textbf{\textcolor{orange!30}{[cardinal direction]}}. \textbf{\textcolor{red!50}{Is the}}  \textbf{\textcolor{blue!30}{[object\_desc]}}, \textbf{\textcolor{red!50}{still visible from your new position and facing direction?}}}
}

\subsection{Data Generation}

\begin{figure*}[t]
    \centering
    \includegraphics[width=\linewidth]{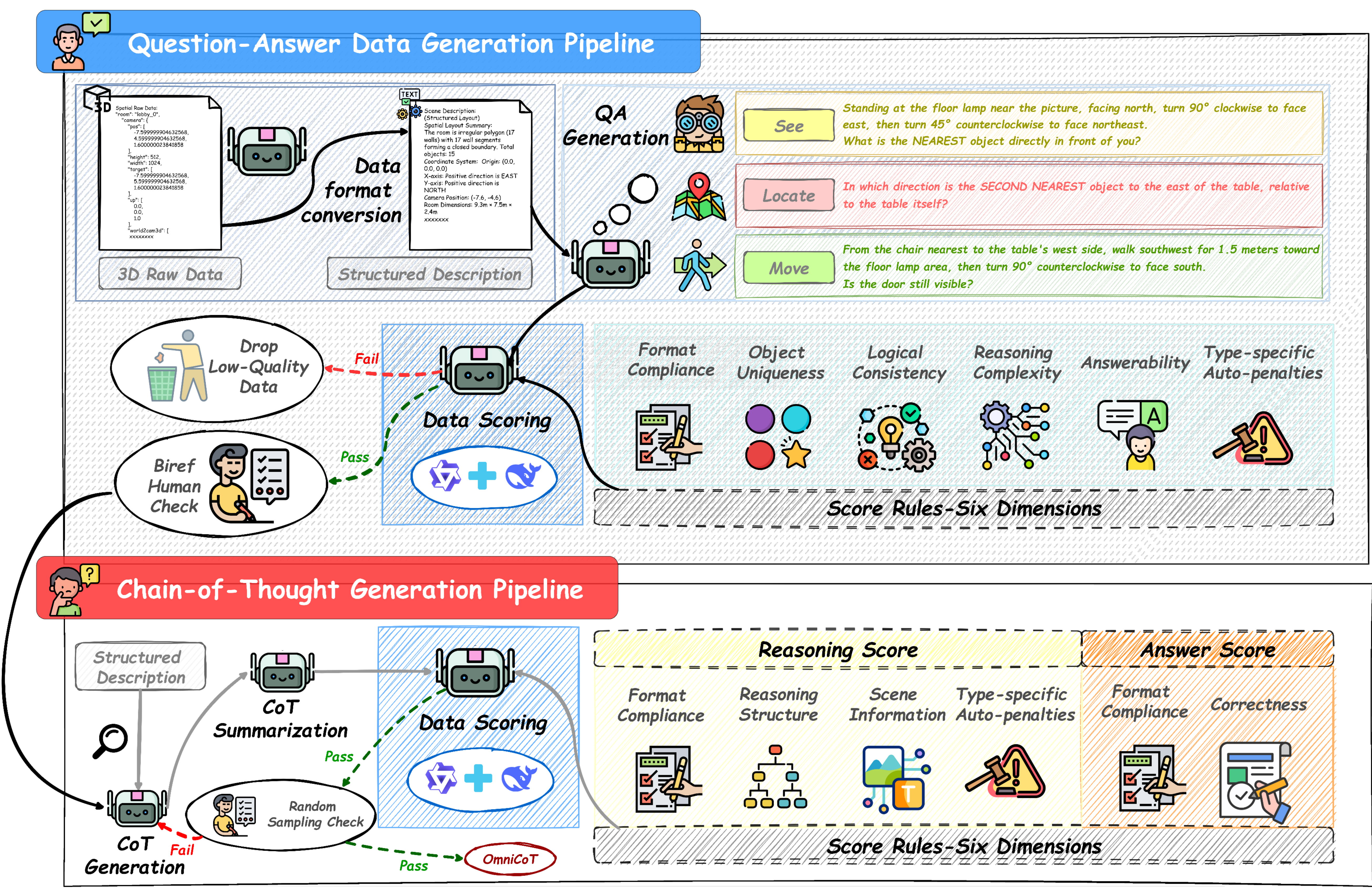}

    \caption{\textbf{Data Generation Pipeline of OmniCoT.} It translates 3D scene geometry into structured language representation, generates multidimensional candidate questions (See, Locate, Move), and applies rigorous dual-LLM reasoning judges to select high-quality QA pairs. Finally, a structured three-stage process synthesizes, refines, and evaluates the corresponding step-by-step Chain-of-Thought (CoT) rationales, culminating in a manually verified dataset.}
    \label{fig:pipeline}

\end{figure*}



\noindent\textbf{Step 1: Question-Answer (QA) Generation.}
In the process of QA generation shown in Fig.~\ref{fig:pipeline}, we propose a hierarchical generation pipeline to gradually generate and optimize the panoramic spatial reasoning QA pairs. 
The generation process begins with reliable raw 3D scene data, which is transformed into structured natural language descriptions. 
This representation provides rich and reliable spatial context for subsequent question generation and enables questions that fully leverage the global perspective afforded by panoramas. 
Building on this representation, we generate multiple batches of candidate questions covering all predefined question types.
We then employ two reasoning-focused large language models (DeepSeekv3.2~\cite{liu2025deepseek} and Qwen3-Max) as judges to score these candidates from six dimensions: format compliance, object uniqueness, logical consistency, reasoning complexity, answerability, and special type-specific auto-penalties.
Only candidate questions that meet the scoring criteria will proceed to the final manual review. 
Experts will quickly check the generated question-answer pairs to ensure there are no significant deviations in format or content.

\begin{figure*}[t!]
    \centering
    \includegraphics[width=\linewidth]{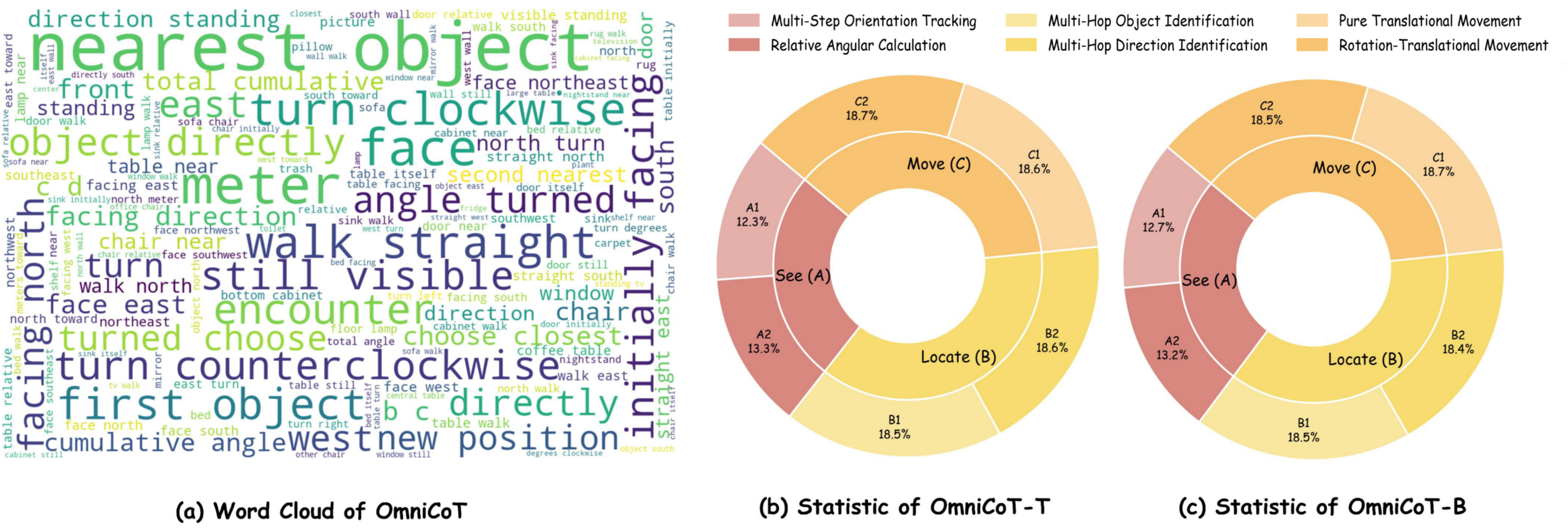}

    \caption{\textbf{OmniCoT exhibits broad lexical coverage and balanced question-type composition.}
    \textbf{(a)} Word cloud over all OmniCoT questions.
    \textbf{(b)} Question-type distribution of OmniCoT-T shows a balanced mixture of \textit{See--Locate--Move} taxonomy.
    \textbf{(c)} Question-type distribution of OmniCoT-B, which closely matches OmniCoT-T and thus mitigates train--test distribution shift.}

    \label{fig:sta}
\end{figure*}

\noindent\textbf{Step 2: Chain-of-Thought Generation.}
Building upon the validated QA pairs, we implement a structured three-stage CoT generation pipeline to produce step-by-step spatial reasoning traces as shown in Fig.~\ref{fig:pipeline}.
Firstly, we utilize a specialized prompting framework that adapts to each question type's unique reasoning requirements, guaranteeing that each generated CoT must demonstrate 2-4 clear reasoning steps, use appropriate transition words, and maintain strict natural language compliance throughout.
Then, the summarizing process distills the core logical flow, preserving 
critical spatial information while eliminating redundancies.
Finally, we implement a rigorous scoring mechanism evaluating both the reasoning process and answer quality across six critical dimensions: reasoning format compliance, reasoning structure, scene information utilization, type-specific auto-penalties, answer format compliance, and answer correctness.
The QA pairs with the passed CoT will go through the final expert check process.
Considering the over-substantial time and cognitive effort required for the careful assessment of CoT data, we randomly sampled near 400 data samples for experts to conduct rigorous and detailed evaluations. Until the random evaluation result satisfies the accuracy requirements of 95\%, the final results will be accepted as the final version of OmniCoT. 
The specific random evaluation details are presented in the \textbf{\textit{Supplementary}} and statistics are shown in Fig.~\ref{fig:sta}

\section{Benchmark}

\subsection{Evaluation Design}
To assess spatial reasoning CoT on OmniCoT-B, we adopt a two-dimensional framework with six metrics capturing both general and spatial reasoning quality.

\noindent \textbf{General Reasoning Quality.} Following MME-CoT~\cite{jiang2025mme}, we employ three interpretable metrics to assess the general reasoning quality: Precision, Recall, and F1-score. 
Precision measures the faithfulness of each generated step by calculating the ratio of correct steps. 
Recall evaluates reasoning informativeness by quantifying the proportion of ground-truth solution steps that appear in the model's response. 
Moreover, F1-score provides a balanced overall assessment.
The evaluation process and prompts of judge LLMs for this part are strictly consistent with MME-CoT~\cite{jiang2025mme}.

\noindent  \textbf{Spatial Reasoning Quality.} We introduce three fine-grained metrics specifically for evaluating spatial reasoning quality in panoramas: Viewpoint Consistency~(VC), Spatial Evidence Sufficiency~(SES), and Reasoning Feasibility~(RF).
VC assesses model's ability to maintain coherent spatial perspectives throughout multi-hop reasoning. All viewpoint-related statements are extracted and scored individually against ground-truth geometry, with the final score being the average consistency.
SES measures whether the reasoning process adequately cites necessary spatial relationships. Key relationships required to answer the question are identified and checked for citation, with the citation rate serving as the final score.
RF evaluates the physical plausibility of execution steps (movement, turning, interaction) within the scene constraints. Each execution step is scored based on feasibility, with the average determining the final score. 

\begin{table*}[t]
    \centering
\caption{\textbf{Performance of SoTA MLLMs on OmniCoT-B.} Accuracy is evaluated on three spatial reasoning dimensions: \textbf{\textit{See}} (Multi-hop Viewpoint Transformation), \textbf{\textit{Locate}} (Inter-object Spatial Relations), and \textbf{\textit{Move}} (Embodied Action Simulation). CoT quality is measured by general metrics~\cite{jiang2025mme}: \textbf{\textit{Pre.}} (Precision), \textbf{\textit{Rec.}} (Recall), and \textbf{\textit{F1}}, along with panorama-specific metrics: \textbf{\textit{VC}} (Viewpoint Consistency), \textbf{\textit{SES}} (Spatial Evidence Sufficiency), and \textbf{\textit{RF}} (Reasoning Feasibility). Best and worst results are highlighted in \colorbox{green!20}{green} and \colorbox{red!20}{red}, respectively.}
    \label{tab:benchmarkr}
    \renewcommand\arraystretch{1} 
    \resizebox{\textwidth}{!}{%
    \begin{tabular}{l|ccc|c|cccccc}
    \toprule
    \multirow{2}{*}{\textbf{\textit{\large Model}}} & 
    \multicolumn{4}{c|}{\textbf{\textit{Accuracy Evaluation}}} & 
    \multicolumn{6}{c}{\textbf{\textit{Chain-of-Thought Evaluation}}} \\
    \cmidrule(l){2-5} \cmidrule(l){6-11}
    &  \textbf{\textit{See}} &  
     \textbf{\textit{Locate}} &
      \textbf{\textit{Move}} &
      \textbf{\textit{Overall}} &
    \textbf{ \textit{Pre.}} & 
    \textbf{ \textit{Rec.}} & 
    \textbf{ \textit{F1.}} & 
    \textbf{ \textit{VC}} & 
    \textbf{ \textit{SES}}  &
    \textbf{ \textit{RF}}
    \\
    \midrule
     \rowcolor{gray!20} \multicolumn{11}{c}{\textit{Open-Source MLLMs}} \\
    \midrule
               Qwen2.5-VL-3B-Instruct~\cite{bai2025qwen25} & 
    34.27 & 17.15 & 31.43 &   26.90 &  \cellcolor{red!20}26.17 & 18.23 & 21.49 & 36.85 & 23.36 & 39.82 \\
        Qwen2.5-VL-7B-Instruct~\cite{bai2025qwen25} & 
    33.77 & 15.53 & 24.33 &   23.53 & 34.60 & 20.52 & 25.76 & 43.85 & 28.05 & 41.40 \\
            Qwen2.5-VL-32B-Instruct~\cite{bai2025qwen25} & 
    47.84 &  21.93 & 31.99 &   32.39 & 42.05 & 27.39 & 33.17 & 57.33 & 31.97 & 49.18 \\
            Qwen2.5-VL-72B-Instruct~\cite{bai2025qwen25} & 
    52.03 & 23.69 & 35.05 &   35.26 & 46.57 & 28.53 & 35.39 & 49.51 & 36.30 & 54.80 \\
                Qwen3-VL-8B-Instruct~\cite{bai2025qwen3} & 
    41.52 & 18.73 & 26.59 &   27.56 & 61.22 & 29.48 & 39.80 & \cellcolor{green!20} 85.09 & \cellcolor{green!20} 79.05 & 63.49 \\
        LLaVA-OneVision-1.5-8B-Instruct~\cite{an2025llava} & 
    42.40 &  14.61 & 29.95 &  27.51 & 36.99 & 19.53 & 25.56 & 49.88 & 21.84 & 44.43 \\
        LLaVA-v1.5-7B~\cite{liu2024improved} & 
    15.00 &  9.82 & 16.80 & \cellcolor{red!20}  13.76 &  30.56 & 12.93 & 20.04 & 37.66 & \cellcolor{red!20} 14.72 & 43.76 \\
        LLaVA-v1.5-13B~\cite{liu2024improved} & 
    21.01 & 14.17 & 11.71 &   15.02 & 40.88 & 16.60 & 23.62 & 35.70  & 14.93 & 42.76 \\
         LLaVA-v1.6-Mistral-7B~\cite{liu2024llavanext} & 
    22.01 & 17.24 & \cellcolor{red!20} 11.40 &   16.31 & 35.18 & 17.64 & 23.49 & 34.80 & 22.56 & 37.64 \\
        LLaVA-v1.6-Vicuna-7B~\cite{liu2024llavanext} & 
    \cellcolor{red!20} 11.13 & 12.63 & 19.24 &   14.70 & 30.56 & \cellcolor{red!20}9.60 & \cellcolor{red!20}14.61 & \cellcolor{red!20} 28.52 & 25.41 & \cellcolor{red!20} 35.66 \\
        LLaVA-v1.6-Vicuna-13B~\cite{liu2024llavanext} & 
    16.32 & 15.00 & 22.33 &   18.07 & 30.40 & 11.31 & 16.48 & 31.36 & 16.41 & 37.12 \\
        InternVL3.5-14B~\cite{wang2025internvl3} & 
        41.77 & 20.35 & 31.64 &   30.11 & 40.54 & 27.99 & 33.11 & 46.94 & 36.18 & 49.61 \\
    \midrule
    \rowcolor{gray!20}   \multicolumn{11}{c}{\textit{Close-Source MLLMs}} \\
    \midrule
    ChatGPT-4o~\cite{hurst2024gpt} & 51.97 & 15.18 & 33.65 &   31.58 & 42.46 & 25.51 & 31.87 & 62.99 & 42.03 & 51.41 \\
    GPT-5~\cite{singh2025openai} & \cellcolor{green!20}57.29 & \cellcolor{green!20}24.40 & \cellcolor{green!20}42.71 &   \cellcolor{green!20}39.72 & 51.43 & \cellcolor{green!20}34.79 & \cellcolor{green!20}41.51 & 63.05 & 57.30 & 57.10 \\
    Doubao-1.5-Vision-Pro-32k~\cite{guo2025seed1} & 54.91 &  8.38 & 28.08 &   27.77 & 45.47 & 22.53 & 30.13 & 71.02 & 42.85 & 47.95 \\
    Doubao-1.8~\cite{seed2025seed1} & 57.09 &  23.87 & 41.14 &   38.90 & 42.56 & 29.92 & 35.13 & 66.29 & 49.94 & 55.79 \\
    Gemini3-Flash~\cite{team2023gemini} & 47.52 & 14.17 & 29.29 &   28.43 &  \cellcolor{green!20}79.30 & 22.46 & 35.01 & 70.22 & 46.54 &  \cellcolor{green!20}68.73 \\
    GLM-4.5v~\cite{hong2025glm} & 53.15 & 15.05 & 28.60 &   29.95 & 47.82 & 23.07 & 31.12 & 66.98 & 38.31 & 56.63 \\
    Step3~\cite{huang2026step3} & 48.40 & 16.71 & 33.69 &   31.23 & 58.59 & 22.53 & 32.55 & 63.71 & 36.96 & 59.49 \\
    Grok-4~\cite{xai2025grok4} & 54.40 & 19.30 & 37.04 &   34.99 & 48.28 & 29.21 & 36.40 & 55.36 & 53.88 & 48.65 \\
    Qwen3-Max~\cite{bai2025qwen3} & 
    52.03 & \cellcolor{red!20} 6.19 & 12.97 &   20.58 & 43.88 & 22.03 & 29.33 & 72.76 & 38.39 & 57.48 \\
    \bottomrule
    \end{tabular}%
    }
\end{table*}

\subsection{Experimental Settings \& Results}

In the evaluation process, we benchmark 12 open-source MLLMs (including Qwen2.5-VL~\cite{bai2025qwen25}, Qwen3-VL~\cite{bai2025qwen3}, LLaVA-OneVision~\cite{an2025llava}, LLaVA-v1.5~\cite{liu2024improved}, LLaVA-v1.6~\cite{liu2024llavanext}, and InternVL3.5~\cite{wang2025internvl3}) and 9 closed-source MLLMs (including ChatGPT-4o~\cite{hurst2024gpt}, GPT-5~\cite{singh2025openai}, Doubao-1.5~\cite{guo2025seed1}, Doubao-1.8~\cite{seed2025seed1}, Gemini3-Flash~\cite{team2023gemini}, GLM-4.5v~\cite{hong2025glm}, Step3~\cite{huang2026step3}, Qwen3-Max~\cite{bai2025qwen3}, and Grok-4~\cite{xai2025grok4}) on the OmniCoT-B. 
The input consists of an ERP panoramic image along with a corresponding question, supplemented with the coordinates and orientation indications (e.g., positive X-axis pointing east, positive Y-axis pointing north) of three randomly selected objects in the scene to establish a unified spatial reference frame. We refer to these coordinate-and-orientation hints as \textit{spatial anchors}. A spatial anchor is an object-level reference point with known position and orientation, used to align the visual panorama with the textual spatial description. \textit{Anchor density} denotes the number of such reference objects provided in the prompt.  We use 3 anchors by default because they provide a lightweight global frame while avoiding exposure of the full scene layout, and random selection prevents the evaluation from favoring object-specific priors. To disentangle panoramic reasoning from ERP distortion, we further evaluate multi-view perspective inputs in the Supplementary. DeepSeekv3.2~\cite{liu2025deepseek} is selected as the judge model for CoT scoring. 

As summarized in Table~\ref{tab:benchmarkr}, our evaluation reveals divergent performance patterns across models, where closed-source titans like GPT-5 and Doubao-1.8 establish a stronghold in the \textbf{\textit{Move}} dimension (over 41\% accuracy), demonstrating superior stability in ego-perspective updates. While large-scale models such as Qwen2.5-VL-72B excel in geometric projection (\textbf{\textit{See}}), a universal bottleneck persists in the \textbf{\textit{Locate}} dimension, where no model surpasses the 25\% threshold, highlighting the difficulty of multi-hop topological reasoning. Interestingly, fine-grained CoT metrics expose a precision-informativeness trade-off: high-scale models like Gemini3-Flash achieve near-perfect Precision (79.30\%) but suffer from sparse reasoning traces (low Recall), whereas smaller models like Qwen3-VL-8B exhibit higher fidelity in spatial anchoring (VC and SES). These results underscore that while scaling enhances raw perception, it does not solve the challenge of consistent environmental grounding in complex panoramic scenes.

\section{Extensive Results}

\subsection{CoT \textit{v.s.} Direct Answer}

\subsubsection{Divergent Impact of CoT on MLLMs: }
\begin{wraptable}{r}{0.48\textwidth}
    \vspace{-3.5em}
    \centering
    \caption{\textbf{Impact of CoT on Overall Accuracy.} We compare \textbf{w/o CoT} (direct answering) with \textbf{w/ CoT} (thinking). $\Delta$ denotes absolute accuracy shift (\textbf{w/CoT} \textit{v.s.} \textbf{w/o CoT}). \textcolor{red!60}{RED} and \textcolor{blue!60}{BLUE} indicate gains and drops.}
    \label{tab:overall_shift}
    \renewcommand\arraystretch{1.1}
    \setlength{\tabcolsep}{10pt}
    \resizebox{0.5\columnwidth}{!}{%
    \begin{tabular}{lccc}
    \toprule
    \textbf{Model} & \textbf{w/o CoT} & \textbf{w/ CoT} & \textbf{$\Delta$} \\
    \midrule
    \rowcolor{gray!15} \multicolumn{4}{c}{\textit{Open-Source MLLMs}} \\
    Qwen3-VL-8B-Instruct~\cite{bai2025qwen3}        & 25.39 & \textbf{27.56} & \textcolor{red!60}{+2.17} \\
    Qwen2.5-VL-3B-Instruct~\cite{bai2025qwen25}      & 21.20 & \textbf{26.90} & \textcolor{red!60}{+5.70} \\
    Qwen2.5-VL-7B-Instruct~\cite{bai2025qwen25}      & 23.01 & \textbf{23.53} & \textcolor{red!60}{+0.52} \\
    Qwen2.5-VL-72B-Instruct~\cite{bai2025qwen25}     & 26.51 & \textbf{35.26} & \textcolor{red!60}{+8.75} \\
    InternVL3.5-14B~\cite{wang2025internvl3}             & 23.79 & \textbf{30.11} & \textcolor{red!60}{+6.32} \\
    LLaVA-v1.5-7B~\cite{liu2024improved}               & 4.91  & \textbf{13.77} & \textcolor{red!60}{+8.86} \\
    LLaVA-v1.5-13B~\cite{liu2024improved}              & 17.38 & \textbf{15.03} & \textcolor{blue!60}{-2.35} \\
    LLaVA-v1.6-Mistral-7B~\cite{liu2024llavanext}       & 16.70 & \textbf{16.31} & \textcolor{blue!60}{-0.39} \\
    LLaVA-v1.6-Vicuna-7B~\cite{liu2024llavanext}        & 9.77  & \textbf{14.70} & \textcolor{red!60}{+4.93} \\
    LLaVA-v1.6-Vicuna-13B~\cite{liu2024llavanext}       & 18.52 & \textbf{18.07} & \textcolor{blue!60}{-0.45} \\ 
    \midrule
    \rowcolor{gray!15} \multicolumn{4}{c}{\textit{Close-Source / API MLLMs}} \\
    Qwen3-Max~\cite{bai2025qwen3}                   & 25.39 & \textbf{20.58} & \textcolor{blue!60}{-4.81} \\
    GPT-5~\cite{singh2025openai}                       & 42.66 & \textbf{39.72} & \textcolor{blue!60}{-2.94} \\
    Doubao-1.5-Vision-Pro-32k~\cite{guo2025seed1}   & 30.96 & \textbf{27.76} & \textcolor{blue!60}{-3.20} \\
    Gemini3-Flash~\cite{team2023gemini}              & 39.97 & \textbf{28.44} & \textcolor{blue!60}{-11.53} \\
    \bottomrule
    \end{tabular}}
    \vspace{-2em}
\end{wraptable}
To further investigate whether Chain-of-Thought (CoT) genuinely enhances spatial reasoning in panoramas, we conduct a controlled comparative experiment.  For each model, we evaluate 
under two distinct response modes: \ding{182} CoT, \ding{183} Direct Answer, with the results shown in Table~\ref{tab:overall_shift}.
For open-source models, CoT acts as an effective panoramic spatial navigator. By externalizing multi-hop logic, open-source models' panoramic reasoning ability is improved. 
Conversely, forcing CoT on closed-source models (\textit{e.g.}, Gemini3-Flash) actively degrades their accuracy. 
Forcing them to write out a step-by-step reasoning path is proven to be unnecessary and tends to hurt their performance.
This divergent impact suggests that \textbf{\textit{the verbose generation of long reasoning steps inherently carries a risk of cumulative hallucination, which currently outweighs the logical benefits of CoT for top-tier closed-source models.}}

\subsubsection{Coupling between Reasoning Quality and Accuracy:}
We further compare the trends of four CoT quality metrics (F1., VC, SES, RF) and the overall accuracy, which are visualized in Fig.~\ref{fig:teaser}. 
Overall, accuracy exhibits a positive directional synergy with spatial reasoning metrics, confirming the intuition that better reasoning generally yields better predictions. However, we also observe a paradoxical ``Reasoning-Answer Mismatch'' in certain outliers. A striking example is Qwen3-VL-8B-Instruct: it achieves relatively high RF, VC, and SES, but fails to deliver a commensurate final accuracy. This exposes a phenomenon we term ``linguistic blind-navigation''. 
The model is highly articulate, capable of writing out a physically flawless, step-by-step navigation plan (\textit{e.g.}, ``turn 90 degrees right, walk past the sofa''). 
Yet, this textual logic is disconnected from the actual visual geometry.
Because panoramic images suffer from severe spherical distortions, the model's pure language skills create a ``persuasive illusion'' of reasoning, but it ultimately fails to map those perfectly written textual steps back to the distorted pixels to locate the correct target.
Ultimately, the ``linguistic blind-navigation'' phenomenon reveals a critical flaw in some MLLMs: \textbf{\textit{their internal reasoning process is often driven by text priors rather than true cross-modal spatial perception.}}

\begin{wrapfigure}{r}{0.6\textwidth}
  \vspace{-2.2em}
   \includegraphics[width=0.6\textwidth]{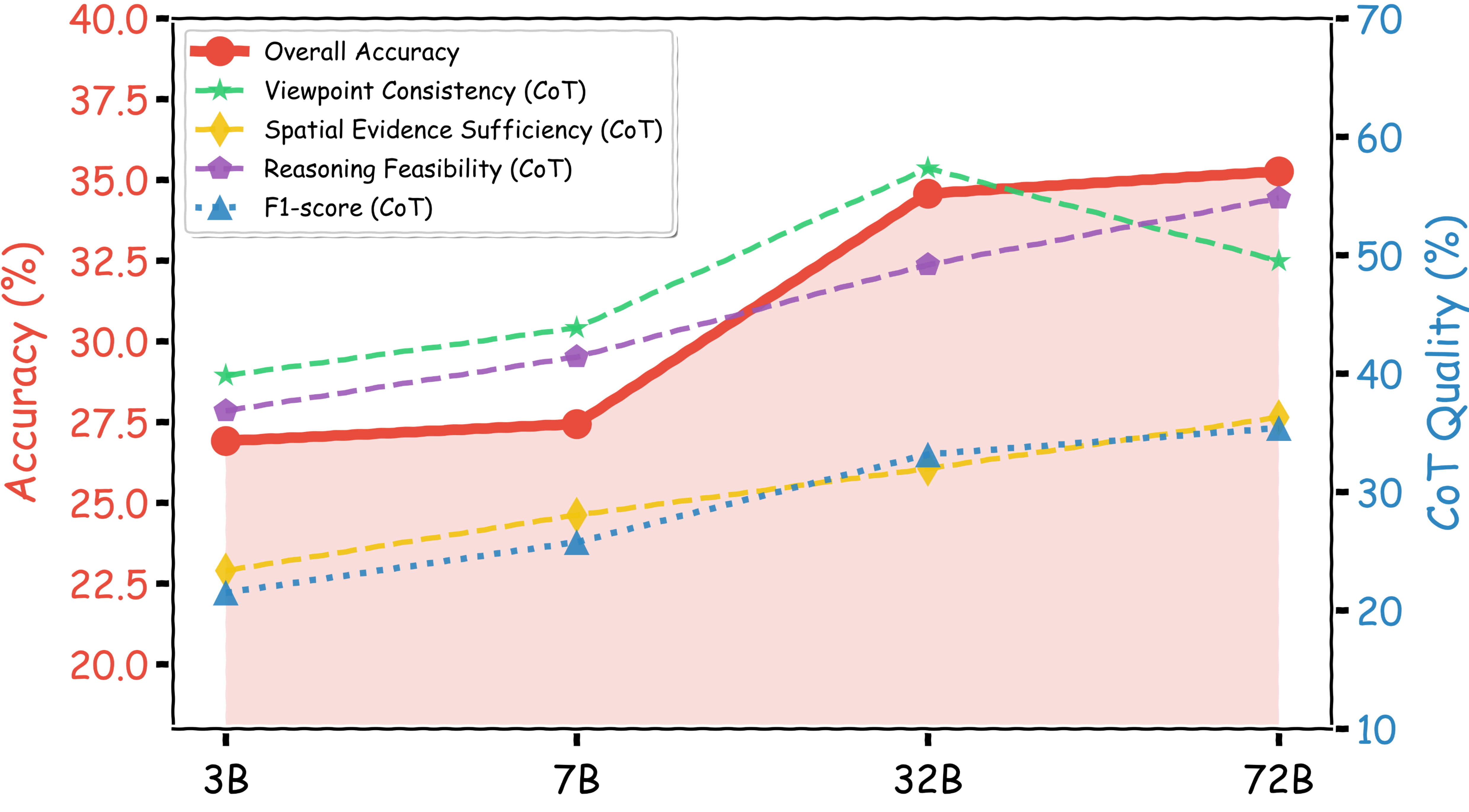}
   \caption{\textbf{Scaling Analysis of Panoramic Spatial Reasoning.} We evaluate the Qwen2.5-VL model family across varying parameter scales (3B to 72B).}
   \label{fig:scaling}
\vspace{-2em}
\end{wrapfigure}
\subsection{What Makes Better Reasoning in Panoramas}
\subsubsection{Scaling Laws in Panoramic Spatial Reasoning:}
We investigate the scaling behavior of panoramic spatial reasoning using the Qwen2.5-VL family (3B, 7B, 32B, and 72B). As shown in Fig.~\ref{fig:scaling}, we observe a clear scaling dividend in overall accuracy, which improves steadily from 26.90\% (3B) to 35.26\% (72B). 
However, this scaling trajectory exhibits a notable marginal effect. The performance leap from 7B to 32B is substantial, yet the subsequent push to 72B yields only minor incremental gains. 
Similarly, the scaling of reasoning quality metrics follows a non-linear pattern, peaking early and plateauing at larger scales. 
This indicates that \textbf{\textit{while increasing the parameter count successfully broadens the MLLM's ability for handling panoramas, simple scaling eventually hits a bottleneck.}} 
To further break through the performance ceiling in panoramas, relying solely on larger models is insufficient; explicitly optimizing the model's spatial reasoning processes (\textit{e.g.}, Post-training) becomes a more promising path.

\begin{wraptable}{r}{0.5\textwidth}
\vspace{-3.4em}
        \caption{\textbf{Performance comparison} of selected models with different numbers of initial reference points (0, 3, 10).}
    \centering
    \renewcommand\arraystretch{1} 
    \resizebox{0.5\textwidth}{!}{%
    \begin{tabular}{c|ccc|c|cccccc}
    \toprule
    \multirow{2}{*}{\textbf{\textit{\large Points}}} & 
    \multicolumn{4}{c|}{\textbf{\textit{Accuracy Evaluation}}} & 
    \multicolumn{6}{c}{\textbf{\textit{Chain-of-Thought Evaluation}}} \\
    \cmidrule(l){2-5} \cmidrule(l){6-11}
    &  \textbf{\textit{See}} &  
     \textbf{\textit{Locate}} &
      \textbf{\textit{Move}} &
      \textbf{\textit{Overall}} &
    \textbf{ \textit{Pre.}} & 
    \textbf{ \textit{Rec.}} & 
    \textbf{ \textit{F1.}} & 
    \textbf{ \textit{VC}} & 
    \textbf{ \textit{SES}}  &
    \textbf{ \textit{RF}}
    \\
    \midrule
    \rowcolor{gray!15} \multicolumn{11}{c}{\textit{Qwen2.5-VL-7B-Instruct~\cite{bai2025qwen25}}} \\
     \midrule
     0 & 30.70 & 22.37 & 25.77 & 25.79 & 40.25 & 22.2 & 28.96 & 38.95 & 35.13 & 42.18 \\
     3 & 33.77 & 15.53 & 24.33 & 23.53 & 34.60 & 20.52 & 25.76 & 43.85 & 28.05 & 41.40 \\
     10 & 35.33 & 22.07 & 33.17 & 29.63 & 48.89 & 29.79 & 37.02 & 48.79 & 37.68 & 48.79 \\
     \midrule
    \rowcolor{gray!15} \multicolumn{11}{c}{\textit{LLaVA-v1.6-Vicuna-7B~\cite{liu2024llavanext}}} \\
     \midrule
     0 & 18.26 & 14.87 & 20.98 & 18.02 & 40.49 & 15.25 & 22.16 & 33.18 & 15.03 & 38.65 \\
     3 & 11.13 & 12.63 & 19.24 & 14.70 & 30.56 & 9.60 & 14.61 & 28.52 & 25.41 & 35.66 \\
     10 & 12.75 & 13.99 & 18.80 & 15.46 & 32.76 & 14.88 & 20.46 & 30.06 & 16.06 & 37.97\\
     \midrule
    \rowcolor{gray!15} \multicolumn{11}{c}{\textit{InternVL3.5-14B~\cite{wang2025internvl3}}} \\
     \midrule
     0 & 36.64 & 19.26 & 31.60 & 28.35 & 41.71 & 27.67 & 33.27 & 40.72 & 35.14 & 45.66 \\
      3 & 41.77 & 20.35 & 31.64 & 30.10 & 40.54 & 27.99 & 33.11 & 46.94 & 36.18 & 49.61 \\
      10 & 44.34 & 25.23 & 36.13 & 34.23 & 44.43 & 29.58 & 35.52 & 54.59 & 43.40 & 55.52 \\
     \midrule
    \rowcolor{gray!15} \multicolumn{11}{c}{\textit{GPT-5~\cite{singh2025openai}}} \\
     \midrule
     0 & 56.47 & 28.21 & 41.84 & 40.60 & 50.58 & 35.48 & 41.71 & 54.88 & 50.67 & 54.75 \\
     3 & 57.29 & 24.40 & 42.71 & 39.72 & 51.43 & 34.79 & 41.51 & 63.05 & 57.30 & 57.10 \\
     10 & 58.66 & 29.74 & 48.01 & 44.03 & 64.80 & 42.62 & 51.42 & 77.98 & 76.96 & 71.18 \\
    \midrule
    \rowcolor{gray!15} \multicolumn{11}{c}{\textit{Gemini3-Flash~\cite{team2023gemini}    }} \\
     \midrule
     0 & 47.28 & 24.97 & 32.13 & 33.41 & 69.18 & 24.13 & 35.78 & 53.26 & 36.77 & 59.01 \\
      3 & 47.52 & 14.17 & 29.29 & 28.43 & 79.30 & 22.46 & 35.01 & 70.22 & 46.54 & 68.73 \\
      10 & 53.28 & 31.11 & 46.36 & 42.52 & 81.49 & 30.64 & 44.53 & 77.19 & 71.42 & 79.87 \\
    \bottomrule
    \end{tabular}%
    }
        \vspace{-2.3em}
    \label{tab:Points_comparison}
\end{wraptable}
\subsubsection{The Impact of Spatial Anchor Density:}
\begin{figure}

    \centering
    \includegraphics[width=0.9\linewidth]{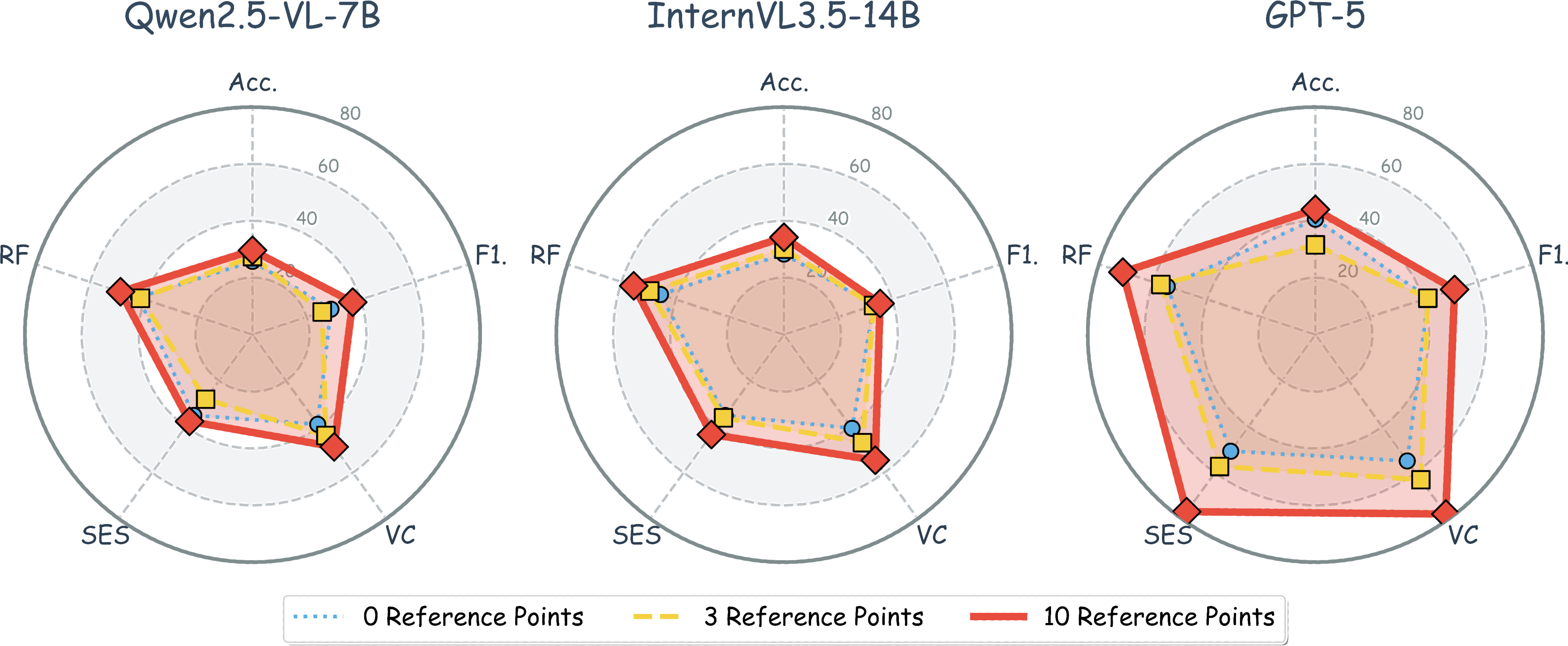}
    \caption{\textbf{Capability Expansion via Spatial Anchors.} Radar charts illustrate model performance across Accuracy and four CoT metrics with varying numbers of reference points (0, 3, 10).}
    \label{fig:anch}

\end{figure}

To understand the role of the environmental grounding anchor, we evaluate model performance under varying densities of reference points  (0, 3, and 10). As summarized in Table~\ref{tab:Points_comparison} and Fig.~\ref{fig:anch}, 
increasing the number of spatial anchors yields a ``precision dividend'' across most MLLMs. These reference points serve as global coordinate ``pins'' that mitigate spatial drift. For instance, InternVL3.5-14B achieves a substantial 5.88\% accuracy boost when anchors are increased from 0 to 10. Interestingly, we observe that providing even a few anchors significantly activates the latent spatial reasoning capabilities of top-tier models like GPT-5 at the 10-point setting.
 This confirms that \textbf{\textit{explicit spatial anchors act as crucial global coordinates that effectively mitigate spatial drift and activate the latent reasoning capabilities of MLLMs.}}

\subsection{Small-Scale Real-world Test}
To evaluate MLLMs' panoramic spatial reasoning capability in real-world environments and quantify the gap between synthetic and real-world settings, we construct \textit{\textbf{OmniCoT-Real}}. Specifically, we collect 200 real-world panoramas 
with  Insta-X5, spanning 13 diverse indoor scene categories, including living rooms, kitchens, bathrooms, corridors, classrooms, and gyms, among others.
All QA pairs are manually
annotated by three domain experts over approximately 320 hours of collaborative effort, yielding 1,073 QA pairs in total. Compared to OmniCoT-B, 
grounded in simulated 3D scenes with precise coordinates, OmniCoT-Real introduces authentic real-world challenges, including lens distortion, uneven lighting, and object occlusion, making it a more demanding testbed for evaluating generalization.

\begin{figure}[!t]

    \centering
    \includegraphics[width=\linewidth]{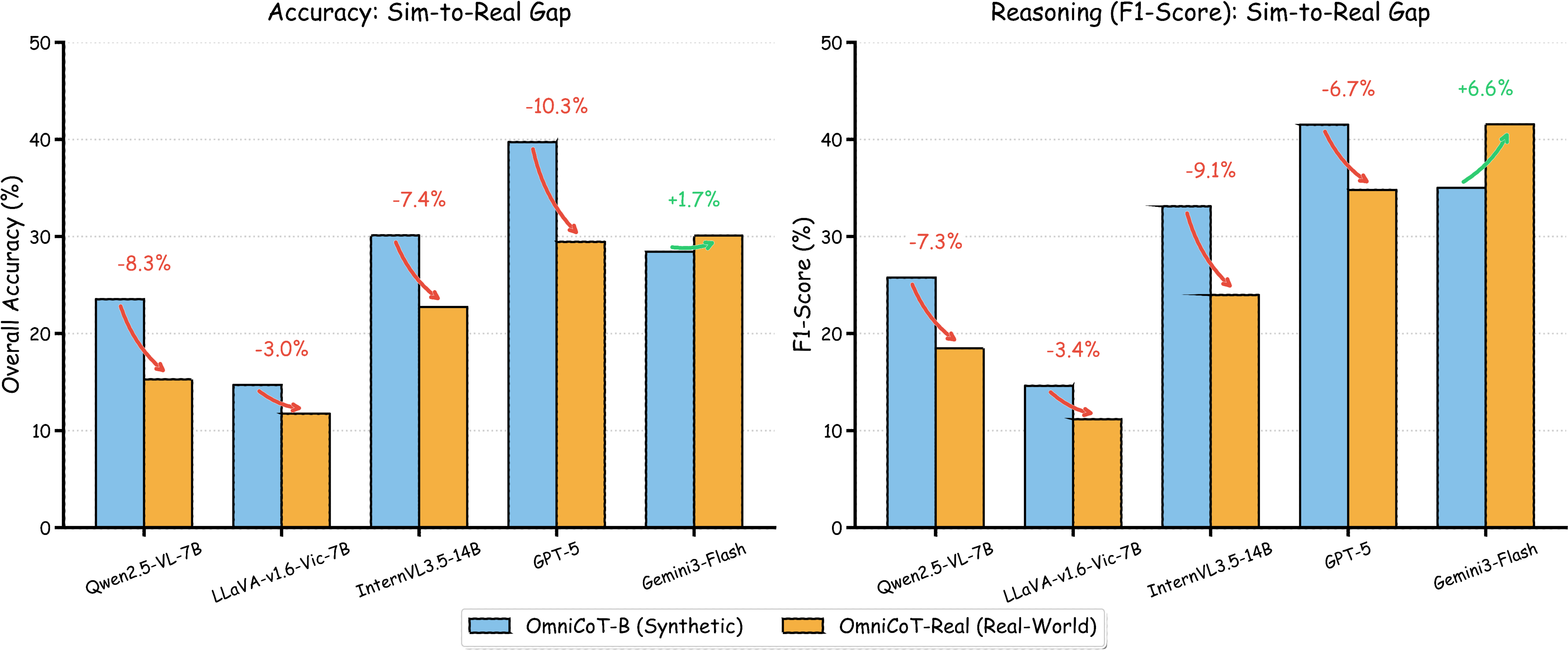}
    \caption{\textbf{Sim-to-Real Performance Gap.} Performance comparison between the synthetic benchmark (OmniCoT-B) and the real-world evaluation set (OmniCoT-Real). We evaluate both Overall Accuracy (left) and Spatial Reasoning Quality via F1-Score (right). 
    A clear performance degradation is observed when transitioning from simulated scenes to real-world continuous environments, highlighting the persistent challenge of real-world generalization for MLLMs in panoramic environments.}
    \label{fig:gap}
\end{figure}

\begin{wraptable}{r}{0.65\textwidth}
\vspace{-3.3em}
    \centering
        \caption{\textbf{Evaluation on OmniCoT-Real,} showing performance comparison on real-world panoramas.}
    \renewcommand\arraystretch{1.1} 
    \resizebox{0.65\textwidth}{!}{%
    \begin{tabular}{l|ccc|c|ccc}
    \toprule
    \multirow{2}{*}{\textbf{\textit{\large Model}}} & 
    \multicolumn{4}{c|}{\textbf{\textit{Accuracy}}} & 
    \multicolumn{3}{c}{\textbf{\textit{CoT}}} \\
    \cmidrule(l){2-5} \cmidrule(l){6-8}
    &  \textbf{\textit{See}} &  
     \textbf{\textit{Locate}} &
      \textbf{\textit{Move}} &
      \textbf{\textit{Overall}} &
    \textbf{ \textit{Pre.}} & 
    \textbf{ \textit{Rec.}} & 
    \textbf{ \textit{F1.}}

    \\
    \midrule
        Qwen2.5-VL-7B-Instruct~\cite{bai2025qwen25} &
        18.48 & 5.74 & 20.59 & 15.28 &
        35.20 & 12.53 & 18.48  \\
        LLaVA-v1.6-Vicuna-7B~\cite{liu2024llavanext} &
        11.74 & 15.51 & 04.23 & 11.74 &
        25.00 & 7.19 & 11.17  \\
        InternVL3.5-14B~\cite{wang2025internvl3} &
        25.54 & 8.70 & 32.35 & 22.73 &
        40.52 & 17.01 & 23.97  \\
        GPT-5~\cite{singh2025openai} &
        30.71 & 17.22 & 39.04 & 29.45 &
        55.69 & 25.31 & 34.80  \\
        Gemini3-Flash~\cite{team2023gemini}     &
        33.15 & 15.11 & 40.37 & 30.10 &
        58.61 & 32.19 & 41.56  \\
    \bottomrule
    \end{tabular}%
    }
    \vspace{-2em}
    \label{tab:benchmarkrreal}
\end{wraptable}
As illustrated in Table~\ref{tab:benchmarkrreal} and Fig.~\ref{fig:gap}, transitioning from synthetic environments to real-world panoramas exposes a substantial sim-to-real performance gap. 
The majority of evaluated MLLMs suffer significant declines in both overall accuracy and spatial reasoning quality (F1-score) on OmniCoT-Real. 
Specifically, all models consistently score lowest on the \textit{Locate} dimension, suggesting that fine-grained inter-object relationship reasoning is particularly vulnerable to real-world visual complexity. 
Furthermore, open-source models exhibit notably low Recall compared to their closed-source counterparts, reflecting a failure to ground sufficient spatial evidence when confronted with real-world panoramas. 
Ultimately, the stark contrast between synthetic and real-world performance underscores that simulated benchmarks alone are insufficient. 
This serves as a compelling call to action, emphasizing \textit{\textbf{the indispensable role of large-scale, real-world panoramic data in developing and evaluating truly robust, generalizable MLLMs.}}

\subsection{OmniCoT-R1}
\label{sec:omnicot_r1}

\begin{wraptable}{r}{0.5\textwidth}
\vspace{-3.5em}
\centering
\caption{\textbf{Implementation details of OmniCoT-R1}, including key hyperparameters for SFT and GRPO.}
\renewcommand\arraystretch{1.1}
\setlength{\tabcolsep}{6pt}
\resizebox{0.5\textwidth}{!}{%
\begin{tabular}{l|l}
\toprule
\textbf{Item} & \textbf{Setting} \\
\midrule
Base model & Qwen2.5-VL-7B-Instruct~\cite{bai2025qwen25} \\
Trainable modules (SFT) & Language model only, vision+proj frozen \\
Trainable modules (GRPO) & Full model fine-tuning \\
Precision & BF16 \\
Max sequence length & 4096 \\
Max completion length & 2048 \\
\midrule
SFT data & OmniCoT-T \\
SFT LR schedule & Cosine, warmup 0.03 \\
\midrule
GRPO data & OmniCoT-T \\
GRPO generations & 8 \\
Rewards & format / accuracy / repetition \\
Reward weights & 0.1 / 1.0 / 0.2 \\
GRPO learning rate & 1e-6 \\
\bottomrule
\end{tabular}}
\vspace{-2em}
\label{tab:r1_params}
\end{wraptable}

\subsubsection{Effectiveness of SFT and GRPO:}
\begin{figure}
    \centering
    \includegraphics[width=\linewidth]{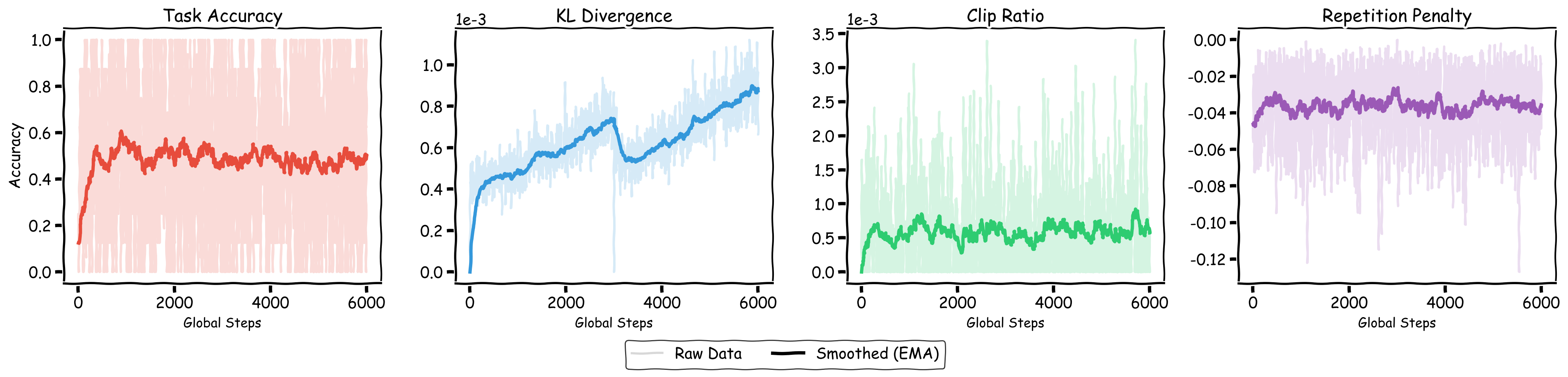}
    \caption{\textbf{Training Dynamics of OmniCoT-R1 (GRPO Stage).} We track the evolution of four key metrics over the initial 6,000 optimization steps. Raw step-wise data is shown as a faded background, overlaid with an Exponential Moving Average (EMA) trendline for clarity.}
    \label{fig:grpo}
\end{figure}

Building on our proposed OmniCoT-T, we develop \textit{\textbf{OmniCoT-R1}} by post-training Qwen2.5-VL with a two-stage strategy including SFT and GRPO.
During the SFT stage, we explicitly instruct the model to follow a structured panoramic reasoning protocol, thereby yielding an explicit \texttt{<think>...<answer>} format where each reasoning step is strictly grounded in visual evidence from the panorama.
Initialized from this SFT checkpoint, we subsequently employ GRPO to further refine the model's long-horizon spatial reasoning capabilities.
The reward combines format compliance to enforce the structured output schema, task accuracy computed by our geometry-grounded executor for OmniCoT, and anti-degeneration regularization to discourage verbose or repetitive chains. The details of the implementation are summarized in
Table~\ref{tab:r1_params} and training dynamics of OmniCoT-R1 is shown in Fig.~\ref{fig:grpo}.

\begin{table*}[t]
    \centering
        \caption{\textbf{Results of OmniCoT-R1.} Performance evaluation on OmniCoT-B comparing the baseline Qwen2.5-VL-7B-Instruct against our models trained via Supervised Fine-Tuning (SFT) and subsequent Group Relative Policy Optimization (GRPO). Both accuracy and spatial reasoning metrics show consistent gains through the two-stage post-training pipeline.}\renewcommand\arraystretch{1.1} 
    \resizebox{\textwidth}{!}{%
    \begin{tabular}{l|ccc|c|cccccc}
    \toprule
    \multirow{2}{*}{\textbf{\textit{\large Model}}} & 
    \multicolumn{4}{c|}{\textbf{\textit{Accuracy Evaluation}}} & 
    \multicolumn{6}{c}{\textbf{\textit{Chain-of-Thought Evaluation}}} \\
    \cmidrule(l){2-5} \cmidrule(l){6-11}
    &  \textbf{\textit{See}} &  
     \textbf{\textit{Locate}} &
      \textbf{\textit{Move}} &
      \textbf{\textit{Overall}} &
    \textbf{ \textit{Pre.}} & 
    \textbf{ \textit{Rec.}} & 
    \textbf{ \textit{F1.}} & 
    \textbf{ \textit{VC}} & 
    \textbf{ \textit{SES}}  &
    \textbf{ \textit{RF}}
    \\
    \midrule Qwen2.5-VL-7B-Instruct~\cite{bai2025qwen25}& 
    33.77 & 15.53 & 24.33 &   23.53 & 34.60 & 20.52 & 25.76 & 43.85 & 28.05 & 41.40 \\
    \rowcolor{yellow!15} OmniCoT-R1 (SFT)& 
     59.41 & 39.44 & 52.06 & 49.31 & 42.89 & 34.22 & 38.07  & 46.25 & 57.87 & 53.93 \\
    \rowcolor{orange!15} OmniCoT-R1 (SFT+GRPO) & 
     67.97 & 51.82 & 61.34 & 59.54 & 55.81 & 47.19 & 51.14 & 59.20 & 66.64 & 63.21 \\
    \bottomrule
    \end{tabular}%
    }

    \label{tab:benchmarkr1}
    \end{table*}

\subsubsection{Training Stability:}
Beyond accuracy, training dynamics exhibit consistently healthy behavior. The ThinkAnswerFormat reward reaches a perfect score from the very first GRPO step and remains saturated throughout training, confirming that SFT reliably establishes the structural prerequisite for reward-driven optimization. The Kullback-Leibler (KL) divergence between the updated policy and the reference model increases only marginally and stabilizes within a narrow bound, indicating that the model acquires enhanced spatial reasoning capabilities without undergoing catastrophic forgetting or deviating significantly from the base policy. Furthermore, the surrogate objective clipping ratios consistently remain near zero, ensuring that policy updates are strictly confined within the GRPO trust region. The Repetition Penalty signal exhibits steady improvement over the training horizon, demonstrating a reduction in output degeneration and promotes the generation of more diverse, non-repetitive reasoning traces.

Collectively, these results provide concrete evidence for the effectiveness of OmniCoT-T as a post-training resource: it drives consistent reward improvements, maintains stable optimization dynamics, and reduces degeneration over long-horizon updates. Overall, OmniCoT-R1 serves as a successful baseline and pipeline for the panoramic spatial reasoning task, providing the community with a validated data foundation and a reproducible Reinforce-Learning paradigm for future panoramic spatial reasoning research.

\section{Conclusion}

In this paper, we identify a fundamental limitation in existing panoramic spatial reasoning benchmarks: the tendency to focus on simplistic, few-hops questions that fail to exploit the native advantage of panoramas.
To address this, we introduce \textit{\textbf{OmniCoT}}, a suite encompassing the \textit{\textbf{OmniCoT-B}} benchmark, the \textit{\textbf{OmniCoT-Real}} subset, which enables the assessment of sim-to-real generalization, and the \textit{\textbf{OmniCoT-T}} training dataset, explicitly designed to demand global and multi-hop reasoning across three progressive dimensions: \textbf{\textit{See}}, \textbf{\textit{Locate}}, and \textbf{\textit{Move}}.
Our experiments reveal that CoT yields divergent effects across model families, and that panoramic reasoning quality is jointly shaped by model scale and explicit spatial grounding, with anchor density proving a particularly effective lever for activating latent reasoning capabilities.
Moving beyond synthetic scenes, the substantial sim-to-real gap observed on OmniCoT-Real further highlights the necessity of real-world panoramic data and more faithful evaluation protocols for robust generalization.
In conclusion, OmniCoT recalibrates the difficulty of panoramic reasoning benchmarks and provides a unified testbed for diagnosing both final-answer accuracy and intermediate reasoning quality. We hope this benchmark will facilitate principled comparisons across MLLMs, inspire stronger panorama-specific grounding and planning strategies. We will release the dataset, evaluation toolkit, and baseline models to support reproducible research and encourage future extensions to broader interaction settings. 

\section*{Acknowledgements}
This work was supported by the EU Horizon projects ELIAS (No. 101120237) and ELLIOT (No. 101214398) and by the FIS project GUIDANCE (No. FIS2023-03251).

\clearpage

%
%
\bibliographystyle{splncs04}
\bibliography{main}

\appendix
\section{Ethics Statement}
The synthetic part of our dataset is derived from DeepPanoContext~\cite{zhang2021deeppanocontext} and ReplicaPano~\cite{dong2024panocontext}, both of which are publicly available for academic research with the MIT license. 
We strictly adhere to the MIT License governing these source datasets. 
Our processing pipeline respects the original authors' intellectual property, and the redistributed data maintains the permissions granted by the original licenses.
Meanwhile, in the real-world part of our dataset, a thorough inspection was conducted to ensure that no personal privacy information is leaked.

\section{More Details}

\subsection{Dataset Construction}

OmniCoT is constructed through a hybrid pipeline that combines structured scene serialization, LLM-based candidate generation, multi-stage automatic filtering, and expert verification, targeting the emerging problem of panoramic and omnidirectional spatial reasoning~\cite{xiao2012recognizing,zheng2025panorama,dongfang2025multimodal,zhang2025towards,yang2025odi}.
The overall objective of this pipeline is to produce panoramic QA pairs and corresponding Chain-of-Thought traces that are simultaneously natural in language, unambiguous in answer space, and sufficiently challenging in spatial reasoning depth.
As described in the main paper, our construction process follows two sequential stages: \textit{\textbf{(1) Question-Answer generation and validation}}, and \textit{\textbf{(2) CoT generation and quality control}}.
In both stages, we deliberately adopt a \textit{\textbf{generate--score--filter--verify}} paradigm to balance scale and quality.

\noindent \textbf{Structured Scene Serialization.}
We begin with reliable 3D scene annotations and convert them into structured natural-language scene descriptions, building on panoramic scene understanding resources developed in prior work~\cite{zhang2021deeppanocontext,dong2024panocontext}.
This representation serves two purposes.
First, it preserves the global geometric and semantic information required for panoramic spatial reasoning, including object identities, relative positions, nearby relations, and structural elements such as walls and windows.
Second, it prevents downstream generation from overfitting to raw coordinates or simulator-specific identifiers, making the final questions closer to realistic language use.
In particular, the subsequent prompts explicitly prohibit coordinate-based references and require all descriptions to be expressed in natural language, such as ``the chair near the floor lamp'' or ``the piano in the southwest corner,'' rather than forms like ``piano$(-6.9,1.2)$''.

\noindent \textbf{Question Generation.}
Given the structured scene description, we prompt an LLM to generate several candidate questions for each of the six predefined OmniCoT types.
To extend recent efforts on visual and multimodal spatial reasoning toward omnidirectional panoramic settings~\cite{liu2023visual,cai2025spatialbot,daxberger2025mm,stogiannidis2025mind,yu2025far,zheng2025multimodal}, these six types span the three progressive reasoning dimensions of See, Locate, and Move.
The question-generation prompt is designed around three explicit goals: natural-language grounding, answer uniqueness, and moderate reasoning complexity.
More specifically, the prompt enforces that all spatial references must be expressed through natural language rather than coordinates or internal IDs.
It also requires each question to admit exactly one unambiguous answer by using qualifiers such as \textbf{\textit{nearest}}, \textbf{\textit{first}}, \textbf{\textit{second closest}}, or \textbf{\textit{directly ahead}}, together with sufficient relational context.
Finally, the prompt constrains the target reasoning depth to approximately two to four steps, thereby avoiding both trivial single-step queries and excessively long reasoning chains.
For question types involving visibility, movement, or viewpoint change, the prompt additionally incorporates structural constraints such as wall occlusion, window transparency, cardinal directions, movement distance, and sequential rotation order.
These type-specific templates substantially improve the consistency and diversity of candidate questions while keeping them aligned with the intended reasoning taxonomy.

\noindent \textbf{Question Scoring and Filtering.}
The generated candidate questions are then evaluated by reasoning-oriented LLM judges before entering the final QA pool.
Following the main paper, we employ two independent judges and score each candidate from six dimensions:
\textit{\textbf{(1) format compliance}},
\textit{\textbf{(2) object uniqueness}},
\textit{\textbf{(3) logical consistency}},
\textit{\textbf{(4) reasoning complexity}},
\textit{\textbf{(5) answerability}}, and
\textit{\textbf{(6) type-specific auto-penalties}}.
This scoring stage is intentionally conservative and functions as a high-precision filter rather than a creativity-oriented evaluator.
Each dimension serves a distinct role.
\textbf{\textit{Format compliance}} checks whether the question obeys the natural-language-only requirement and avoids any coordinate leakage.
\textit{\textbf{Object uniqueness}} verifies that every referenced object can be uniquely resolved in the scene, especially in layouts with repeated furniture categories.
\textit{\textbf{Logical consistency}} checks whether the question is compatible with the scene description, including object existence, direction relations, and physical feasibility.
\textit{\textbf{Reasoning complexity}} ensures that the question requires genuine multi-step reasoning rather than a direct lookup, but remains within a verifiable difficulty range.
\textit{\textbf{Answerability}} verifies that the scene description contains sufficient information to derive exactly one correct answer.
Finally, \textbf{\textit{type-specific auto-penalties}} reject malformed questions that violate the structural requirements of their corresponding type, such as missing distance qualifiers, unspecified movement directions, non-specific starting positions, or invalid angle constraints.
Only candidates that satisfy the criteria are retained.
This stage is crucial because panoramic scenes often contain multiple semantically similar objects, and even a small amount of ambiguity can significantly reduce the reliability of both automatic supervision and downstream evaluation.
After automatic filtering, experts perform a rapid manual review of the retained QA pairs to remove residual format issues or obvious content deviations.

\noindent \textbf{CoT Generation.}
Building upon the validated QA pairs, we generate step-by-step CoT traces through a structured three-stage pipeline.
First, we use a type-aware prompting framework that adapts the reasoning instructions to the corresponding question family.
Instead of using a single generic CoT prompt, we provide dedicated guidance for viewpoint transformation, multi-hop relational reasoning, movement simulation, and visibility analysis.
This type-aware design improves reasoning faithfulness by explicitly specifying the required intermediate operations, such as tracking sequential rotations, resolving intermediate relational anchors, simulating translational movement, or checking visibility under wall and window constraints.
A key design principle in CoT generation is again strict natural-language grounding.
The prompt forbids any use of coordinates, position vectors, or technical object IDs, and requires the model to explain the reasoning process using object-centric and relation-centric descriptions only.
In addition, the generated CoT is constrained to contain clear step boundaries, which follows the broader practice of step-by-step reasoning and process-sensitive supervision in large models~\cite{wei2022chain,lightman2023let,zheng2025processbench,jiang2025mme}, usually two to four core reasoning steps, with explicit discourse markers such as ``First,'' ``Next,'' ``Then,'' and ``Finally.''
This requirement improves both readability and verifiability, and makes the CoT traces more suitable for later post-training.

\noindent \textbf{Answer Summarization and CoT Normalization.}
The raw reasoning output is further summarized into a standardized answer$+$CoT format.
This step preserves the critical logical flow while removing redundancy and normalizing style across question types.
For each example, the final output contains a concise answer together with a short, structured reasoning chain.
The summarization prompt is also type-aware: for instance, object-identification questions are normalized toward final object names, angle-calculation questions toward discrete angle outputs, direction-identification questions toward compass labels, and visibility questions toward binary yes/no conclusions.
This normalization makes the resulting CoT corpus more consistent for both evaluation and post-training.

\noindent \textbf{CoT Quality Scoring and Final Validation.}
After normalization, the CoT traces are scored jointly on the reasoning process and the final answer.
As stated in the main paper, we evaluate CoT quality from six dimensions:
\textit{\textbf{(1) reasoning format compliance}},
\textit{\textbf{(2) reasoning structure}},
\textit{\textbf{(3) scene information utilization}},
\textit{\textbf{(4) type-specific auto-penalties}},
\textit{\textbf{(5) answer format compliance}}, and
\textit{\textbf{(6) answer correctness}}.
The scoring prompt separately assesses the quality of the reasoning chain and the final answer, which helps distinguish between failures caused by flawed intermediate logic and those caused by incorrect final extraction.
Concretely, \textit{\textbf{reasoning format compliance}} checks whether the CoT follows the natural-language-only requirement and avoids coordinate leakage.
\textit{\textbf{Reasoning structure}} evaluates whether the chain is decomposed into coherent intermediate steps with appropriate transitions.
\textit{\textbf{Scene information utilization}} measures whether the reasoning explicitly uses scene-supported evidence, such as relative positions, nearby relations, movement distances, wall barriers, or window transparency.
\textit{\textbf{Type-specific auto-penalties}} verify that each question family uses the correct reasoning procedure, such as multi-hop intermediate resolution for relational questions or movement-then-rotation ordering for embodied questions.
\textit{\textbf{Answer format compliance}} checks whether the final answer matches the expected output form of the question type.
Finally, \textit{\textbf{answer correctness}} verifies consistency with the validated QA label.
Only QA pairs with passed CoT traces are retained for the final dataset.
Since careful CoT validation is substantially more expensive than answer-only verification, we combine automatic filtering with expert auditing rather than manually reviewing every single reasoning chain in full detail.
Specifically, after the automatic scoring stage, the retained CoT data undergo final expert checking, and a random sample of nearly $400$ examples is subjected to rigorous human evaluation.
The dataset is accepted only after the sampled quality satisfies the target reliability threshold reported in the main paper.
This design allows us to maintain both scalability and annotation rigor in constructing a large-scale panoramic reasoning benchmark.
Overall, the data construction pipeline of OmniCoT is centered on one principle: each example should require real panoramic spatial reasoning rather than superficial local matching.
The question-generation stage controls linguistic naturalness and answer uniqueness, while the CoT-generation stage ensures that the final reasoning traces remain interpretable, structured, and scene-grounded.
By combining prompt-level constraints, dual-judge filtering, and expert verification, our pipeline produces a benchmark that is both large-scale and sufficiently reliable for evaluating and training panoramic reasoning MLLMs.

\subsection{Paraphrase Robustness}
\begin{table*}
\centering
\caption{\textbf{Paraphrase robustness of OmniCoT.}
We compare model accuracy on the original question group and the paraphrased question group.
The small performance shifts indicate that OmniCoT is not overly sensitive to surface-level wording.}
\label{tab:paraphrase_robustness}
\setlength{\tabcolsep}{4pt}
\renewcommand{\arraystretch}{0.9}
\resizebox{\textwidth}{!}{%
\begin{tabular}{lcccccc}
\toprule
\textbf{\textit{Group}} &
\textbf{\textit{OmniCoT-R1}} &
\textbf{\textit{GPT-5}} &
\textbf{\textit{Qwen2.5-VL-72B}} &
\textbf{\textit{Step3}} &
\textbf{\textit{GLM-4.5v}} &
\textbf{\textit{Qwen2.5-VL-7B}} \\
\midrule
Original Acc. &
59.54 &
39.72 &
35.26 &
31.23 &
29.95 &
23.53 \\
Paraphrase Acc. &
58.08~($-1.46$) &
40.62~($+0.90$) &
34.27~($-0.99$) &
30.93~($-0.30$) &
28.91~($-1.04$) &
24.02~($+0.49$) \\
\bottomrule
\end{tabular}%
}
\end{table*}
To examine whether OmniCoT is overly sensitive to surface-level wording, we further construct a paraphrased version of the benchmark questions.
Specifically, we select $6175$ questions as the original group and generate $12350$ paraphrased questions by rewriting each question with synonymous expressions while preserving its underlying spatial reasoning structure and ground-truth answer.
This experiment is designed to test whether model performance is mainly driven by rigid templates or by genuine spatial reasoning ability.
As shown in Table~\ref{tab:paraphrase_robustness}, the performance of representative models remains stable between the original and paraphrased groups.
The absolute accuracy shifts are small across both open-source and closed-source models.
These results suggest that OmniCoT is not primarily solved through question-template memorization, and that its linguistic formulation is reasonably robust to synonymous paraphrasing.

\subsection{Benchmark Experiments}

In this section, we provide additional details for the benchmark protocol, including the evaluation metrics, the model input format, and the annotation procedure for the real-world subset.
Our evaluation is designed to assess both \textit{\textbf{task success}} and \textit{\textbf{reasoning quality}}, following recent benchmark trends that distinguish final-answer performance from intermediate reasoning quality in multimodal spatial reasoning~\cite{jiang2025mme,liu2023visual,cai2025spatialbot,daxberger2025mm,stogiannidis2025mind,xu2025spatialbench}. The former measures whether a model reaches the correct final answer, while the latter examines whether the generated Chain-of-Thought (CoT) remains faithful, informative, and geometrically grounded in panoramic scenes.

\noindent \textbf{Evaluation Overview.}
Following the main paper, benchmark performance is evaluated from two complementary aspects.
First, we report final-answer accuracy over the three progressive reasoning dimensions of \textbf{\textit{See, Locate, and Move}}, together with the overall average.
Second, we evaluate CoT quality using six metrics, including three general reasoning metrics and three panorama-specific spatial reasoning metrics.
This design allows us to distinguish models that happen to output the correct answer from those that genuinely perform structured panoramic reasoning.

\noindent \textbf{General CoT Quality Metrics.}
For general reasoning quality, we follow MME-CoT~\cite{jiang2025mme} and adopt \textbf{\textit{Precision}}, \textbf{\textit{Recall}}, and \textbf{\textit{F1-score}}.
Precision measures the faithfulness of the generated reasoning, \emph{i.e.}, the proportion of valid reasoning content among all generated steps.
Recall measures informativeness, \emph{i.e.}, the extent to which the model covers the key reasoning content required for solving the problem.
F1-score is the harmonic mean of Precision and Recall and provides a balanced summary of these two aspects.
These metrics mainly capture generic reasoning quality, but not explicitly evaluating whether the reasoning is geometrically consistent in panoramic spatial setting.
To address this limitation, we further introduce three panorama-specific spatial reasoning metrics.

\noindent \textbf{Viewpoint Consistency (VC).}
Viewpoint Consistency evaluates whether the model maintains a coherent spatial perspective throughout the reasoning process.
Instead of assigning a single score to the entire CoT directly, the judge first extracts all viewpoint-related statements from the reasoning.
These include any claims involving direction, orientation, relative position, or agent pose, such as left/right, front/back, facing, turn clockwise, look toward, behind, and cardinal-direction updates such as north or east.
The extraction is restricted to statements actually used to support the answer, rather than merely restating the question.
Near-duplicate statements are merged to avoid artificially inflating the count.

Each extracted statement is then scored independently against the ground-truth scene description on a continuous scale from $0.0$ to $1.0$ with a step size of $0.1$.
A score of $0.0$ indicates a clear contradiction with the scene geometry, scores between $0.1$ and $0.4$ indicate that the statement is largely incorrect or highly ambiguous, scores between $0.5$ and $0.8$ indicate that the statement is mostly consistent but somewhat underspecified, and scores between $0.9$ and $1.0$ indicate that the statement is fully consistent and precise.
If no viewpoint-related statement is found, the output is treated as empty rather than automatically corrected.
For a given sample, the final VC score is computed as the arithmetic mean of the per-statement scores:
\begin{equation}
\mathrm{VC}(x) =
\frac{1}{|\mathcal{S}_{\mathrm{vp}}(x)|}
\sum_{s \in \mathcal{S}_{\mathrm{vp}}(x)} \mathrm{score}(s),
\end{equation}
where $\mathcal{S}_{\mathrm{vp}}(x)$ denotes the set of extracted viewpoint-related statements for sample $x$.
At the dataset level, we report the average VC over all samples with valid extracted statements.
This metric, therefore measures whether the model preserves viewpoint coherence during intermediate reasoning, rather than only whether the final answer is correct.

\noindent \textbf{Spatial Evidence Sufficiency (SES).}
Following the definition in the main paper, SES measures
whether the minimal required spatial relations are explicitly cited in the reasoning.
In our implementation, the judge further assigns a graded support score (0.0–1.0) for each relation. The reported SES value is computed as the average of these relation-level scores.
For each sample, the judge first identifies the \textit{\textbf{minimal set of spatial relationships}} that are necessary for solving the given question.
Importantly, this set is intentionally minimal: the judge is instructed not to enumerate all available scene facts, but only the relations that are essential for deriving the answer.
Typical examples include adjacency, relative direction, containment, in-front/behind relations, nearest-object constraints, distance ordering, and object location with respect to a landmark.

After identifying these necessary relationships, the judge checks whether each one is actually supported by the model's reasoning.
The evaluation follows a strict evidence rule: a relationship can only receive a non-zero score if the judge can point to a direct quote or faithful near-quote from the generated reasoning as supporting evidence.
If no such evidence can be found, the corresponding relationship must receive a score of 0.0 and its evidence field must be set to \texttt{null}, even if the model might have implicitly relied on the correct fact.
This makes SES a stricter measure than semantic similarity or answer overlap, because it requires the reasoning to explicitly expose the necessary spatial support.

Each key relationship is scored on a $0.0 \sim 1.0$ scale with a step size of $0.1$.
A score of $0.0$ means that the relationship is unsupported or has no evidence quote, scores between $0.1$ and $0.4$ indicate only weak or partial support, scores between $0.5$ and $0.8$ indicate that the relationship is clearly supported but slightly underspecified, and scores between $0.9$ and $1.0$ indicate that the required spatial evidence is clearly and precisely provided.
For sample $x$, let $\mathcal{R}(x)$ denote the minimal set of necessary spatial relationships.
The SES score is computed as
\begin{equation}
\mathrm{SES}(x) =
\frac{1}{|\mathcal{R}(x)|}
\sum_{r \in \mathcal{R}(x)} \mathrm{score}(r).
\end{equation}
In addition to the average SES score, we also track an \textit{\textbf{evidence rate}}, defined as the fraction of necessary relationships for which the reasoning contains a valid evidence quote or near-quote.
This auxiliary statistic helps distinguish between low scores caused by weak support and those caused by entirely missing evidence.
Overall, SES measures whether a reasoning trace provides enough explicit spatial grounding to justify its answer.

\noindent \textbf{Reasoning Feasibility (RF).}
Reasoning Feasibility evaluates whether the execution steps described in the reasoning are physically and geometrically plausible under the scene constraints.
Here, an execution step refers to any implied action, movement, viewpoint update, or perceptual operation performed by the agent, including body movement (walk, move, approach), viewpoint actions (turn, rotate, face, look toward, scan), and simple interactions like reach, when applicable.
The judge extracts such steps from the reasoning and evaluates each step individually against the ground-truth scene.
Moreover, the judge follows an explicit coverage rule: if the reasoning contains viewpoint-action cues, it must extract at least one executable step rather than returning an empty result.

Each execution step is scored on a $0.0 \sim 1.0$ scale with a step size of $0.1$.
A score of $0.0$ indicates that the step is clearly infeasible, scores between $0.1$ and $0.4$ indicate strong doubt or a major constraint violation, scores between $0.5$ and $0.8$ indicate that the step is feasible but lacks some details, and scores between $0.9$ and $1.0$ indicate that the step is clearly feasible and well grounded.
In addition, several hard constraints are enforced during scoring.
If a step refers to an object or location not present in the scene, its score is capped at $0.2$.
If a step implies moving through walls or closed barriers without valid supporting evidence, its score is capped at $0.2$.
If a step requires an impossible movement, such as teleportation or an unrealistic displacement in a single step, its score is also capped at 0.2.

For sample $x$, let $\mathcal{E}(x)$ denote the extracted set of execution steps.
The RF score is defined as
\begin{equation}
\mathrm{RF}(x) =
\frac{1}{|\mathcal{E}(x)|}
\sum_{e \in \mathcal{E}(x)} \mathrm{score}(e).
\end{equation}
We additionally report a feasibility rate, namely the fraction of extracted steps whose feasibility score is at least $0.5$.
This statistic reflects how often the reasoning remains executable at the step level, even when some details are partially underspecified.
Compared with VC and SES, RF focuses more directly on embodied plausibility and whether the described reasoning process could actually be carried out in the scene.

\noindent \textbf{Implementation Details of the Judge Outputs.}
For the three panorama-specific metrics, the judge does not directly assign one holistic score to the entire reasoning trace.
Instead, it first decomposes the reasoning into metric-specific evaluation units and then scores each unit independently.
For \textbf{VC}, the judge returns a list of extracted viewpoint-related statements, where each item contains the statement text and its consistency score.
For \textbf{SES}, the judge returns the minimal necessary spatial relationships together with their sufficiency scores and evidence quotes from the reasoning.
For \textbf{RF}, the judge returns the extracted execution steps and their feasibility scores.
The final sample-level metric value is then computed automatically by averaging over these extracted units.
This design improves interpretability and reduces the instability of assigning a single coarse score to a long reasoning chain.

\subsection{Additional Benchmark Protocol Analyses}

\noindent \textbf{Tolerance Sensitivity for Angle-sensitive Questions.}
Most OmniCoT answers are represented as discrete object identities, visibility labels, or cardinal/inter-cardinal directions.
Therefore, the benchmark does not generally rely on strict free-form matching of continuous quantities.
For angle-sensitive RAC questions, we convert angular reasoning into closest-option selection, such as choosing the closest angle among $45^\circ$, $90^\circ$, $135^\circ$, and $180^\circ$.
This design avoids penalizing models for insignificant numeric formatting differences while still requiring them to perform correct angular reasoning.

To further verify that the evaluation is not sensitive to minor angular deviations, we conduct a tolerance sensitivity analysis under different angular tolerance margins.
As shown in Table~\ref{tab:tolerance_sensitivity}, the overall accuracy remains nearly unchanged under $\pm 5^\circ$, $\pm 10^\circ$, and $\pm 20^\circ$ tolerance settings.
This confirms that OmniCoT does not unfairly penalize small numerical deviations and that the reported results are stable with respect to tolerance choices.

\begin{table*}[t]
\centering
\caption{\textbf{Tolerance sensitivity for angle-sensitive questions.}
We report overall accuracy under different angular tolerance margins.
The results remain stable, indicating that the benchmark is insensitive to minor angular deviations.}
\label{tab:tolerance_sensitivity}
\setlength{\tabcolsep}{5pt}
\renewcommand{\arraystretch}{0.9}
\resizebox{0.85\textwidth}{!}{%
\begin{tabular}{lccccc}
\toprule
\textbf{\textit{Tolerance}} &
\textbf{\textit{GPT-5}} &
\textbf{\textit{Qwen2.5-VL-72B}} &
\textbf{\textit{Step3}} &
\textbf{\textit{GLM-4.5v}} &
\textbf{\textit{Qwen2.5-VL-7B}} \\
\midrule
$\pm 0^\circ$ Acc.  & 39.72 & 35.26 & 31.24 & 29.95 & 23.53 \\
$\pm 5^\circ$ Acc.  & 39.76~($+0.04$) & 35.27~($+0.01$) & 31.27~($+0.03$) & 29.98~($+0.03$) & 23.55~($+0.02$) \\
$\pm 10^\circ$ Acc. & 39.76~($+0.04$) & 35.27~($+0.01$) & 31.27~($+0.03$) & 29.98~($+0.03$) & 23.55~($+0.02$) \\
$\pm 20^\circ$ Acc. & 39.77~($+0.05$) & 35.27~($+0.01$) & 31.27~($+0.03$) & 29.99~($+0.04$) & 23.56~($+0.03$) \\
\bottomrule
\end{tabular}%
}
\end{table*}

\noindent \textbf{ERP vs. Multi-view Projection.}
The main benchmark uses ERP panoramas as the default input format because ERP preserves the continuous $360^\circ$ spatial context.
However, ERP images also introduce geometric distortion.
To examine whether model failures are caused merely by ERP distortion, we additionally evaluate representative models under two perspective-projection settings: a 4-view perspective input without the up/down views, and a complete 6-view cubemap input.
The 4-view setting reduces local distortion while covering the horizontal surroundings, whereas the 6-view cubemap further includes the upward and downward directions.

As shown in Table~\ref{tab:projection_ablation}, multi-view projection affects individual models but does not solve OmniCoT.
Most models remain far from high accuracy under both 4-view and 6-view settings.
This indicates that the difficulty of OmniCoT cannot be attributed merely to ERP distortion; the benchmark also requires assembling global multi-hop spatial evidence across the full scene.
Meanwhile, the slight gain of GPT-4o under the 4-view setting suggests that projection format can still affect model-specific behavior, making multi-view evaluation a useful diagnostic complement.

\begin{table*}[t]
\centering
\caption{\textbf{Input projection ablation.}
We compare ERP input with 4-view perspective input and 6-view cubemap input.
Multi-view projection mitigates local distortion but does not eliminate the global multi-hop reasoning challenge.}
\label{tab:projection_ablation}
\setlength{\tabcolsep}{5pt}
\renewcommand{\arraystretch}{0.9}
\resizebox{\textwidth}{!}{%
\begin{tabular}{lccccc}
\toprule
\textbf{\textit{Format}} &
\textbf{\textit{OmniCoT-R1}} &
\textbf{\textit{GPT-5}} &
\textbf{\textit{Qwen2.5-VL-72B}} &
\textbf{\textit{GPT-4o}} &
\textbf{\textit{Gemini3-Flash}} \\
\midrule
ERP Acc. &
59.54 & 39.72 & 35.26 & 31.58 & 28.43 \\
4-view Acc. &
52.87~($-6.67$) &
34.40~($-5.32$) &
30.59~($-4.67$) &
32.47~($+0.89$) &
24.47~($-3.96$) \\
6-view Cubemap Acc. &
53.86~($-5.68$) &
34.72~($-5.00$) &
30.77~($-4.49$) &
31.95~($+0.37$) &
25.38~($-3.05$) \\
\bottomrule
\end{tabular}%
}
\end{table*}

\begin{wraptable}{r}{0.4\textwidth}
\vspace{-1.2em}
\caption{\textbf{Scene-category distribution of OmniCoT-Real.} 
OmniCoT-Real contains $1073$ entries spanning $13$ indoor scene categories, covering a diverse range of real-world environments for evaluating panoramic spatial reasoning.}
\centering
\resizebox{0.4\textwidth}{!}{%
\begin{tabular}{l|c|c}
\toprule
\textbf{Scene Type} & \textbf{Count} & \textbf{Percentage} \\
\midrule
Living Room & 180 & 16.78\% \\
Kitchen & 131 & 12.21\% \\
Bathroom & 108 & 10.07\% \\
Corridor & 98 & 9.13\% \\
Classroom & 85 & 7.92\% \\
Gym & 78 & 7.27\% \\
Workshop & 72 & 6.71\% \\
Pantry & 66 & 6.15\% \\
Others & 66 & 6.15\% \\
Office & 66 & 6.15\% \\
Bedroom & 65 & 6.06\% \\
Balcony & 54 & 5.03\% \\
Canteen & 4 & 0.37\% \\
\bottomrule
\end{tabular}}

\label{table:scene_distribution}
\vspace{-1.5em}
\end{wraptable}
\noindent \textbf{Coverage-aware Statistics}
A practical challenge in LLM-as-judge evaluation is that some model outputs may be too short, too underspecified, or insufficiently structured for reliable extraction.
For example, a reasoning trace may contain no explicit viewpoint-related statement, no quotable spatial evidence, or no executable action step.
In our implementation, such cases are treated as not applicable rather than being assigned a default score.
Accordingly, in addition to the average value of each metric, we also record several coverage-aware statistics, including the number of successfully scored samples and the proportion of empty outputs.

\noindent \textbf{OmniCoT-Real}
The OmniCoT-Real dataset comprises a total of $1073$ entries, distributed across $13$ scene categories as shown in Table~\ref{table:scene_distribution}. These scene types represent a variety of typical indoor environments, ensuring diverse testing conditions for spatial reasoning models. The dataset is distributed across $6$ distinct types of spatial reasoning tasks. 
These tasks cover various aspects of spatial understanding and their distribution is shown in the Table~\ref{table:task_distribution}.

\begin{table}[htbp]
\centering
\caption{\textbf{Reasoning-task distribution of OmniCoT-Real.} 
The real-world subset covers all six OmniCoT task types defined in the main paper: MOT, RAC, MOI, MDI, PTM, and RTM. The distribution is relatively balanced across the six reasoning tasks.}
\resizebox{0.7\textwidth}{!}{%
\begin{tabular}{l|c|c}
\toprule
\textbf{Reasoning Task Type} & \textbf{Count} & \textbf{Percentage (\%)} \\
\midrule
Multi-Step Orientation Tracking & 198 & 18.45\% \\
Pure Translational Movement & 197 & 18.36\% \\
Rotation-Translational Movement & 177 & 16.5\% \\
Multi-Hop Object Identification & 173 & 16.12\% \\
Relative Angular Calculation & 170 & 15.84\% \\
Multi-Hop Direction Identification & 158 & 14.73\% \\
\bottomrule
\end{tabular}}
\label{table:task_distribution}
\end{table}

\begin{table}
\centering
\caption{\textbf{Comparison between the base model and OmniCoT-R1 on OmniCoT-Real.} 
Accuracy is reported across the three reasoning dimensions defined in the main paper: 
\textbf{\textit{See}} (Multi-hop Viewpoint Transformation), 
\textbf{\textit{Locate}} (Inter-Object Spatial Relationship), and 
\textbf{\textit{Move}} (Embodied Action Simulation), together with the overall accuracy.}
\resizebox{0.5\textwidth}{!}{%
\begin{tabular}{l|cccc}
\toprule
\textbf{Model} & \textbf{See} & \textbf{Locate} & \textbf{Move} & \textbf{Overall} \\
\midrule
Qwen2.5-VL-7B & 18.48\% & 5.74\% & 20.59\% & 15.28\% \\
OmniCoT-R1 & 19.02\% & 9.37\% & 28.34\% & 19.29\% \\
\bottomrule
\end{tabular}}
\label{tab:real_world_results}
\end{table}
\noindent \textbf{Real-World Test Results}
The performance of the OmniCoT-R1 model was evaluated on a real-world test set from the OmniCoT-Real dataset. The results in Table~\ref{tab:real_world_results} indicate that while the model shows promising performance, it faces challenges in certain spatial reasoning tasks. The accuracy of the model on various reasoning tasks is presented below:
The highest accuracy is achieved in the Move reasoning task (28.34\%). The Locate and See tasks show significantly lower performance, with accuracy rates of 9.37\% and 19.02\%, respectively. These results suggest that while the model is capable of handling simpler tasks, it struggles with more complex reasoning scenarios involving occlusion and dynamic object relationships. 
In summary, while the OmniCoT-R1 model shows reasonable success in certain tasks, its performance in real-world scenarios remains a challenge. Future improvements will focus on improving robustness and generalization to a wider range of real-world data, particularly in complex, cluttered environments.

\subsection{Post-Training}

Building upon OmniCoT-T, we develop \textbf{OmniCoT-R1} through a two-stage post-training pipeline consisting of supervised fine-tuning (SFT) followed by Group Relative Policy Optimization (GRPO)~\cite{shao2024deepseekmath}.
The overall design is motivated by a central observation of panoramic reasoning: unlike conventional VQA, solving OmniCoT questions requires not only recognizing objects, but also maintaining a stable global frame over long-horizon intermediate steps, including viewpoint updates, multi-hop relational grounding, and movement-conditioned visibility reasoning.
As a result, post-training should not merely improve answer matching; it must also encourage the model to produce structured reasoning traces that remain grounded in the geometry of the panorama.

\noindent \textbf{Stage 1: SFT for Structured Panoramic Reasoning.}
In the first stage, we initialize from \textbf{Qwen2.5-VL-7B-Instruct}~\cite{bai2025qwen25} and perform supervised fine-tuning on OmniCoT-T.
The goal of this stage is to teach the model a stable reasoning protocol before reinforcement learning.
Concretely, the model is trained to produce responses in an explicit
\texttt{<think>...\allowbreak</think><answer>...\allowbreak</answer>} schema, where the \texttt{<think>} block contains a natural-language Chain-of-Thought grounded in the panoramic scene and the \texttt{<answer>} block contains the final concise answer~\cite{wei2022chain,lightman2023let,zheng2025processbench}.
During SFT, we update only the language model parameters while freezing the vision encoder and projection layers, so that the model first learns a panoramic reasoning format on top of an already capable visual backbone.
This stage is important because GRPO is substantially more stable when the initial policy already knows how to produce valid structured outputs rather than discovering the target response format from scratch. 

The two-stage design is especially important in panoramic spatial reasoning.
In pinhole-image tasks, many questions can often be solved through local recognition plus short verbal justification.
By contrast, OmniCoT questions frequently require the model to preserve scene-wide consistency across several latent updates: turning from one heading to another, resolving intermediate anchors, simulating motion through space, and checking whether a target remains visible after movement or rotation.
These long-horizon operations are exactly where free-form reasoning is prone to repeat collapse into locally plausible but globally inconsistent chains.
SFT therefore provides a structured scaffold for panoramic reasoning, while the subsequent GRPO stage further sharpens the model's ability to choose answer-relevant and geometrically coherent reasoning trajectories.

\noindent \textbf{Stage 2: GRPO for Long-Horizon Spatial Refinement.}
Starting from the SFT checkpoint, we then apply GRPO on OmniCoT-T to refine the model's reasoning behavior under reward supervision.
Following our implementation in SWIFT~\cite{zhao2025swift}, each prompt is sampled into multiple candidate completions, and the policy is optimized according to their relative rewards within the group.
In our experiments, the number of sampled completions per prompt is set to \textbf{8}, which provides sufficient diversity for relative policy improvement while keeping generation cost manageable.

A key reason for introducing GRPO is that supervised targets alone do not fully determine which reasoning behaviors are most useful for panoramic reasoning.
Different completions may all follow the same output schema, yet differ substantially in whether they preserve viewpoint consistency, avoid spatial drift, and terminate in a concise answer instead of a repetitive or degenerate chain.
GRPO provides a direct mechanism for preferring those generations that are both structurally valid and task-effective.

\noindent \textit{Reward Design.}
Our GRPO reward is defined as a weighted combination of three rule-based components:
\begin{equation}
R = \lambda_{\mathrm{fmt}} R_{\mathrm{fmt}}
+ \lambda_{\mathrm{acc}} R_{\mathrm{acc}}
+ \lambda_{\mathrm{rep}} R_{\mathrm{rep}},
\end{equation}
where $R_{\mathrm{fmt}}$ is the format reward,
$R_{\mathrm{acc}}$ is the answer-accuracy reward, and
$R_{\mathrm{rep}}$ is the repetition regularization term.
In our implementation,
\begin{equation}
\lambda_{\mathrm{fmt}} = 0.1,\qquad
\lambda_{\mathrm{acc}} = 1.0,\qquad
\lambda_{\mathrm{rep}} = 0.2.
\end{equation}
This weighting reflects our design principle that correct task execution should dominate the reward, while response structure and anti-degeneration behavior act as auxiliary but necessary constraints.

\paragraph{Format reward.}
The format reward encourages the model to preserve the explicit reasoning schema established during SFT.
Given a completion $y$, we define
\begin{equation}
R_{\mathrm{fmt}}(y)=
\begin{cases}
1, & \text{if } y \text{ matches }
\texttt{<think>...\allowbreak</think><answer>...\allowbreak</answer>},\\
0, & \text{otherwise.}
\end{cases}
\end{equation}
In implementation, this is checked by a regular-expression matcher that searches for a valid \texttt{<think>} block followed by a valid \texttt{<answer>} block.
This reward is deliberately lightweight rather than dominant: its role is not to force a particular writing style, but to ensure that the model continues to externalize a reasoning trace before giving the final answer.
For panoramic tasks, this explicit separation is particularly useful because many errors arise in the intermediate reasoning process rather than in the final answer token alone.

\textbf{\noindent \textit{Accuracy Reward.}}
The core optimization signal is the answer-accuracy reward.
Given a completion $y$ and the ground-truth answer $a^\star$, we first extract the predicted answer from the \texttt{<answer>} block:
\begin{equation}
\hat{a} = \mathrm{ExtractAnswer}(y).
\end{equation}
We then normalize both the prediction and the target:
\begin{equation}
\tilde{a} = \mathrm{Normalize}(\hat{a}),\qquad
\tilde{a}^{\star} = \mathrm{Normalize}(a^\star).
\end{equation}
Finally, the accuracy reward is defined as exact match after normalization:
\begin{equation}
R_{\mathrm{acc}}(y,a^\star)=
\begin{cases}
1, & \text{if } \tilde{a} = \tilde{a}^{\star},\\
0, & \text{otherwise.}
\end{cases}
\end{equation}

The extraction and normalization process is tailored to the answer space of OmniCoT.
Specifically, the parser first reads the content inside the \texttt{<answer>} tag and then canonicalizes it according to the expected answer type.
For binary visibility questions, outputs are mapped to \texttt{yes}/\texttt{no}.
For angle questions, the parser extracts the numeric degree value and removes option prefixes such as \texttt{A)} or \texttt{B)}.
For open-form object or direction answers, normalization lowercases the string, removes articles such as \texttt{the}, \texttt{a}, and \texttt{an}, strips punctuation, and collapses superficial formatting differences.
This design is important because OmniCoT contains heterogeneous answer forms, including object names, compass directions, angle values, and yes/no judgments.
A panoramic reasoning model should be rewarded for getting the geometrically correct answer, rather than being penalized for irrelevant surface variation.

Although the implementation operates on normalized answer strings, the supervision signal is geometry-grounded at the dataset level.
The gold answers in OmniCoT-T are derived from validated scene annotations and multi-stage filtering, so a correct normalized answer corresponds to a correct execution of the underlying spatial reasoning task.
This is particularly meaningful in panoramic scenes, where many questions depend on global geometric consistency rather than local object appearance alone.
For example, in viewpoint transformation and move-turn questions, the answer can only be correct if the model implicitly tracks heading changes, spatial anchors, and visibility constraints over the full panorama.
Therefore, the binary exact-match reward serves as a compact proxy for successful panoramic reasoning execution.

\textbf{\noindent \textit{Repetition Regularization.}}
Long-horizon reinforcement learning on reasoning tasks is prone to degeneration, especially when the model discovers that longer chains can sometimes preserve reward while avoiding commitment.
This issue is amplified in panoramic reasoning, where the model may repeatedly restate directions, anchors, or movement steps without contributing new information.
To suppress this behavior, we include a repetition regularization term:
\begin{equation}
R_{\mathrm{rep}}(y) \le 0,
\end{equation}
which penalizes repeated $n$-gram patterns in the generated completion.
In our implementation, this reward follows the built-in repetition penalty in SWIFT~\cite{zhao2025swift} and computes a negative score based on the proportion of repeated $n$-grams in the output sequence.
The final reward therefore discourages verbose or circular CoT traces while still allowing sufficient length for multi-step reasoning.

\textbf{\noindent \textit{Combined Training Objective.}}
Let $\pi_{\theta}$ denote the current policy and let $\{y_i\}_{i=1}^{G}$ be the $G$ sampled completions for a given prompt, where $G=8$ in our experiments.
Each completion is assigned the scalar reward
\begin{equation}
R_i =
0.1\,R_{\mathrm{fmt}}(y_i)
+ 1.0\,R_{\mathrm{acc}}(y_i,a^\star)
+ 0.2\,R_{\mathrm{rep}}(y_i).
\end{equation}
GRPO then updates the policy by preferring completions with higher relative rewards within the sampled group.
This means the model is encouraged to choose responses that simultaneously satisfy three conditions:
they follow the structured \texttt{<think>}--\texttt{<answer>} protocol,
they arrive at the correct answer under OmniCoT supervision,
and they avoid repetitive degeneration during long-form reasoning.
For panoramic spatial reasoning, these pressures are complementary: the first stabilizes response structure, the second anchors learning to task success, and the third prevents the reasoning trace from collapsing into low-quality loops.

\textbf{\noindent \textit{Training Configuration.}}
The main training hyperparameters are summarized as follows.
The base model is \textbf{Qwen2.5-VL-7B-Instruct}.All SFT and GRPO experiments are conducted on NVIDIA RTX PRO 6000 GPUs.
During SFT, only the language model is trainable, while the vision encoder and projection layers are frozen.
During GRPO, we perform full-model fine-tuning.
All training is conducted in \textbf{BF16} precision with a maximum sequence length of \textbf{4096} and a maximum completion length of \textbf{2048}.
Both SFT and GRPO use \textbf{OmniCoT-T} as the training data source.
For GRPO, the learning rate is set to \textbf{1e-6}, the number of generations per prompt is \textbf{8}, and the training is run for one epoch in our current implementation.

The relatively long sequence budget is necessary because panoramic reasoning often requires the model to jointly process a wide-field visual input, a structured prompt with global anchors, and a multi-step CoT trace.
Similarly, a low GRPO learning rate is important because the SFT checkpoint already provides a reasonably good reasoning prior, and the purpose of GRPO is to refine this prior rather than to relearn the task from scratch.
Using multiple sampled generations per prompt further improves the stability of relative preference learning by exposing the optimizer to diverse reasoning candidates for the same problem.

\textbf{\noindent \textit{Training Dynamics and Stability.}}
As shown in the main paper, the GRPO stage exhibits healthy optimization behavior.
The KL divergence between the updated policy and the reference model remains bounded, indicating that the model improves its spatial reasoning capability without drifting excessively far from the SFT initialization.
The clipping ratio stays near zero, suggesting that policy updates remain within a stable trust region.
At the same time, the repetition-related signal improves steadily during training, which is consistent with reduced degeneration in the generated reasoning traces.
Together, these observations indicate that OmniCoT-T provides a suitable post-training resource: it supports reward improvement, preserves optimization stability, and suppresses repetitive failure modes in long-horizon panoramic reasoning.

\textbf{\noindent \textit{Effectiveness of the Two-Stage Pipeline.}}
Empirically, the two-stage pipeline yields consistent gains over the base model on both answer accuracy and CoT quality.
Compared with Qwen2.5-VL-7B-Instruct, the SFT stage already provides substantial improvements by teaching the model a structured panoramic reasoning protocol.
The subsequent GRPO stage further improves overall accuracy and all three spatial reasoning metrics, indicating that reinforcement learning is particularly effective for refining long-horizon geometric consistency after the basic reasoning format has been established.

\textbf{\noindent \textit{Data Scaling on OmniCoT-T.}}
Beyond the full-data setting, we further study how OmniCoT-R1 scales with different proportions of OmniCoT-T.
Specifically, we train Qwen2.5-VL-7B with $25\%$, $50\%$, $75\%$, and $100\%$ of the training data using the same SFT+GRPO pipeline.
This experiment examines whether the proposed training set provides scalable supervision, rather than only improving performance at a single fixed data scale.
As shown in Table~\ref{tab:data_scaling}, both answer accuracy and CoT quality improve consistently as more training data is used.
The overall accuracy increases from $44.91\%$ with $25\%$ data to $59.54\%$ with the full data, while CoT F1 improves from $30.60\%$ to $51.14\%$.
The gains are especially clear in the Locate dimension, suggesting that multi-hop inter-object reasoning benefits strongly from increased structured spatial supervision.
These results indicate that OmniCoT-T is not only effective as a fixed post-training set, but also provides scalable supervision for panoramic spatial reasoning.

\begin{table*}[t]
\centering
\caption{\textbf{Data scaling on OmniCoT-T.}
We train OmniCoT-R1 with different proportions of OmniCoT-T using the same SFT+GRPO pipeline.
Both answer accuracy and CoT quality improve as the amount of training data increases.}
\label{tab:data_scaling}
\setlength{\tabcolsep}{4pt}
\renewcommand{\arraystretch}{0.9}
\resizebox{\textwidth}{!}{%
\begin{tabular}{c|cccc|cccccc}
\toprule
\textbf{\textit{Data}} &
\textbf{\textit{See}} &
\textbf{\textit{Locate}} &
\textbf{\textit{Move}} &
\textbf{\textit{Overall}} &
\textbf{\textit{Pre.}} &
\textbf{\textit{Rec.}} &
\textbf{\textit{F1}} &
\textbf{\textit{VC}} &
\textbf{\textit{SES}} &
\textbf{\textit{RF}} \\
\midrule
25\%  & 55.97 & 33.87 & 48.15 & 44.91 & 38.68 & 25.31 & 30.60 & 39.15 & 57.25 & 53.29 \\
50\%  & 57.72 & 36.33 & 50.85 & 47.27 & 43.65 & 29.70 & 35.35 & 42.71 & 60.33 & 56.45 \\
75\%  & 59.54 & 41.51 & 53.55 & 50.66 & 47.30 & 32.64 & 38.63 & 47.77 & 63.52 & 59.89 \\
100\% & 67.97 & 51.82 & 61.34 & 59.54 & 55.81 & 47.19 & 51.14 & 59.20 & 66.64 & 63.21 \\
\bottomrule
\end{tabular}%
}
\end{table*}

Overall, the post-training design of OmniCoT-R1 is centered on a panoramic-specific principle: a good reasoning model must not only answer correctly, but must also reason in a way that remains globally organized over the full $360^\circ$ scene.
SFT provides this organization explicitly through a structured reasoning schema, while GRPO further aligns generation behavior with three desirable properties: valid reasoning format, correct task execution, and non-degenerate long-form reasoning.
This combination yields a simple yet effective reinforcement-learning paradigm for panoramic spatial reasoning and serves as a reproducible baseline for future OmniCoT-style post-training.

\section{Human Evaluation}
We have performed two complementary human studies to assess (1) the large-scale accuracy of the QA generation pipeline, and (2) the performance gap between MLLMs and humans.

\subsection{Large-Scale QA Accuracy Estimation}

To estimate the overall error rate of the QA generation system, we sampled $348$ questions from the full set of $20560$ QA pairs. 
Each sampled question was manually checked by 4 annotators to judge whether the generated answer was correct with respect to the scene and question.

With the selected $348$ samples, the estimated accuracy of the full set ($N = 20560$) can be bounded using classical confidence interval analysis for proportions.
Let $n=348$ be the sample size, and let $\hat{p}$ be the observed sample error rate (\textit{i.e.}, the proportion of incorrect answers among the $348$ samples). 
Since the sampling fraction $n/N = 0.0169 < 0.05$, we can ignore the finite population 
correction factor. The standard error of $\hat{p}$ is:
\begin{equation}
\text{SE}(\hat{p}) = \sqrt{\frac{\hat{p}(1 - \hat{p})}{n}}.
\end{equation}

\begin{table}[h]
    \caption{\textbf{Human verification of QA quality on a random sample from OmniCoT.} 
This table reports the error counts and error rates for $348$ randomly sampled QA pairs from the full dataset. Results are broken down by the six OmniCoT task types defined in the main paper: MOT, RAC, MOI, MDI, PTM, and RTM.} 
    \centering
    \resizebox{\textwidth}{!}{%
    \begin{tabular}{l|ccc}
    \toprule
        \textbf{\textit{Type}} & \textbf{\textit{ Total Number}} & \textbf{\textit{ Wrong Number}} &  \textbf{\textit{Error Rate}} \\
        \midrule
        Multi-Step Orientation Tracking  &36 &2 &5.56\%  \\
        Relative Angular Calculation & 48 & 3 &6.25\% \\
        Multi-Hop Object Identification &61 &1 &1.64\% \\
        Multi-Hop Direction Identification &68 &2 &2.94\% \\
        Pure Translational Movement &68 &0 &0.00\% \\
        Rotation-Translational Movement & 67 &2 &2.99\% \\
        \midrule
        Total & 348& 10 & 2.87\% \\
    \bottomrule
    \end{tabular}
    }
    \label{tab:qa_verification}
        \vspace{-2em}
\end{table}
Substituting the values with $\hat{p}=0.0287$ (As shown in Table~\ref{tab:qa_verification}):
\begin{equation}
\text{SE}(\hat{p}) = \sqrt{\frac{0.0287 \times 0.9713}{348}} \approx 0.008947
\end{equation}

For a $95\%$ confidence interval, we use $z_{0.975} \approx 1.96$. The margin of error is:
\begin{equation}
\text{ME} = z_{0.975} \times \text{SE}(\hat{p}) \approx 1.96 \times 0.00895 \approx 0.01755
\end{equation}
Thus, the $95\%$ confidence interval for the true population error rate $p$ is:
\begin{equation}
\hat{p} \pm \text{ME} = 0.0287 \pm 0.01755 \approx [1.12\%, 4.63\%]
\end{equation}

Thus, with a 95\% confidence level, the accuracy of the entire data generation is basically [95.37\%, 98.88\%], meeting the requirements of actual testing.

\subsection{Large-Scale CoT Quality Estimation}

Building upon the 338 QA-verified samples above, we further evaluate the quality of their corresponding CoT traces. Since a valid reasoning chain presupposes a correct answer, only the QA-verified subset is meaningful for CoT assessment. 
Each of the $338$ CoT traces was independently reviewed by 4 annotators across five dimensions: format compliance, reasoning structure, scene information utilization, type-specific requirement adherence, and logical coherence. 
A CoT trace was marked as erroneous if any critical reasoning step was judged to be logically incorrect or spatially 
inconsistent with the scene.

As shown in Table~\ref{tab:cot_quality}, only 5 out of 338 
traces were identified as erroneous, yielding an observed 
error rate of $\hat{p} = 5/338 \approx 0.0148$. Following 
the same statistical framework, with $n = 338$:
\begin{equation}
\text{SE}(\hat{p}) = \sqrt{\frac{0.0148 \times 0.9852}{338}} 
\approx 0.006566
\end{equation}

For a $95\%$ confidence interval ($z_{0.975} \approx 1.96$), the margin of error is:
\begin{equation}
\text{ME} = 1.96 \times 0.006566 \approx 0.01290
\end{equation}

Thus, the $95\%$ confidence interval for the true population 
CoT error rate is:
\begin{equation}
\hat{p} \pm \text{ME} = 0.0148 \pm 0.0129 \approx [0.19\%,\ 2.77\%]
\end{equation}

Accordingly, with $95\%$ confidence, the overall CoT quality 
of the full dataset falls within $[97.23\%,\ 99.81\%]$, 
substantially exceeding the required $95\%$ acceptance threshold.
Notably, errors are mainly observed in Relative Angular Calculation (4.44\%), Multi-Hop Object Identification (3.33\%), and Multi-Step Orientation Tracking (2.94\%)
, which require precise geometric reasoning over spherical projections, while all Move-type and 
Locate-Direction questions achieved zero errors.

\begin{table}[h]
    \caption{\textbf{Human verification of CoT quality on the QA-verified subset of OmniCoT.} 
This table reports the error counts and error rates for $338$ COT traces whose corresponding QA pairs were verified as correct. Results are reported across the six OmniCoT task types defined in the main paper, showing that CoT errors are relatively rare and mainly concentrated in geometry-intensive reasoning tasks.} 
    \centering
    \resizebox{\textwidth}{!}{%
    \begin{tabular}{l|ccc}
    \toprule
        \textbf{\textit{Type}} & \textbf{\textit{Total Number}} & \textbf{\textit{Wrong Number}} & \textbf{\textit{Error Rate}} \\
        \midrule
        Multi-Step Orientation Tracking        & 34  & 1 & 2.94\% \\
        Relative Angular Calculation      & 45  & 2 & 4.44\% \\
        Multi-Hop Object Identification & 60 & 2 & 3.33\% \\
        Multi-Hop Direction Identification & 66 & 0 & 0.00\% \\
        Pure Translational Movement       & 68  & 0 & 0.00\% \\
        Rotation-Translational Movement       & 65  & 0 & 0.00\% \\
        \midrule
        Total & 338 & 5 & 1.48\% \\
    \bottomrule
    \end{tabular}
    }
    \label{tab:cot_quality}
\end{table}

\subsection{Human--Judge Agreement for CoT Evaluation}

Beyond estimating the intrinsic quality of the generated QA pairs and CoT traces, we further evaluate whether the LLM-as-judge used in our benchmark produces scores that are consistent with human judgment.
This study focuses on the reliability of the automatic process-level evaluation rather than dataset correctness.
Specifically, we randomly sample $315$ CoT responses and ask human annotators to score them under the same rubric as the LLM judge.
Each sample receives a human score $s_i^{\mathrm{human}}$ and an LLM-judge score $s_i^{\mathrm{LLM}}$.
We then compare the two score sequences using average score difference, Pearson correlation, Spearman rank correlation, and consistency rates under different tolerance margins.

Let $n=315$ be the number of sampled CoT responses.
The average score difference is defined as:
\begin{equation}
\Delta_{\mathrm{avg}} =
\frac{1}{n}
\sum_{i=1}^{n}
\left(s_i^{\mathrm{LLM}} - s_i^{\mathrm{human}}\right).
\end{equation}
A value close to zero indicates that the LLM judge does not systematically over-score or under-score the CoT responses compared with human annotators.

We further compute the Pearson correlation coefficient to measure the linear agreement between human scores and LLM-judge scores:
\begin{equation}
r =
\frac{
\sum_{i=1}^{n}
\left(s_i^{\mathrm{LLM}}-\bar{s}^{\mathrm{LLM}}\right)
\left(s_i^{\mathrm{human}}-\bar{s}^{\mathrm{human}}\right)
}{
\sqrt{
\sum_{i=1}^{n}
\left(s_i^{\mathrm{LLM}}-\bar{s}^{\mathrm{LLM}}\right)^2
}
\sqrt{
\sum_{i=1}^{n}
\left(s_i^{\mathrm{human}}-\bar{s}^{\mathrm{human}}\right)^2
}
},
\end{equation}
where $\bar{s}^{\mathrm{LLM}}$ and $\bar{s}^{\mathrm{human}}$ denote the mean LLM-judge score and the mean human score, respectively.

To measure whether the two scoring methods induce similar rankings over samples, we also report the Spearman rank correlation:
\begin{equation}
\rho =
\mathrm{Pearson}
\left(
\mathrm{rank}(s^{\mathrm{LLM}}),
\mathrm{rank}(s^{\mathrm{human}})
\right).
\end{equation}
This metric is complementary to Pearson correlation because it focuses on ranking consistency rather than absolute score scale.

Finally, we compute the consistency rate under a tolerance margin $\tau$:
\begin{equation}
\mathrm{Consistency}(\tau) =
\frac{1}{n}
\sum_{i=1}^{n}
\mathbb{I}
\left(
\left|s_i^{\mathrm{LLM}} - s_i^{\mathrm{human}}\right|
\leq \tau
\right),
\end{equation}
where $\mathbb{I}(\cdot)$ is the indicator function.
We report consistency rates under $\tau=0.1$ and $\tau=0.2$.

As shown in Table~\ref{tab:human_judge_agreement}, the average score difference between the LLM judge and human annotators is only $+0.0019$, indicating negligible systematic bias.
The Pearson correlation reaches $0.9383$, and the Spearman correlation reaches $0.9095$, showing strong agreement in both absolute score space and ranking space.
Moreover, the consistency rate reaches $82.80\%$ under a $\pm 0.1$ score tolerance and $96.77\%$ under a $\pm 0.2$ score tolerance.
These results suggest that, when provided with reference answers and precomputed spatial evidence, the LLM judge aligns well with human judgment and provides reliable process-level evaluation for panoramic CoT quality.

\begin{table}[h]
\centering
\caption{\textbf{Human--judge agreement for CoT evaluation.}
We compare human scores with LLM-as-judge scores on $315$ randomly sampled CoT responses.
The high correlation and consistency rates indicate that the automatic CoT judge aligns well with human judgment.}
\resizebox{0.8\textwidth}{!}{%
\begin{tabular}{ccccc}
\toprule
\textbf{\textit{Avg. Score Diff.}} &
\textbf{\textit{Pearson}} &
\textbf{\textit{Spearman}} &
\textbf{\textit{Consistency ($\pm 0.1$)}} &
\textbf{\textit{Consistency ($\pm 0.2$)}} \\
\midrule
$+0.0019$ & 0.9383 & 0.9095 & 82.80\% & 96.77\% \\
\bottomrule
\end{tabular}%
}
\label{tab:human_judge_agreement}
\end{table}

\section{Robustness of LLM as Judge}

Our Chain-of-Thought (CoT) evaluation relies on LLM-as-judge scoring for three spatial reasoning metrics: Viewpoint Consistency (VC), Spatial Evidence Sufficiency (SES), and Reasoning Feasibility (RF).
A natural question is whether changing the judge model alters the relative ranking of evaluated MLLMs.
We therefore compare two independent judges and quantify agreement using Spearman rank correlation for ranking agreement and Pearson correlation for linear agreement in score space. 
To ensure the generalizability and robustness of our evaluation, we curate a diverse set of open- and closed-source models exhibiting significant performance disparities, while explicitly excluding models homologous to the judge LLMs to mitigate potential bias.

Table~\ref{tab:judge_robustness_raw5} reports the original per-model metric scores produced by each judge.
All values are the judge outputs for the corresponding metric (higher is better).

\begin{table}[t]
\caption{\textbf{Robustness of LLM-as-judge evaluation across different judge models.} 
This table reports the raw panorama-specific CoT metric scores assigned by two independent judges, \textbf{\textit{DeepSeek-as-Judge}} and \textbf{\textit{Qwen3-Max-as-Judge}}, on a five-model comparison subset. The reported metrics are Viewpoint Consistency, Spatial Evidence Sufficiency, and Reasoning Feasibility, where higher scores indicate better reasoning quality.} 
\centering
\resizebox{0.8\textwidth}{!}{%
\begin{tabular}{l|ccc|ccc}
\toprule
\textbf{\textit{Model}} &
\multicolumn{3}{c|}{\textbf{\textit{DeepSeek-as-Judge}}} &
\multicolumn{3}{c}{\textbf{\textit{Qwen3-Max-as-Judge}}} \\
\cmidrule(l){2-4} \cmidrule(l){5-7}
& \textbf{\textit{VC}} & \textbf{\textit{SES}} & \textbf{\textit{RF}}
& \textbf{\textit{VC}} & \textbf{\textit{SES}} & \textbf{\textit{RF}} \\
\midrule
Doubao-1.8             & 0.6629 & 0.4994 & 0.5579 & 0.4954 & 0.5148 & 0.5185 \\
ChatGPT-4o             & 0.6299 & 0.4203 & 0.5141 & 0.4678 & 0.4230 & 0.4889 \\
InternVL3.5-14B        & 0.4694 & 0.3618 & 0.4961 & 0.3698 & 0.3780 & 0.4443 \\
Grok-4  & 0.5536 & 0.5388 & 0.4865 & 0.4188 & 0.5334 & 0.4370 \\
LLaVA-OneVision-1.5-8B & 0.4988 & 0.2184 & 0.4443 & 0.3712 & 0.3108 & 0.4010 \\
\bottomrule
\end{tabular}%
}
\label{tab:judge_robustness_raw5}
\end{table}

Let $\{(x_i, y_i)\}_{i=1}^{n}$ be the paired scores from two judges for a given metric, where
$x_i$ is the DeepSeek-as-judge score and $y_i$ is the Qwen3-Max-as-judge score for the same evaluated model.
We report two complementary correlations:

\textbf{(1) Pearson correlation.}
Pearson measures linear consistency between absolute scores:
\begin{equation}
r \;=\;
\frac{\sum_{i=1}^{n} (x_i-\bar{x})(y_i-\bar{y})}
{\sqrt{\sum_{i=1}^{n} (x_i-\bar{x})^2}\;
 \sqrt{\sum_{i=1}^{n} (y_i-\bar{y})^2}},
\end{equation}
where $\bar{x}=\frac{1}{n}\sum_i x_i$ and $\bar{y}=\frac{1}{n}\sum_i y_i$.

\textbf{(2) Spearman rank correlation.}
Spearman measures agreement in ordering by applying Pearson to ranked values:
\begin{equation}
\rho \;=\; \mathrm{Pearson}(\mathrm{rank}(x),\, \mathrm{rank}(y)).
\end{equation}
Equivalently, when there are no tied ranks, Spearman can be written as:
\begin{equation}
\rho \;=\; 1 - \frac{6\sum_{i=1}^{n} d_i^2}{n(n^2-1)},
\end{equation}
where $d_i = R_i - S_i$ is the difference between the rank $R_i$ of $x_i$ and the rank $S_i$ of $y_i$.

On the set in Table~\ref{tab:judge_robustness_raw5}, the rankings induced by the two judges are identical for each metric, yielding a perfect Spearman correlation $\rho = 1.0$. Despite judge-dependent score scaling, the linear agreement remains strong:
\textit{\textbf{VC}}: Pearson $r=0.9927$,
\textit{\textbf{SES}}: Pearson $r=0.9808$,
\textit{\textbf{RF}}: Pearson $r=0.9767$.
These findings suggest that, for this consistent core set of models, changing the judge does not alter the relative ordering across the three spatial CoT metrics, mainly behaves as a near-monotonic and close-to-linear transformation of the absolute scores.

\section{Stress Test}

To further examine whether OmniCoT can be solved by language priors alone, we conduct a stress test in which the panoramic image is completely removed and the model(Qwen2.5-VL-7B~\cite{bai2025qwen25}) is asked to answer the benchmark questions using only the textual query.
This setting is designed to test whether the benchmark can be solved by superficial answer priors or question-pattern memorization, rather than by genuine visual-spatial reasoning over the panorama~\cite{agrawal2018don,geirhos2020shortcut}.

\noindent \textbf{Text-only Stress-Test Setting.}
In this experiment, we remove the visual panoramic input and keep only the question text.
As a result, the model has no access to the scene layout, object arrangement, visibility constraints, or movement trajectories.
Any non-trivial performance under this setting must therefore come from weak language priors, answer-space bias, or dataset regularities, rather than grounded panoramic reasoning.

\noindent \textbf{Interpretation.}
As shown in the Table~\ref{tab:stress_test_text_only} under the text-only setting, the model achieves an overall accuracy of 11.76\%. This stress test supports the conclusion that OmniCoT is not primarily driven by linguistic shortcuts~\cite{agrawal2018don,geirhos2020shortcut}.
If the benchmark could be solved through question-form memorization or answer-pattern bias alone, the text-only baseline would remain reasonably competitive.
Instead, the model performs extremely poorly once the panoramic image is unavailable, indicating that solving OmniCoT requires genuine visual grounding in the scene.

\begin{wraptable}{r}{0.5\textwidth}
\vspace{-3.3em}
\caption{\textbf{Stress-test comparison under image removal.}
We compare the overall accuracy of Qwen2.5-VL-7B-Instruct in the original multimodal setting and a text-only stress-test setting where the panoramic image is removed and only the question text is provided. }
\centering
\resizebox{0.5\textwidth}{!}{%
\begin{tabular}{lc}
\toprule
\textbf{Setting} & \textbf{Overall Accuracy (\%)} \\
\midrule
Original Model & 23.53 \\
Text-only Setting (Stress Test) & 11.76 \\
\bottomrule
\end{tabular}
}
\vspace{-1.8em}
\label{tab:stress_test_text_only}
\end{wraptable}

\noindent \textbf{Implications.}
The stress-test results provide an important sanity check for the benchmark.
They show that the difficulty of OmniCoT does not mainly come from textual ambiguity or annotation artifacts, but from the need to reason over the visual structure of the scene.
In other words, the panoramic image is not merely supplementary context; it is a necessary source of information for solving the benchmark reliably.

\noindent \textbf{Discussion.}
Overall, the text-only stress test demonstrates that OmniCoT is a genuinely vision-dependent benchmark.
The severe drop in performance after removing the panoramic image confirms that benchmark success cannot be explained by language priors alone.
This result complements the main benchmark evaluation and further supports the validity of OmniCoT as a testbed for grounded $360^\circ$ spatial reasoning.

\section{Visualization}

\subsection{Visualization of OmniCoT}
\begin{figure}[t]
    \centering
    \includegraphics[width=\linewidth]{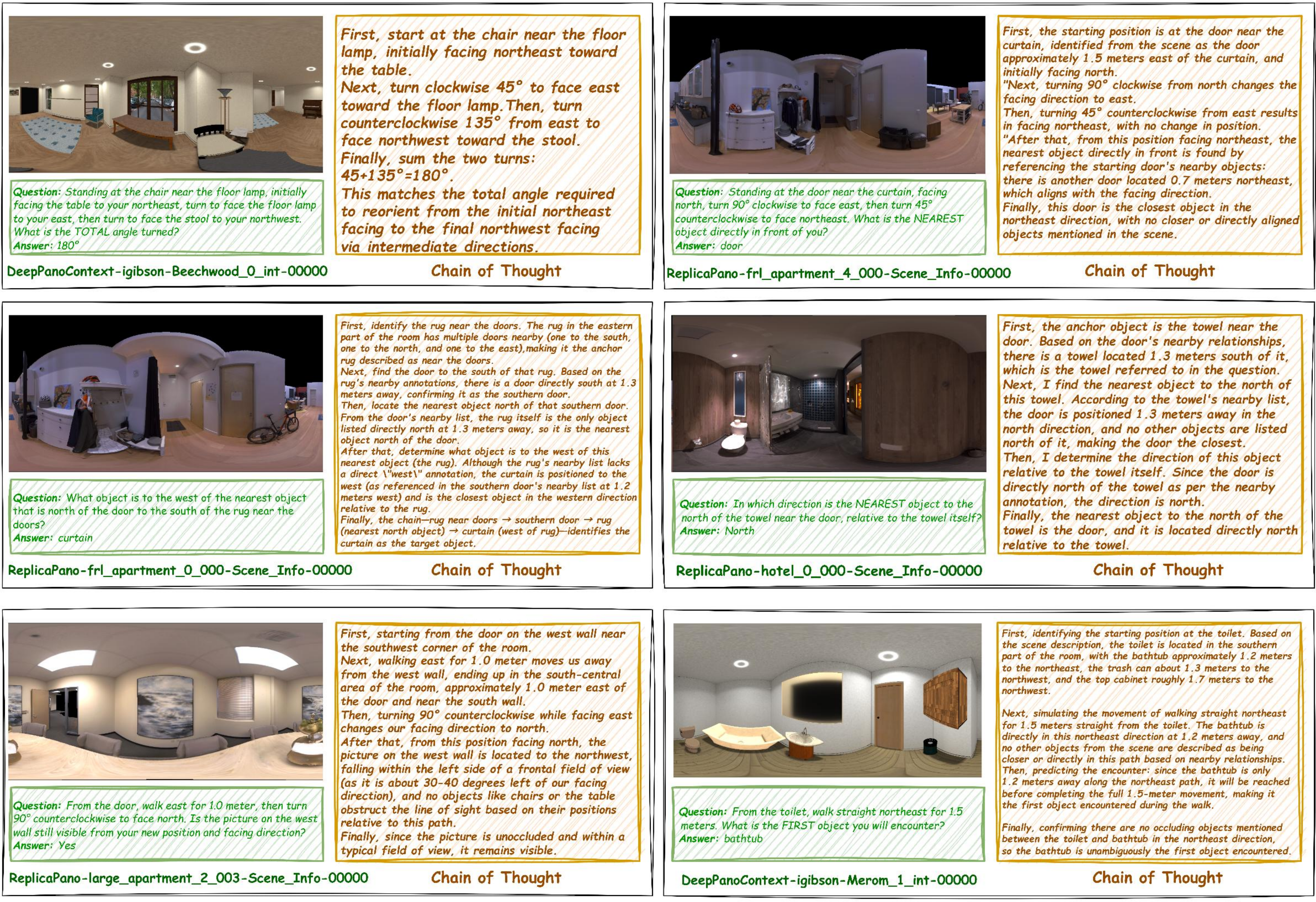}
    \caption{
\textbf{Visualization of the six OmniCoT question types.}
Each example illustrates a representative reasoning pattern from the three reasoning dimensions defined in the main paper. The figure visualizes the panoramic scene together with the corresponding question, key intermediate reasoning steps, and the final answer. 
From left to right, the examples correspond to
\textit{\textbf{Multi-Step Orientation Tracking}},
\textit{\textbf{Relative Angular Calculation}},
\textit{\textbf{Multi-Hop Object Identification}},
\textit{\textbf{Multi-Hop Direction Identification}},
\textit{\textbf{Pure Translational Movement}},
and \textit{\textbf{Rotation-Translational Movement}},
demonstrating how OmniCoT requires coordinated reasoning over viewpoint updates, multi-hop spatial relations, and embodied movement in a $360^\circ$ environment.
}
    \label{fig:vis1}
\end{figure}

To provide an intuitive understanding of the proposed benchmark, we present representative examples for all six question types in OmniCoT as shown in Fig~\ref{fig:vis1}. 
These examples are selected to illustrate how panoramic spatial reasoning in our benchmark goes beyond local object recognition and instead requires coordinated reasoning over viewpoint transformation, inter-object relations and embodied movement.
For each type, we visualize the panorama together with the corresponding question, the key reasoning steps, and the final answer.
The goal of this section is not only to show the diversity of the benchmark but also to clarify the distinct cognitive demands imposed by different question families.

\noindent \textbf{Type 1: Multi-Step Orientation Tracking (MOT).}
This type evaluates whether the model can maintain a consistent egocentric viewpoint while performing sequential rotations in a $360^\circ$ scene.
A typical example starts from a specific anchor object and asks the model to imagine facing a given direction or another object, followed by two consecutive turns.
The model must then infer which object is nearest or directly ahead under the updated orientation.
Compared with standard panoramic VQA, this type requires the model to explicitly track viewpoint updates rather than rely on local appearance cues.

\noindent  \textbf{Type 2: Relative Angular Calculation  (RAC).}
This type focuses on cumulative angular reasoning under viewpoint changes.
Instead of directly asking for an object identity, the question requires the model to compute the total angle turned across multiple re-orientations and select the closest option from a predefined set.
Such examples test whether the model can correctly align object positions with egocentric headings and aggregate multiple angular transitions into a single quantitative judgment.

\noindent  \textbf{Type 3: Multi-Hop Object Identification (MOI).}
This type evaluates relational reasoning over multiple spatial hops.
The model must first resolve an intermediate object using one relation chain anchored at a reference object, and then identify a second target object through another relation applied to the intermediate result.
The key challenge lies in preserving relational consistency across steps, especially when multiple nearby objects share similar semantic categories.

\noindent \textbf{Type 4: Multi-Hop Direction Identification (MDI).}
This type extends multi-hop reasoning from object retrieval to directional inference.
Given an anchor object, the model must identify another object through a relational description and determine the final direction of resolved object relative to the anchor.
Unlike direct object identification, this type requires the model to convert relational understanding into a directional judgment in the global scene layout.

\noindent \textbf{Type 5: Pure Translational Movement (PTM).}
This type evaluates embodied reasoning under translational movement without additional rotation.
Starting from a specific object, the model is asked to walk in a given direction for a certain distance and then determine the first object encountered along the path.
Compared with static relation questions, this type introduces a simple but important form of action-conditioned reasoning, where the model must simulate a trajectory in the scene rather than only compare fixed object positions.

\noindent \textbf{Type 6: Rotation Translational Movement (RTM).}
This type is the most comprehensive question family in OmniCoT, combining movement, viewpoint transformation, and visibility reasoning.
The model first translates to a new location, then performs an additional turn, and finally judges whether a specific target object remains visible under the updated position and facing direction.
This setup requires coordinated reasoning over pose update, field-of-view adjustment, and possible occlusion by scene structures.

Taken together, these six example groups demonstrate that OmniCoT is not a collection of isolated visual lookup questions, but a structured benchmark for multi-step panoramic spatial reasoning.
They also reveal the progressive difficulty of the benchmark: from viewpoint tracking, to relational composition, and finally to embodied action simulation.

\subsection{Visualization of the Results between Different MLLMs}

\begin{figure}
    \centering
    \includegraphics[width=\linewidth]{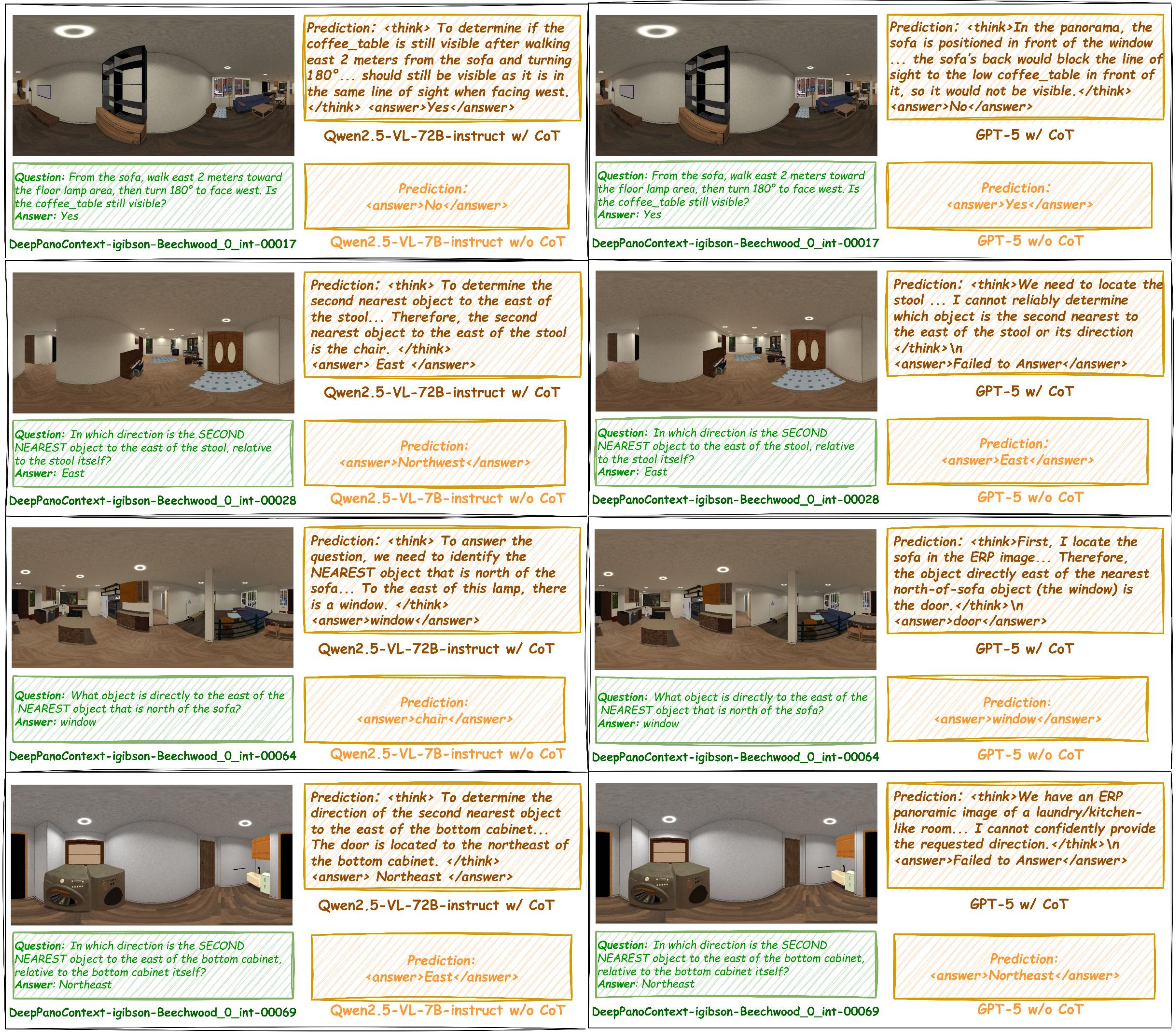}
    \caption{
\textbf{Impact of Chain-of-Thought prompting and model scale on panoramic spatial reasoning.}
We compare representative responses generated by different MLLMs under both direct-answer prompting and explicit Chain-of-Thought (CoT) prompting.
The visualization highlights how CoT often helps models expose intermediate reasoning steps such as viewpoint tracking, relational grounding, and movement simulation.
At the same time, differences between smaller and larger MLLMs reveal that stronger models maintain more consistent multi-step spatial reasoning, while weaker models frequently rely on local visual cues or produce inconsistent intermediate steps.
These examples qualitatively illustrate how both reasoning format and model capacity influence performance on OmniCoT tasks.
}

    \label{fig:mlm_compare}
\end{figure}

Beyond showing benchmark examples, we further visualize representative response differences across MLLMs, as shown in Fig.~\ref{fig:mlm_compare}.
These case studies are intended to complement the quantitative results in the main paper by revealing how different models succeed or fail on the same panoramic reasoning problem.

In particular, we focus on three aspects that are especially relevant to our benchmark: the contrast between direct answering and Chain-of-Thought reasoning, the qualitative differences between stronger and weaker MLLMs, and the remaining challenges under real-world panoramic conditions.

\begin{figure}
    \centering
    \includegraphics[width=0.7\linewidth]{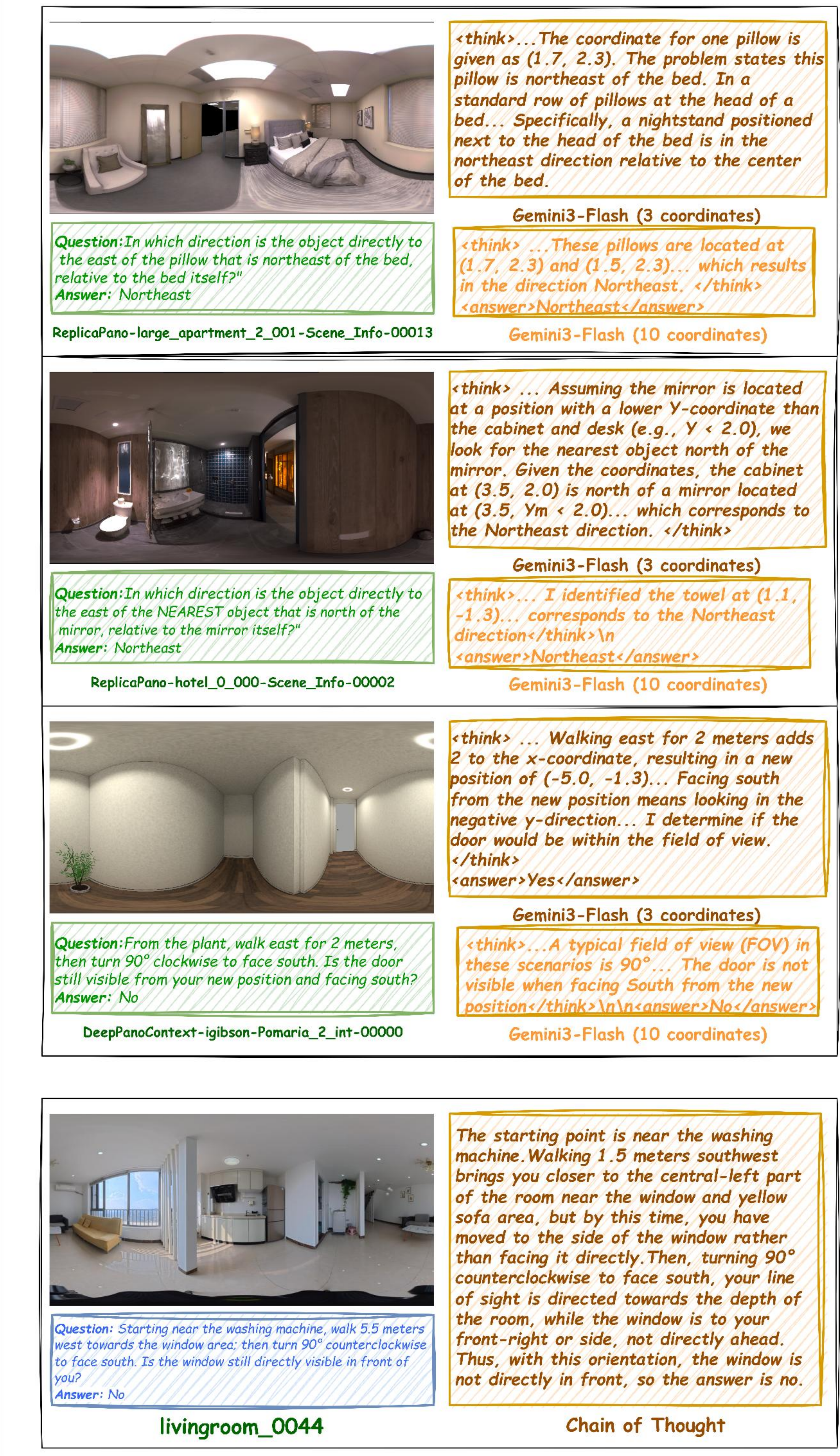}
    \caption{
\textbf{Impact of reference coordinate density and real-world scenes on panoramic reasoning.}
Examples compare reasoning with \textbf{3} vs.\ \textit{\textbf{10}} reference anchors and include cases from \textbf{\textit{real-world}} panoramas, illustrating how spatial references and scene complexity affect reasoning outcomes.
}

    \label{fig:coord_real}
\end{figure}

\begin{figure}
    \centering
    \includegraphics[width=\linewidth]{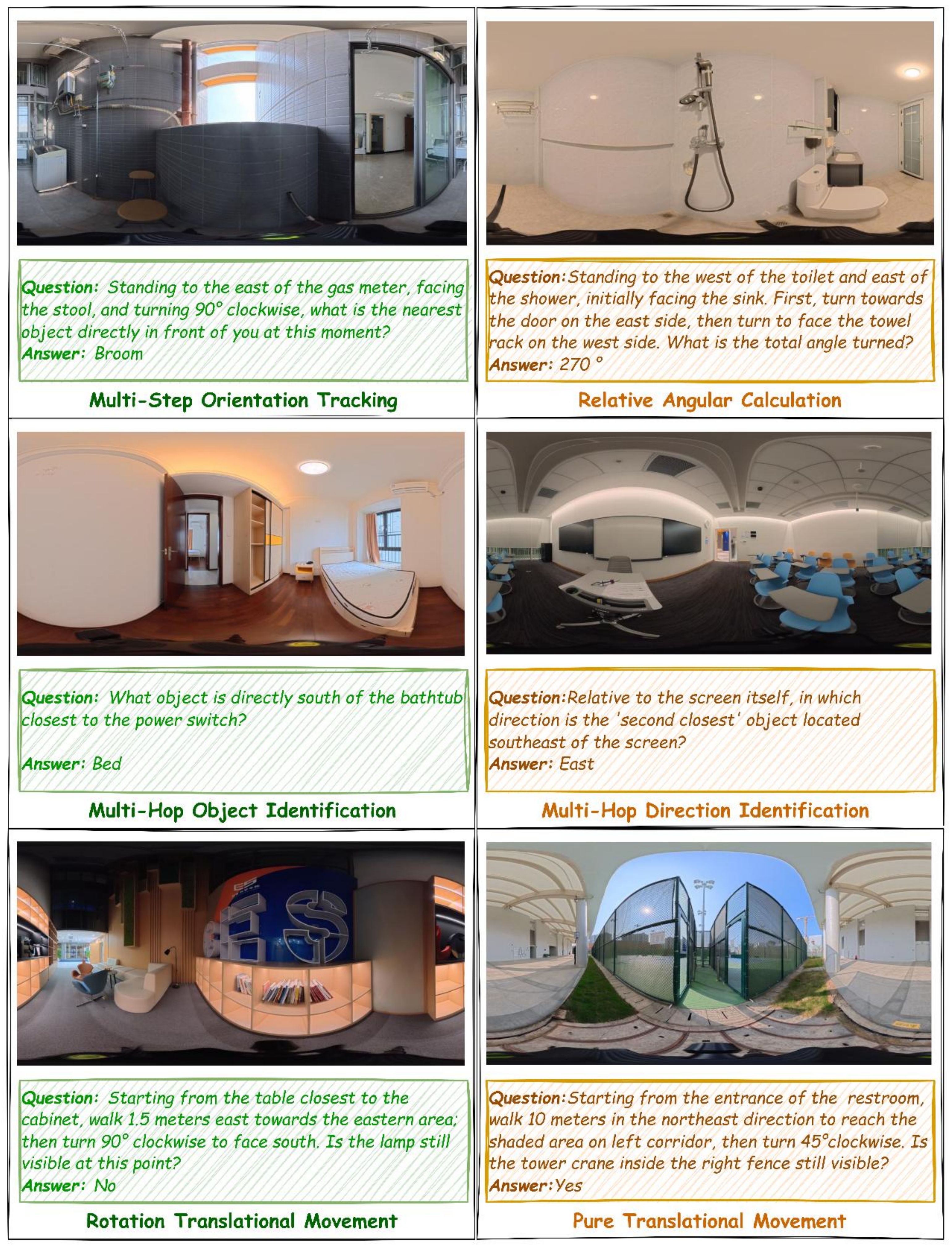}
    \caption{
    \textbf{Visualization of OmniCoT question types in real-world panoramas.}
    Examples from OmniCoT-Real covering the six reasoning types in the benchmark, including viewpoint transformation, relational reasoning, and embodied movement.
    These real-world cases illustrate additional challenges such as clutter, occlusion, and lighting variation compared with synthetic scenes.
    }
    \label{fig:vis_real}
\end{figure}

\noindent \textbf{Direct Answer vs.\ Chain-of-Thought.}
For representative examples, we compare the outputs obtained under direct-answer prompting and under explicit CoT prompting.
These visualizations show that CoT often helps when the task requires explicit viewpoint tracking, multi-hop relation composition, or embodied movement simulation.
In such cases, direct-answer predictions tend to collapse to local object priors or visually salient distractors, whereas CoT responses more often expose the intermediate reasoning path that leads to the final answer.
At the same time, the visualized examples also highlight that CoT is not universally beneficial: when the model lacks sufficient geometric grounding, the additional reasoning steps may amplify spatial drift and produce internally plausible but incorrect conclusions.
We therefore present both successful and failure cases to provide a balanced interpretation of the role of CoT in panoramic reasoning.

\noindent \textbf{Cross-model Qualitative Comparison.}
We also compare different MLLMs on the same questions to illustrate their distinct reasoning behaviors.
Stronger models usually exhibit better global scene awareness, maintain more stable directional consistency across multiple steps, and make more effective use of anchor objects or sparse spatial references.
By contrast, weaker models often fail at one of three stages: identifying a unique intermediate object, preserving orientation after viewpoint updates, or integrating movement and visibility constraints into a coherent final judgment.
These examples help explain why model performance differences on OmniCoT cannot be reduced to object recognition alone, but instead reflect broader differences in structured spatial reasoning ability.

\noindent \textbf{Synthetic vs.\ Real-world Cases.}
Finally, we visualize representative examples from real-world panoramas to highlight the sim-to-real challenge, as shown in Fig.~\ref{fig:vis_real}.
Compared with synthetic scenes, real-world panoramas introduce additional factors such as clutter, uneven lighting, and partial occlusion, all of which can disrupt long-horizon reasoning.
We further analyze the impact of different spatial reference settings on the reasoning process, as illustrated in Fig.~\ref{fig:coord_real}.
The qualitative comparisons show that even when a model performs reasonably well on synthetic examples, it may still fail to preserve relational consistency or visibility judgment in real-world scenes.
These cases provide an intuitive explanation for the performance degradation observed in the real-world evaluation and further justify the need for OmniCoT-Real as a complementary benchmark.

Overall, the visualizations in this subsection serve as qualitative evidence for the main empirical conclusions of the paper.
They demonstrate that performance on OmniCoT depends not only on recognizing objects in a panoramic image, but also on maintaining coherent intermediate reasoning over the entire $360^\circ$ space.

\section{Limitations}
\noindent \textbf{Environmental Scope.} OmniCoT-B and OmniCoT-T are primarily constructed from synthetic indoor scenes, as these provide the precise geometric ground truth necessary for generating high-quality QA pairs. However, it does not fully capture the complexity of outdoor scenarios, which remains unexplored. Future work will expand the benchmark to include more diverse outdoor scenarios, enriching the diversity of synthetic panoramic CoT data.

\noindent \textbf{Verification Scalability.} Manually annotating long-chain reasoning (CoT) is extremely labor-intensive. Consequently, we employed small-batch expert sampling rather than exhaustive verification of the entire dataset (which still costs considerable time). Future efforts will explore automated verifiers or model-based feedback to enhance the efficiency of CoT quality control, which will help the scaling of panoramic reasoning datasets.

\noindent \textbf{Methodological Evolution.} The current proposed OmniCoT-R1 framework relies on the existing post-training paradigm, serving as a \textbf{\textit{baseline}} of panoramic spatial reasoning. While effective, there is still room for further optimization through better architectures that more tightly couple panoramic perception with spatial reasoning. Our future work will focus on more sophisticated MLLMs for panoramic spatial reasoning based on our proposed OmniCoT, such as equipping MLLMs with active viewpoint selection capabilities to dynamically attend to task-relevant regions within a panoramic scene.

\section{Prompts} 

\subsection{Question Generation Prompt}
\begin{lstlisting} 
prompt = f"""Generate 6 spatial reasoning questions (one per type) for this 360\textdegree scene.

**Scene Description:**
{scene_description}

**Core Rules:**

1. **Natural Language Only:**
    -  While generating questions, Use ONLY natural language to describe spatial relations between objects for determining their positions, instead of using coordinate-based positioning.
    -  CORRECT: "the chair near the floor lamp", "the table closest to the north wall"
    -  CORRECT: "the piano in the southwest corner", "the stool next to the chair"
    -  WRONG: "piano(-6.9,1.2)" or any coordinate-based reference
    -  WRONG: "chair_1" or object IDs


2. **Answer Uniqueness (CRITICAL):**
   Every question must have EXACTLY ONE unambiguous answer.
   - Use spatial qualifiers: "NEAREST", "FIRST", "closest", "directly"
   - Specify relative positions: "to the left of", "north of", "between X and Y"
   - For visibility questions: Ask about ONE specific target object

3. **Wall & Window Information:**
   - Wall segments BLOCK line-of-sight (solid barriers)
   - Windows are TRANSPARENT (can see through walls with windows)

4. **Reasoning Complexity:**
    - Target: 3-4 reasoning steps 
    - Each step should be clear and verifiable from scene data
    - Avoid overly convoluted multi-hop chains

**Question Types (Generate All 6):**

 TYPE 1: Viewpoint Transform - Identify Object 
**Structure:** "Standing at [object], facing [direction/object], turn [angle1]\textdegree [dir1], then turn [angle2]\textdegree [dir2], what is the NEAREST object?"

**Requirements:**
- Start from a SPECIFIC object mentioned in the scene
- Initial facing: cardinal direction (N/S/E/W) OR another object
- Include 2 sequential rotations (e.g., "turn 90\textdegree right, then turn 45\textdegree left")
- End with distance qualifier: "NEAREST", "FIRST visible", "directly ahead"

**Example:**
"Standing at the floor lamp near the chair, facing north, turn 90\textdegree clockwise to face east, then turn 45\textdegree counterclockwise to face northeast. What is the NEAREST object directly in front of you?"

**Reasoning Steps:** 2-3 steps
1. Locate starting object position
2. Apply rotations to determine final facing direction
3. Identify nearest object in that direction

 TYPE 2: Viewpoint Transform - Angle Calculation 
**Structure:** "Standing at [object A], facing [direction/object B], turn to face [object C], then turn to face [object D]. What is the TOTAL angle turned? (Choose: 45\textdegree/90\textdegree/135\textdegree/180\textdegree)"

**Requirements:**
- Start at specific object
- Sequence: initial facing \rightarrow turn to object C \rightarrow turn to object D
- **Answer MUST be approximately 45\textdegree, 90\textdegree, 135\textdegree, or 180\textdegree (\pm10\textdegree tolerance)**

**Example:**
"Standing at the table near the window, initially facing east toward the door, turn to face the piano to your south, then turn to face the chair to your west. What is the TOTAL cumulative angle turned? (Choose closest: A) 90\textdegree B) 135\textdegree C) 180\textdegree D) 225\textdegree)"

**Reasoning Steps:** 3-4 steps
1. Calculate angle from initial direction to object C
2. Calculate angle from object C to object D
3. Sum the two angles
4. Match to closest option

 TYPE 3: Multi-Hop Object Identification 
**Structure:** "What is [relation2 + qualifier] of the [relation1 + qualifier] object of [anchor]?"

**Requirements:**
- Create 2-hop spatial relation chain
- Use distance qualifiers: "nearest", "second closest", "farthest"
- Use direction qualifiers from scene data: "north of", "to the east of", "behind"
- Ensure intermediate object is unique

**Example:**
"What object is directly to the west of the NEAREST object that is north of the piano?"

**Reasoning Steps:** 2-3 steps
1. Identify anchor object (piano)
2. Find nearest object north of piano (using Nearby + direction info)
3. Identify object west of that intermediate object

 TYPE 4: Multi-Hop Direction Identification 
**Structure:** "In which direction is the [relation + qualifier] object of [anchor], relative to [anchor]?"

**Requirements:**
- Create spatial relation chain
- Query final direction (answer: N/S/E/W/NE/NW/SE/SW)
- Use available "Nearby" direction annotations

**Example:**
"In which direction is the SECOND NEAREST object to the east of the table, relative to the table itself?"

**Reasoning Steps:** 2-3 steps
1. Identify anchor object (table)
2. Find second nearest object east of table
3. Calculate direction from table to that object

 TYPE 5: Move - Pure Translation 
**Structure:** "From [start_object], walk straight [direction] for [distance]m. What is the FIRST object you will encounter?"

**Requirements:**
- Straight-line movement (NO rotation)
- Specify cardinal direction (N/S/E/W/NE/NW/SE/SW)
- Distance: 1-3 meters (reasonable walking distance)
- Ask about FIRST object encountered

**Example:**
"From the stool near the chair, walk straight east for 2 meters. What is the FIRST object you will encounter?"

**Reasoning Steps:** 2-3 steps
1. Identify starting position
2. Calculate endpoint after 2m east movement
3. Find first object along that path

 TYPE 6: Move-Turn Combined 
**Structure:** "From [start], walk [direction] for [distance]m, then turn [angle]\textdegree [dir]. Is [specific_object] still visible?"

**Requirements:**
- Compound transformation: movement + rotation
- Specify movement direction and distance
- Specify turn angle and direction
- Ask about ONE specific target object's visibility

**Example:**
"From the floor lamp, walk north 2 meters toward the table area, then turn 90\textdegree clockwise to face east. Is the piano still visible from your new position and facing direction?"

**Reasoning Steps:** 3-4 steps
1. Calculate new position after movement
2. Apply rotation to determine final facing
3. Check if target object is in new field of view
4. Account for potential obstacles

**Output Format (valid JSON, no markdown):**
{{
  "questions": [
    {{"question": "...", "type": "viewpoint_transform_identify", "current_position_description": "...", "current_facing_description": "...", "action_description": "...", "expected_reasoning_steps": 2|3|4}},
    {{"question": "...", "type": "viewpoint_transform_angle", "current_position_description": "...", "initial_facing_target": "...", "action_description": "...", "answer_constraint": "\approx45/90/135/180\textdegree", "multiple_choice_options": ["A)45\textdegree","B)90\textdegree","C)135\textdegree","D)180\textdegree"], "expected_reasoning_steps": 2|3|4}},
    {{"question": "...", "type": "multi_hop_object", "current_position_description": "...", "current_facing_description": "...", "action_description": "...", "expected_reasoning_steps": 2|3|4}},
    {{"question": "...", "type": "multi_hop_direction", "current_position_description": "...", "current_facing_description": "...", "action_description": "...", "expected_reasoning_steps": 2|3|4}},
    {{"question": "...", "type": "move_translation", "current_position_description": "...", "current_facing_description": "...", "action_description": "...", "expected_reasoning_steps": 2|3|4}},
    {{"question": "...", "type": "move_turn_combined", "current_position_description": "...", "current_facing_description": "...", "action_description": "...", "expected_reasoning_steps": 2|3|4}}
    ]
}}


**JSON Requirements:**
- Valid JSON (no trailing commas, proper quotes)
- Question text on single line
- Use double quotes only
- Close all brackets/braces

**Critical Reminders:**
- Generate EXACTLY 6 questions (one per type)
- Use ONLY natural language for object references (NO coordinates like "piano(-6.9,1.2)")
- Leverage "Nearby" direction information
- Keep reasoning steps between 2-4 (not overly complex)
- Ensure each question has ONE unambiguous answer
- Output valid JSON as JSON Requirements section

Generate all 6 types now."""
\end{lstlisting}

\subsection{Question Scoring Prompt}
\begin{lstlisting}
prompt = f"""You are evaluating a spatial reasoning question for a 360\textdegree panoramic scene. Rate it on a 1-10 integer scale.

**Scene Description (used as reference for validation):**
{scene_description}

**Question to Evaluate:**
{question['question']}

**Question Metadata:**
- Type: {original_type}
- Expected reasoning steps: {question.get('expected_reasoning_steps', 2)}

**Evaluation Checklist (Check each point against the Scene Description):**

1. **Format Compliance - Natural Language ONLY (CRITICAL):**
    PASS: Uses natural language to describe objects and positions
      - "the chair near the floor lamp in the northern area"
      - "the piano in the southwest corner"
      - "the table closest to the north window"
   
    FAIL (Auto score \leq 2): Contains coordinate references in question text
      - "piano(-6.9,1.2)" or "chair at position (-7.0, 5.0)"
      - "(X, Y)" notation
      - "object_2.5_3.1" style IDs
   
   **Note**: Scene description contains technical IDs like "piano(-6.9,1.2)" for precision, but question text MUST NOT reference them.

2. **Object Uniqueness - Unambiguous Identification (CRITICAL):**
    PASS: Every mentioned object can be UNIQUELY determined from scene
      - Uses spatial qualifiers: "NEAREST", "FIRST", "second closest", "directly ahead"
      - Provides disambiguating context: "the chair near the table, closer to the window"
      - References nearby objects: "the stool next to the chair"
   
    FAIL (Score \leq 5): Ambiguous object references
      - "the chair" (when multiple chairs exist without clarification)
      - "what object will you see?" (without NEAREST/FIRST qualifier)
      - "turn around" (no specific angle given)

3. **Logical Consistency - Internal Coherence:**
    PASS: Objects mentioned in question exist in the scene description
      - Initial and final objects are both present
      - Spatial relationships are factually correct
      - Movement paths are physically possible
   
    FAIL (Score \leq 3): Logical errors
      - References non-existent objects
      - Contradictory spatial relations (e.g., "north of X but also south of X")

4. **Reasoning Complexity - Appropriate Depth:**
    PASS: Requires 2-4 reasoning steps based on expected_reasoning_steps
      - Not trivial (single-step lookup)
      - Not overly complex (excessive multi-hop chains)
      - Tests genuine spatial understanding
   
    FAIL (Score \leq 5): Inappropriate complexity
      - Too simple: "Where is the chair?" (direct lookup)
      - Too complex: 5+ hop chains or excessive transformations


5. **Answerability - Question Can Be Solved:**
    PASS: Has EXACTLY ONE correct answer derivable from scene data
      - All information needed is present in scene description
      - Answer is determinate (not opinion-based)
      - Question structure allows for clear conclusion
   
    FAIL (Score \leq 3): Cannot be answered definitively
      - Missing critical information
      - Multiple valid interpretations possible
      - Requires external knowledge not in scene

**Special Type-Specific Auto-Penalties:**
- **Viewpoint Transform (Type 1):**
  * Score \leq 3 if starts from abstract position (e.g., "center of room") instead of specific object
  * Score \leq 3 if angle calculation type but answer NOT close to 45\textdegree/90\textdegree/135\textdegree/180\textdegree (\pm10\textdegree)
  * Score \leq 5 if angle calculation type missing multiple choice options

- **Multi-Hop (Type 2):**
  * Score \leq 4 if missing critical qualifiers (NEAREST/second/farthest)

- **Movement (Type 3):**
  * Score \leq 4 if movement direction or distance not specified

- **Occlusion (Type 4):**
  * Score \leq 5 if missing prompt "Consider wall structures and window transparency"

**Scoring Scale:**
- **9-10 (Exceptional):**
  * Passes ALL 6 checklist items 
  * Creative use of scene features
  * Appropriate complexity for the type
  
- **7-8 (Good):**
  * Passes all critical checks (1, 2, 5, 6)
  * Minor issues in complexity or phrasing
  * Solid spatial reasoning required
  
- **5-6 (Acceptable):**
  * Passes critical checks but has issues:
    - Some ambiguity in object references
    - Slightly too simple or too complex
    - Minor factual imprecision
  
- **3-4 (Poor):**
  * Fails 1-2 critical checks:
    - Ambiguous object identification
    - Logical inconsistency
    - Factual errors
  
- **1-2 (Unacceptable):**
  * Fails multiple critical checks:
    - Uses coordinate references in question text
    - Unanswerable or nonsensical
    - Major factual errors

**Grading Instructions:**
1. Go through each checklist item systematically
2. Cross-reference object mentions with Scene Description
3. Verify spatial relationships match scene data
4. Apply auto-penalties strictly if triggered
5. Be critical: reserve 9-10 for truly exceptional questions
6. Most questions should fall in 4-8 range

**Output Format (JSON only):**
{{
  "score": <1-10 integer, or 0 for fatal flaws>
}}

Output ONLY the JSON."""
\end{lstlisting}

\subsection{CoT Generation}
\begin{lstlisting}
base_prompt = f"""
You are a Chain of Thought (CoT) reformatting expert. Your task is to 
transform descriptive reasoning into natural, step-by-step thinking
that reflects how one would logically work through a problem. Ensure
each step builds on the previous one, retains all critical information,
and concludes with a direct answer to the original question.        


**Scene Description:**
{scene_description}

**Question:**
{question}

**Scene Data Format:**
- Determine object positions exclusively through natural language descriptions of spatial relationships between objects, rather than using object coordinates.
- Camera Position: Observer's viewpoint
- Wall Vertices: Counter-clockwise boundary

**ABSOLUTE RULE - Natural Language ONLY:**
 NEVER reference coordinates: "piano(-6.9,1.2)", "(X, Y)", "position vector"
 ALWAYS use natural language: "the piano in the southwest corner", "the chair near the floor lamp"


**Question Type:** {original_type} (Reasoning Category: {mapped_type})
"""
        
        # Type-specific reasoning guidance
        if mapped_type == 'occlusion_reasoning':
            type_specific = """**Reasoning Task: Wall-Occlusion Analysis (Type 4)**
Determine whether two objects can see each other by checking wall barriers.

**NEW SCENE FORMAT FEATURES:**
- Wall Vertices listed in counter-clockwise order
- Wall Segments show connections
- Windows are explicitly listed as objects with "(direction: ...)"

**Reasoning Steps Required (2-4 steps):**

1. **Identify Object Positions (Step 1):**
   - Locate Object A and Object B in scene description
   - Describe their positions using natural language
   -  AVOID: "piano(-6.9,1.2)" or coordinates
   -  USE: "the piano in the southern area", "the table near the north window"
   - Note their general spatial relationship (e.g., "on opposite sides of the room")

2. **Determine Spatial Relationship (Step 2):**
   - Describe which direction Object B is from Object A
   - Check if they are in the same general area or separated
   - Example: "The table is to the north of the piano"

3. **Check for Wall Barriers (Step 3):**
   - Review the "Wall Segments" in scene description
   - Identify if any wall segments lie between the two objects
   - **Simplified approach:**
     * If both objects are in the same open area (no walls mentioned between them): Likely DIRECT LINE OF SIGHT
     * If they are described in different "rooms" or "areas" separated by walls: Check for windows

4. **Check for Windows (Step 4):**
   - If walls exist between objects, check if windows are present
   - Windows are TRANSPARENT (can see through)
   - Look for window objects in the wall segment area
   - Example: "window near the north wall" indicates visibility through that wall

5. **Final Visibility Conclusion (Step 4):**
   Just state "Yes" or "No" based on visibility

"""
        
        elif mapped_type == 'multi_step_viewpoint':
            type_specific = """**Reasoning Task: Multi-Hop Relationship Reasoning (Type 2)**
This includes:
- Type 2.1: Multi-hop object identification (what is X of Y of Z?)
- Type 2.2: Multi-hop direction identification (in which direction is X relative to Y?)

**NEW SCENE FORMAT FEATURES YOU CAN USE:**
- "Nearby" field lists neighboring objects with DIRECTION 
- Use this to quickly find objects in specific directions
- Distance information helps apply qualifiers like "nearest"

**Reasoning Steps Required (2-3 steps):**

1. **Identify Anchor Object (Step 1):**
   - Locate the starting/anchor object in scene description
   - Describe its location using natural language
   -  AVOID: "table(-8.2,5.7)" or coordinate references
   -  USE: "the table near the north window", "the table in the northern area"

2. **Apply First Spatial Relation + Qualifier (Step 2):**
   - Identify the first relation (e.g., "south of", "behind", "to the east of")
   - Apply distance qualifier: "nearest", "second closest"
   - **Use "Nearby" direction annotations to filter candidates:**
     * Example: Looking for "nearest object north of table"
     * Check table's "Nearby" list for objects with "(direction: north)"
     * Sort by distance, select based on qualifier
   - Clearly state the intermediate object you selected

3. **Apply Second Relation (if multi-hop) (Step 3):**
   - From the intermediate object, apply second spatial relation
   - Use intermediate object's "Nearby" list
   - Select final target based on qualifiers
   - For Type 2.2 (direction): Calculate compass direction from anchor to target

4. **State Final Answer (Step 3 or 4):**
   - For Type 2.1: State the final target object name
   - For Type 2.2: State the compass direction (N/S/E/W/NE/NW/SE/SW)

**Key Reminders:**
- Each hop builds on the previous result
- Show intermediate objects explicitly
- Use "Nearby" direction annotations to speed up identification
- Natural language descriptions ONLY
"""
        
        elif mapped_type == 'move_translation':
            type_specific = """**Reasoning Task: Movement & Translation Simulation (Type 3)**
This includes:
- Type 3.1: Pure translation (move_translation) - straight-line movement, no rotation
- Type 3.2: Combined move-turn (move_turn_combined) - movement + rotation

**Reasoning Steps Required:**

1. **Identify Starting Position:**
   - Locate the starting object/position mentioned
   - Extract its position from scene description,describe using natural language and NOT coordinates.
   - Determine initial facing direction (if specified)

2. **Simulate Movement Vector:**
   - Identify movement direction (e.g., "walk north", "move toward X")
   - Note distance if specified (e.g., "3 meters", "until reaching X")
   - Calculate approximate endpoint position
   - Consider which objects will be passed during movement
   - Update observer's position mentally

3. **Apply Rotation (if Combined Move-Turn):**
   - After movement, check if rotation is specified
   - Note rotation angle (e.g., 90 degrees, 180 degrees) and direction (left/right)
   - Calculate new facing direction after rotation
   - Determine what becomes visible after turning

4. **Predict Field of View:**
   - Based on final position and facing direction:
     * For "what is the FIRST object you see": Identify closest object in direct line of sight
     * For "what is NEAREST": Find closest object from new position
     * For "is X visible": Check if target object is within view cone and not occluded
   - Consider viewing angle (typically 360 degrees for panoramic, or frontal cone)
   - Account for occlusions (walls, other objects)


**Key Distinctions:**
- Pure Translation (Type 3.1): Movement ONLY, no rotation involved
- Combined Move-Turn (Type 3.2): Movement FOLLOWED BY rotation (check temporal order)
"""
        
        else:  
            type_specific = """**Reasoning Task: Viewpoint Transformation (Type 1)**
This includes:
- Type 3.1: Pure translation - straight movement, no rotation
- Type 3.2: Combined move-turn - movement + rotation

**NEW SCENE FORMAT FEATURES YOU CAN USE:**
- "Distance to Camera" helps estimate object positions
- "Nearby" direction annotations show spatial layout
- Use these to predict movement paths

**Reasoning Steps Required (2-4 steps):**

1. **Identify Starting Position & Facing (Step 1):**
   - Locate starting object in scene description
   - Use natural language: "the stool near the chair in the northern area"
   - Determine initial facing direction if specified

2. **Simulate Movement Vector (Step 2):**
   - Identify movement direction 
   - Note distance if specified
   - Consider which objects will be encountered along path

3. **Apply Rotation (if Combined Move-Turn) (Step 3):**
   - Check if rotation is specified
   - Note angle (e.g., 90\textdegree) and direction (clockwise/counterclockwise)
   - Calculate new facing direction after rotation
   - Determine what becomes visible after turning

4. **Predict Visible Object (Step 3 or 4):**
   - Based on final position and facing:
     * For "FIRST object": Identify closest object in direct path
     * For "NEAREST": Find closest object from new position using "Nearby" info
     * For "is X visible?": Check if target is in view cone and not occluded
   - Use "Nearby" lists of objects in target area to identify candidates
   - Account for distance and direction

**Key Distinctions:**
    - Use "Nearby" distances to estimate encounter points
    - Natural language ONLY for positions   

"""
        
        # Unified output requirements
        full_prompt = base_prompt + type_specific + """

**Output Requirements:**
- Use natural language descriptions (NO coordinate IDs like "Chair_2.5_3.1" or (2.5, 3.1))
- Show clear step-by-step reasoning following the structure above
- Use transition words: "First," "Next," "Then," "Finally"
- Reference specific details from scene description (object names, positions, wall segments)
- Be thorough but concise (aim for 200-400 words depending on complexity)
- End with a clear, direct answer to the question

Provide your detailed reasoning now:"""
        
        return full_prompt
\end{lstlisting}
\subsection{Answer Summarization Prompt}
\begin{lstlisting}
            if mapped_type == 'occlusion_reasoning':
            type_guidance = """**For Occlusion Reasoning (Type 4):**

            
**COT Requirements (3-5 steps, natural language ONLY):**
1. **Step 1** - Describe both objects' positions:
   - CORRECT: "Object A is near the west wall in the northern section"
   - WRONG: "Object A is at position (-10.0, 4.9)"
   
2. **Step 2** - Determine spatial relationship:
   - Use cardinal directions and relative positions
   - CORRECT: "Object B is to the northeast of Object A"
   
3. **Step 3** - Check wall barriers:
   - Describe wall segments in natural language
   - CORRECT: "The northern and southern areas are separated by a wall segment running east-west"
   - WRONG: "Wall segment from V5 to V6"
   
4. **Step 4** (if needed) - Check for windows:
   - Mention window presence using natural language
   - CORRECT: "There is a window on the north wall that would allow visibility"
   
5. **Step 5** - State conclusion:
   - Clear "Yes" or "No" with brief reason
   - CORRECT: "Yes, they can see each other through the north window."

**Answer Format Example:**
Answer: Yes
COT:
1. First, Object A (sofa chair) is located near the west wall in the northern area of the room.
2. Next, Object B (table) is positioned near the north window, to the east-northeast of the sofa chair.
3. Then, I check for wall barriers between them. Both objects are in the northern section of the room, which appears to be an open area without interior walls separating them.
4. Finally, since no solid walls obstruct the line of sight between the sofa chair and the table, the answer is: Yes, there is a direct line of sight.


"""

        elif mapped_type == 'viewpoint_transform':
            type_guidance = """**For Viewpoint Transformation (Type 1):**
**COT Requirements (3-5 steps, natural language ONLY):**
1. **Step 1** - Identify starting object:
   - CORRECT: "The chair is in the northern area, near the floor lamp and table"
   - WRONG: "Starting at chair(-7.0, 5.0)"
   
2. **Step 2** - State initial facing:
   - CORRECT: "Initially facing north toward the table"
   - Use cardinal directions or relative to objects
   
3. **Step 3-4** - Apply rotations sequentially:
   - For each rotation, state angle and direction
   - CORRECT: "After turning 90\textdegree clockwise, now facing east"
   - Track cumulative orientation using compass directions
   
4. **Step 4-5** - Identify final object (Type 1.1) or calculate angle (Type 1.2):
   - **For Type 1.1:**
     * CORRECT: "Looking northeast, the nearest object is the floor lamp, about 0.9m away"
     * Use distance qualifiers from question
   
   - **For Type 1.2:**
     * CORRECT: "Total angle turned: 90\textdegree + 90\textdegree = 180\textdegree, which matches option C"
     * State calculated angle AND multiple choice option

**Answer Format Example (Type 1.1):**
Answer: Floor lamp
COT:
1. First, I identify the starting position at the chair in the northern area, near the floor lamp and table.
2. Next, the initial facing direction is north, toward the table.
3. Then, turning 90\textdegree clockwise changes the facing from north to east.
4. After that, turning 45\textdegree counterclockwise from east results in facing northeast.
5. Finally, looking northeast from the chair, the nearest object in that direction is the floor lamp, approximately 0.9m away (as indicated in the chair's nearby objects list).

**Answer Format Example (Type 1.2):**
Answer: 180\textdegree
COT:
1. First, starting at the table near the north window, initially facing east toward the door.
2. Next, turning to face the piano (south of the table) requires a 90\textdegree turn to the right (from east to south).
3. Then, turning to face the chair (west of the table) requires another 90\textdegree turn to the right (from south to west).
4. Finally, the total angle turned is 90\textdegree + 90\textdegree = 180\textdegree, which corresponds to option C in the multiple choice.
"""


        elif mapped_type == 'move_translation':
            type_guidance = """**For Move & Translation Simulation (Type 3):**

Includes Type 3.1 (pure translation) and Type 3.2 (move + turn)

Use "Nearby" distances to predict movement paths**
**For Type 3.1 (Pure Translation - No Rotation):**

1. **Step 1** - Identify starting position:
   - CORRECT: "Starting from the stool in the northern area, near the chair and table"
   - WRONG: "Starting from position (-7.3, 4.5)"

2. **Step 2** - Describe movement vector:
   - CORRECT: "Walking straight east for 2 meters. The stool's Nearby list shows the table is 0.9m east"
   - WRONG: "Movement vector: +2 in X-axis"

3. **Step 3** - Predict objects along path:
   - CORRECT: "At 0.9m, I pass the table. At 2m, I reach near the floor lamp area"
   - Use Nearby distances to estimate encounters

4. **Step 4** - Identify first/nearest object:
   - CORRECT: "The FIRST object encountered is the table at 0.9m"
   - Apply qualifier from question (FIRST/NEAREST)

**For Type 3.2 (Combined Move-Turn - Movement + Rotation):**

1. **Step 1** - Identify starting position & initial facing:
   - CORRECT: "Starting from the floor lamp in the northern area, initially facing north"

2. **Step 2** - Simulate movement:
   - CORRECT: "Walking north 2 meters brings me near the north wall/window area"
   - WRONG: " position: (-6.2, 7.2)"
   - Describe endpoint using nearby landmarks

3. **Step 3** - Apply rotation:
   - CORRECT: "Turning 90\textdegree clockwise changes facing from north to east"
   - Calculate new facing direction after rotation

4. **Step 4** - Determine target object visibility:
   - CORRECT: "The piano is in the southern area, behind me (facing east). Not visible in frontal view"
   - Or: "In 360\textdegree panoramic view, the piano remains visible"
   - Consider viewing mode (frontal vs panoramic)

5. **Step 5** - State visibility conclusion:
   - Answer "Yes" or "No" with reason

**Answer Format Example (Type 3.1 - Pure Translation):**
Answer: Table
COT:
1. First, starting from the stool in the northern area, near the chair and table.
2. Next, walking straight east for 2 meters. The Nearby list shows the table is 0.9m to the east, and the floor lamp is 1.4m to the northeast.
3. Then, along the eastward path, the table is encountered first at 0.9m, followed by approaching the floor lamp area.
4. Finally, the FIRST object encountered is the table.

**Answer Format Example (Type 3.2 - Move-Turn, Frontal View):**
Answer: No
COT:
1. First, starting from the floor lamp in the northern area.
2. Next, walking north 2 meters brings me close to the north wall near the window.
3. Then, turning 90\textdegree clockwise changes my facing from north to east.
4. After that, the piano is in the southern area of the room. From my position near the north wall facing east, the piano is behind me to the south.
5. Finally, in a frontal field of view, the piano is not visible. Answer: No.

**Answer Format Example (Type 3.2 - Move-Turn, 360\textdegree Panoramic):**
Answer: Yes
COT:
1. First, starting from the floor lamp, walking north 2 meters toward the north wall area.
2. Next, turning 90\textdegree clockwise to face east.
3. Then, checking the piano's position: it's in the southern area, behind my current facing direction.
4. After that, in a 360\textdegree panoramic scene, all objects remain visible regardless of facing direction.
5. Finally, the piano is still visible in the panoramic view. Answer: Yes.

**Key Distinction:**
- Type 3.1: Movement ONLY \rightarrow identify object along path
- Type 3.2: Movement \rightarrow Rotation \rightarrow check visibility
"""

        else:  # Multi-Hop Relationship Reasoning
            type_guidance = """**For Multi-Hop Relationship Reasoning (Type 2):**
Includes Type 2.1 (object identification) and Type 2.2 (direction identification)

Use "Nearby" direction annotations from scene

**COT Requirements (3-5 steps, natural language ONLY):**
1. **Step 1** - Identify anchor object:
   - CORRECT: "The piano is in the southern area of the room"
   - WRONG: "Starting from piano(-6.9,1.2)"

2. **Step 2** - Apply first relation + qualifier:
   - CORRECT: "Finding the nearest object north of the piano. The chair is 0.4m away in the north direction (from Nearby list)"
   - WRONG: "Finding object where Y > Y_piano, minimum distance"
   - Explicitly name the intermediate object

3. **Step 3** - Apply second relation (if multi-hop):
   - CORRECT: "From the chair, looking west. The stool is 0.6m away to the southwest"
   - Use intermediate object's Nearby list

4. **Step 4** - State final answer:
   - **For Type 2.1 (object):** Name the target object clearly
   - **For Type 2.2 (direction):** State compass direction (N/S/E/W/NE/NW/SE/SW)

**Answer Format Example (Type 2.1 - Multi-Hop Object):**
Answer: Stool
COT:
1. First, the anchor object is the piano, located in the southern area.
2. Next, I find the nearest object north of the piano. According to the Nearby list, the chair is 0.4m away in the north direction.
3. Then, from the chair, I look for an object to the west. The chair's Nearby list shows the stool is 0.6m away to the southwest, closest to the west direction.
4. Finally, the two-hop chain (piano \rightarrow chair \rightarrow stool) identifies the stool as the target object.

**Answer Format Example (Type 2.2 - Multi-Hop Direction):**
Answer: Northeast
COT:
1. First, the anchor object is the table near the north window.
2. Next, I find the second nearest object to the east. The chair is 1.4m southeast, and the floor lamp is farther east at about 1.7m, making it the second nearest.
3. Then, I calculate the direction from the table to the floor lamp. Based on their positions, the floor lamp is to the northeast of the table.
4. Finally, the direction is northeast.
"""

        # Build complete prompt
        prompt = f"""You are an expert at summarizing spatial reasoning into clear, concise answers.
**Question:**
{question}
**Question Type:** {original_type} (Answer Format Category: {mapped_type})
**Reasoning Process:**
{reasoning}
{type_guidance}

**Your Task:**
Based on the reasoning above, provide:
1. **Answer**: The final concise answer
2. **COT**: 3-5 numbered steps leading to the answer

**ABSOLUTE REQUIREMENTS:**
- Use natural language descriptions ONLY in COT
- NO coordinate IDs like "piano(-6.9,1.2)" or "(X, Y)"
- Each COT step must start with transition word: "First,", "Next,", "Then,", "After that,", "Finally,"
- COT should read naturally, like explaining to a human
- Be specific: reference nearby objects, directions, distances mentioned in reasoning

**Output Format:**

Answer: [object name, or angle, or "Yes"/"No"]
COT: [ a step-by-step chain-of-thought that leads to the answer, structured into 3-5 numbered reasoning steps.(NO coordinate IDs like "Chair_2.5_3.1" or (2.5, 3.1)]

Provide your response now:"""
\end{lstlisting}

\subsection{CoT Quality Scoring Prompt}
\begin{lstlisting}
prompt = f"""Rate the quality of spatial reasoning and answer on a 1-10 scale for each.

    **Scene Description (for reference validation):**
    {scene_description}

    **Question:**
    {question}

    **Reasoning Process:**
    {cot}

    **Final Answer:**
    {final_answer}
    {optional_info}

    **Question Type:** {original_type} (Category: {mapped_type})

    **CRITICAL RULE - Natural Language ONLY:**
    Both reasoning and answer MUST use natural language descriptions.
     AUTO-FAIL (Score \leq 3) if contains:
    - Coordinate IDs: "piano(-6.9,1.2)", "chair at (-7.0, 5.0)"
    - Position vectors: "(X, Y)", "position (2.5, 3.1)"
    - Object IDs: "table_-8.2_5.7"
     REQUIRED format:
    - "the piano in the southwest corner"
    - "the chair near the floor lamp in the northern area"
    - "the table close to the north window"

    **Reasoning Score Evaluation (1-10):**

    **Check 1: Format Compliance (CRITICAL - Auto-fail if violated)**
    CORRECT: Uses natural language ONLY to describe objects and positions
    CORRECT: NO coordinate references anywhere in reasoning
    WRONG: Contains "object(-X,Y)" \rightarrow Score \leq 2

    **Check 2: Reasoning Structure (Expected: 2-4 steps)**
    CORRECT: Structured into clear logical steps (typically 2-4 steps based on question type)
    CORRECT: Each step follows from the previous one
    CORRECT: Uses transition words: "First,", "Next,", "Then,", "Finally,"
    WRONG: Too few steps (trivial, single-step lookup) \rightarrow Score \leq 4
    WRONG: Too many steps (overly complex, >5 steps) \rightarrow Score \leq 6

    **Check 3: Use of Scene Information**
    CORRECT: References specific objects from scene description
    CORRECT: Spatial relationships match scene data (verify directions/distances against scene)
    WRONG: Contradicts scene information \rightarrow Score \leq 4
    WRONG: Ignores available spatial clues \rightarrow Score \leq 6

    **Check 4: Type-Specific Requirements**

    **Type 1 (Viewpoint Transform):**
    - CORRECT: Identifies starting SPECIFIC OBJECT clearly
    - CORRECT: Tracks initial facing direction
    - CORRECT: Type 1.1: Applies each rotation sequentially (\geq2 rotations required)
    - CORRECT: Type 1.2: Calculates angles and sums total (should \approx 45\textdegree/90\textdegree/135\textdegree/180\textdegree)
    - WRONG: Starts from abstract position \rightarrow Score \leq 3
    - WRONG: Type 1.1 with only 1 rotation \rightarrow Score \leq 4

    **Type 2 (Multi-Hop):**
    - CORRECT: Shows intermediate object at each hop explicitly
    - CORRECT: Applies distance/direction qualifiers correctly (nearest/second/north of)
    - CORRECT: Type 2.1: Reasoning chain leads to target object
    - CORRECT: Type 2.2: Calculates final direction from anchor to target
    - WRONG: Only 1 hop (no intermediate) \rightarrow Score \leq 4
    - WRONG: Missing qualifiers \rightarrow Score \leq 5

    **Type 3 (Movement):**
    - CORRECT: Describes movement vector (direction + distance)
    - CORRECT: Type 3.1: Identifies objects along straight path
    - CORRECT: Type 3.2: Applies movement THEN rotation in sequence
    - CORRECT: Type 3.2: Checks target visibility from new position/facing
    - WRONG: Movement vector unclear \rightarrow Score \leq 5

    **Type 4 (Occlusion):**
    - CORRECT: Identifies both object positions
    - CORRECT: Checks for wall segments between them
    - CORRECT: Considers window transparency (windows = can see through)
    - CORRECT: Draws correct visibility conclusion

    **Check 5: Logical Coherence**
    CORRECT: Each step logically follows from previous
    CORRECT: No contradictory statements
    CORRECT: Reasoning leads to the stated final answer
    WRONG: Contradictory logic \rightarrow Score \leq 3
    WRONG: Reasoning doesn't support answer \rightarrow Score \leq 4

    **Reasoning Score Scale:**
    - **9-10 (Excellent):** Passes all checks, perfect format, leverages scene features effectively, appropriate complexity
    - **7-8 (Good):** Solid reasoning, passes critical checks, minor imperfections in detail or structure
    - **5-6 (Acceptable):** Meets basic requirements, some issues in clarity or completeness
    - **3-4 (Poor):** Fails multiple checks, significant logic gaps, or format violations
    - **1-2 (Unacceptable):** Contains coordinates, contradictory logic, or fundamentally flawed

    ---

    **Answer Score Evaluation (1-10):**

    **Check 1: Format Compliance (CRITICAL)**
    CORRECT: Uses natural language description ONLY
    WRONG: Contains coordinate IDs \rightarrow Score \leq 2

    **Check 2: Correctness**
    CORRECT: Factually correct based on scene data
    CORRECT: Consistent with reasoning process
    CORRECT: Matches expected answer type:
    - Type 1.1 / 2.1 / 3.1: Specific object name
    - Type 1.2: Angle value (e.g., 45\textdegree, 90\textdegree)
    - Type 2.2: Compass direction (N/S/E/W/NE/NW/SE/SW)
    - Type 3.2 / 4: Yes/No
    WRONG: Factually wrong \rightarrow Score \leq 4
    Wrong answer type \rightarrow Score \leq 6


    **Answer Score Scale:**
    - **9-10 (Excellent):** Correct, precise, perfectly aligned with reasoning, ideal format
    - **7-8 (Good):** Correct answer, minor format or precision issues
    - **5-6 (Acceptable):** Partially correct or lacks precision
    - **3-4 (Poor):** Wrong answer or contradicts reasoning
    - **1-2 (Unacceptable):** Completely wrong, contains coordinates, or nonsensical

    ---

    **Output Format (JSON only, no explanation):**
    {{
    "reasoning_score": <1-10 integer>,
    "answer_score": <1-10 integer>
    }}

    **Grading Instructions:**
    1. Check CRITICAL format compliance first (coordinates \rightarrow auto-fail)
    2. Verify reasoning structure (2-4 steps expected)
    3. Cross-reference spatial claims with scene description
    4. Apply type-specific requirements strictly
    5. Be critical: reserve 9-10 for exceptional quality
    6. Most should fall in 4-8 range

    Output ONLY the JSON."""
\end{lstlisting}

\subsection{Evaluation Prompt (w/ Reference Points)}
\begin{lstlisting}
Coordinate System: X-axis: Positive direction is EAST\nY-axis: Positive direction is NORTH \n

Reference Objects Locations (Randomly selected, not represent all objects):{coord_info}\n\n

Question: {question}\n\n

Requirements:Answer the question based on the provided ERP-formatted image. 

Example: <think> Your reasoning here </think> <answer> Your answer here </answer> \n\n 

IMPORTANT: You MUST follow this format EXACTLY:\n
1. Start with <think> tag\n
2. Write your complete step-by-step reasoning (at least 3-5 sentences)\n
3. End the reasoning with </think> tag\n4. 

Provide your final answer between <answer> and </answer> tags using one to three words (e.g., Yes, No, Southwest, 135\textdegree). If you can not answer the question, answer with '<answer>Failed to Answer</answer>'. \n\n

ANSWER FORMAT CONSTRAINTS:\n
- If the question involves direction, answer with one of the eight cardinal directions: North, South, East, West, Northeast, Northwest, Southeast, Southwest\n
- If the question involves angle, answer in the format: xxx\textdegree (e.g., 135\textdegree, 180\textdegree, 90\textdegree)\n
- If the question involves a yes/no judgment, answer with either \"Yes\" or \"No\"\n
- If your answer is an object name, you MUST choose from the following object pool:\nwindow, door, chair, table, picture, shelf, bottom_cabinet, lamp, cabinet, sofa, rug, sink, bed, trash_can, mirror, tv, coffee_table, top_cabinet, counter, sofa_chair, nightstand, toilet, fridge, box, pillow, stool, plant, basket, floor_lamp, curtain, wall_clock, office_chair, oven, dishwasher, standing_tv, washer, wall_mounted_tv, piano, dryer, desk, chest, bathtub, console_table, microwave, table_lamp, shower, crib, heater, monitor, cooktop, stove, towel_rack, fence, laptop, range_hood, loudspeaker, towel, cushion, guitar, pool_table, coffee_machine, speaker_system\n\n

Do NOT output anything outside the <think></think> and <answer></answer> tags.\n
Do NOT skip the reasoning process even for simple questions.
\end{lstlisting}

\subsection{Evaluation Prompt (w/o Reference Points)}
\begin{lstlisting}
Coordinate System: X-axis: Positive direction is EAST\n
Y-axis: Positive direction is NORTH\n\n

Question: {question}\n\n

Requirements:Answer the question based on the provided ERP-formatted image. 

Example: <think> Your reasoning here </think> <answer> Your answer here </answer> \n\n 

IMPORTANT: You MUST follow this format EXACTLY:\n
1. Start with <think> tag\n
2. Write your complete step-by-step reasoning (at least 3-5 sentences)\n
3. End the reasoning with </think> tag\n4. 

Provide your final answer between <answer> and </answer> tags using one to three words (e.g., Yes, No, Southwest, 135\textdegree). If you can not answer the question, answer with '<answer>Failed to Answer</answer>'. \n\n

ANSWER FORMAT CONSTRAINTS:\n
- If the question involves direction, answer with one of the eight cardinal directions: North, South, East, West, Northeast, Northwest, Southeast, Southwest\n
- If the question involves angle, answer in the format: xxx\textdegree (e.g., 135\textdegree, 180\textdegree, 90\textdegree)\n
- If the question involves a yes/no judgment, answer with either \"Yes\" or \"No\"\n
- If your answer is an object name, you MUST choose from the following object pool:\nwindow, door, chair, table, picture, shelf, bottom_cabinet, lamp, cabinet, sofa, rug, sink, bed, trash_can, mirror, tv, coffee_table, top_cabinet, counter, sofa_chair, nightstand, toilet, fridge, box, pillow, stool, plant, basket, floor_lamp, curtain, wall_clock, office_chair, oven, dishwasher, standing_tv, washer, wall_mounted_tv, piano, dryer, desk, chest, bathtub, console_table, microwave, table_lamp, shower, crib, heater, monitor, cooktop, stove, towel_rack, fence, laptop, range_hood, loudspeaker, towel, cushion, guitar, pool_table, coffee_machine, speaker_system\n\n

Do NOT output anything outside the <think></think> and <answer></answer> tags.\n
Do NOT skip the reasoning process even for simple questions.
\end{lstlisting}

\subsection{Evaluation Prompt (Viewpoint Consistency)}
\begin{lstlisting}
You are an expert in evaluating spatial reasoning quality.

Task: Evaluate viewpoint consistency in the reasoning.

Definitions:
- "Viewpoint-related statement" = a claim about directions/orientation/relative position that depends on the agent pose
  (left/right/front/back, facing, turn, look toward, behind, clockwise, north/east, etc.).

Rules:
1) Extract viewpoint-related statements from the REASONING.
   - Prefer statements that are USED TO SUPPORT the final answer (e.g., "after turning, I face northeast so X is in front").
   - Do NOT list mere restatements of the QUESTION instruction unless the reasoning relies on it to infer geometry.
   - Merge duplicates / near-duplicates (avoid splitting one idea into many tiny statements).
2) Score each extracted statement against the Ground Truth.
3) If NO viewpoint-related statements are found in the reasoning, return {"statements": []}.

Scoring:
- score must be a float in [0.0, 1.0] with exactly 1 decimal place (step=0.1).
  - 0.0: Clear contradiction with Ground Truth.
  - 0.1-0.4: Mostly incorrect or highly ambiguous.
  - 0.5-0.8: Mostly consistent, minor underspecification.
  - 0.9-1.0: Fully consistent and precise.

Inputs:
[Ground Truth]
{ground_truth}

[Question]
{question}

[Reasoning to Evaluate]
{reasoning}

[Reference Reasoning]
{reference_cot}

[Correct Answer]
{answer}

Output Requirements:
- Output ONLY valid JSON.
- No extra keys other than "statements".
- Each statement item must include:
  - "text": quote or faithful paraphrase from reasoning
  - "score": float with 1 decimal

Output JSON Schema:
{
  "statements": [
    { "text": "...", "score": 0.3 },
    { "text": "...", "score": 0.9 }
  ]
}
\end{lstlisting}

\subsection{Evaluation Prompt (Spatial Evidence Sufficiency)}
\begin{lstlisting}
You are an expert in evaluating spatial reasoning quality.

Task: Evaluate spatial evidence sufficiency.

Goal:
Identify the MINIMAL set of key spatial relationships NECESSARY to justify the conclusion for THIS question,
then check whether the reasoning provides evidence for each relationship.

Definitions:
- "Key spatial relationship" = a spatial relation necessary to justify the final answer (not just related facts).
  Examples: adjacency, relative direction, containment, above/below, in front/behind, nearest, distance ordering,
  object location with respect to a landmark.
- "Evidence" must be a direct quote or near-quote from the provided reasoning. Do NOT invent evidence.

Rules:
1) List only the minimal necessary relationships. Do NOT enumerate everything in Ground Truth.
2) Recognize alternative valid reasoning paths: do NOT require matching the reference reasoning.
   The reference reasoning is ONLY a hint for possible relationships, not a required path.
3) STRICT EVIDENCE RULE:
   - If you cannot point to a quote/near-quote in the reasoning, you MUST set score=0.0 and evidence=null for that relationship.
4) If the reasoning contains no usable spatial evidence at all, return {"relationships": []}.

Scoring:
- score must be a float in [0.0, 1.0] with exactly 1 decimal place (step=0.1).
For each relationship:
  - 0.0: Not supported by the reasoning OR no evidence quote available.
  - 0.1-0.4: Very weak/partial (mentions objects but not the needed relation).
  - 0.5-0.8: Clearly supported but slightly underspecified.
  - 0.9-1.0: Clearly and sufficiently supported with precise relation.

Inputs:
[Ground Truth]
{ground_truth}

[Question]
{question}

[Reference Reasoning]
{reference_cot}

[Reasoning to Evaluate]
{reasoning}

Output Requirements:
- Output ONLY valid JSON.
- No extra keys other than "relationships".
- Each relationship item must include:
  - "text": concise relationship sentence
  - "score": float with 1 decimal
  - "evidence": quote/near-quote from reasoning if score>0; if score==0.0 set null

Output JSON Schema:
{
  "relationships": [
    { "text": "...", "score": 0.0, "evidence": null },
    { "text": "...", "score": 0.7, "evidence": "..." }
  ]
}

\end{lstlisting}

\subsection{Evaluation Prompt (Reasoning Feasibility)}
\begin{lstlisting}
You are an expert in evaluating spatial reasoning quality.

Task: Evaluate reasoning feasibility by extracting execution steps and judging feasibility against Ground Truth.

Definitions:
- "Execution step" = any implied action/movement/perceptual operation by the agent, including:
  (a) body movement: walk/move/go/approach
  (b) viewpoint actions: turn/rotate/facing/clockwise/counterclockwise/look toward/look at/scan
  (c) interaction: reach/grab/open/close/pick up

Rules:
1) Evaluate feasibility of EACH step using Ground Truth.
2) IMPORTANT COVERAGE RULE:
   - If the reasoning contains ANY viewpoint-action cue (e.g., "turn", "face", "clockwise", "counterclockwise", "look"),
     you MUST extract at least 1 step (do NOT return empty).
3) Return {"steps": []} ONLY if the reasoning contains no actions AND no viewpoint-action cues (pure static description).

Hard constraints (apply strictly):
- If a step refers an object/location not present in Ground Truth, score <= 0.2.
- If a step implies passing through walls/closed barriers without evidence, score <= 0.2.
- If a step requires an impossible move (teleport, impossible distance in one step), score <= 0.2.

Scoring:
- score must be a float in [0.0, 1.0] with exactly 1 decimal place (step=0.1).
  - 0.0: Clearly infeasible.
  - 0.1-0.4: Strongly doubtful / violates constraints.
  - 0.5-0.8: Feasible with minor missing details.
  - 0.9-1.0: Clearly feasible and consistent with Ground Truth.

Inputs:
[Ground Truth]
{ground_truth}

[Question]
{question}

[Reasoning to Evaluate]
{reasoning}

[Reference Reasoning]
{reference_cot}

[Correct Answer]
{answer}

Output Requirements:
- Output ONLY valid JSON.
- No extra keys other than "steps".
- Each step item must include:
  - "text": a quote or faithful paraphrase from the reasoning
  - "score": float with 1 decimal

Output JSON Schema:
{
  "steps": [
    { "text": "...", "score": 0.9 },
    { "text": "...", "score": 0.1 }
  ]
}

\end{lstlisting}

\end{document}